\pdfoutput=1
\documentclass[11pt]{article}

\makeatletter
\@ifl@t@r\fmtversion{2018/12/01}{}{\PassOptionsToPackage{draft}{hyperref}}
\makeatother
\PassOptionsToPackage{table,xcdraw}{xcolor}
\usepackage[preprint]{acl}

\usepackage{times}
\usepackage{latexsym}
\usepackage[T1]{fontenc}
\usepackage[utf8]{inputenc}
\usepackage{microtype}
\IfFileExists{inconsolata.sty}{\usepackage{inconsolata}}{}

\usepackage{float}
\usepackage{amsmath}   %
\usepackage{amssymb}
\newif\ifhavemultirow
\IfFileExists{multirow.sty}{\havemultirowtrue}{}
\ifhavemultirow \usepackage{multirow}\else \newcommand{\multirow}[3]{#3}\fi
\usepackage{booktabs}
\usepackage{pifont}
\usepackage{algorithm}
\usepackage{algorithmic}
\usepackage{graphicx}
\usepackage{subcaption}
\usepackage{enumitem}
\newif\ifhavetcolorbox
\IfFileExists{tcolorbox.sty}{\havetcolorboxtrue}{}
\ifhavetcolorbox
\usepackage{tcolorbox}
\tcbset{
  colback=white,
  colframe=black,
  boxrule=0.4pt,
  arc=1mm,
  left=4pt,
  right=4pt,
  top=4pt,
  bottom=4pt
}

\newenvironment{promptbox}[1]{%
  \begin{tcolorbox}
  \textbf{#1}\\[0.3em]
  \footnotesize\ttfamily
}{%
  \end{tcolorbox}
}
\else
\newsavebox{\promptboxbox}

\fi

\newcommand{\nCrossUnique}{122}

\newcommand{\detBestFlashCross}{0.75}
\newcommand{\detBestFlashImage}{0.82}
\newcommand{\detBestFlashLiteCross}{0.72}
\newcommand{\detBestFlashLiteImage}{0.73}
\newcommand{\detBestFlashLiteText}{0.80}
\newcommand{\detBestFlashText}{0.85}
\newcommand{\detBestGemmaCross}{0.81}
\newcommand{\detBestGemmaImage}{0.77}
\newcommand{\detBestGemmaText}{0.84}
\newcommand{\detBestInternVLCross}{0.77}
\newcommand{\detBestInternVLImage}{0.83}
\newcommand{\detBestInternVLText}{0.85}
\newcommand{\detBestIsSECount}{16}
\newcommand{\detBestMistralCross}{0.81}
\newcommand{\detBestMistralImage}{0.87}
\newcommand{\detBestMistralText}{0.87}
\newcommand{\detBestProCross}{0.71}
\newcommand{\detBestProImage}{0.73}
\newcommand{\detBestProText}{0.87}
\newcommand{\detBestQwenThreeEightCross}{0.67}
\newcommand{\detBestQwenThreeEightImage}{0.76}
\newcommand{\detBestQwenThreeEightText}{0.83}
\newcommand{\detBestQwenThreeFiveCross}{0.73}
\newcommand{\detBestQwenThreeFiveImage}{0.82}
\newcommand{\detBestQwenThreeFiveText}{0.86}
\newcommand{\detBestQwenVLCross}{0.82}
\newcommand{\detBestQwenVLImage}{0.85}
\newcommand{\detBestQwenVLText}{0.89}
\newcommand{\detCFApiCells}{9}
\newcommand{\detCFApiPool}{4}
\newcommand{\detCFCrossLowestCount}{8}
\newcommand{\detCFCrossLowestExceptions}{Gemma 4 12B (Image lowest)}
\newcommand{\detCFCrossLowestNearTieCount}{1}
\newcommand{\detCFCrossLowestNearTieGap}{0.015}
\newcommand{\detCFFlashCross}{0.74}
\newcommand{\detCFFlashImage}{0.81}
\newcommand{\detCFFlashLiteCross}{0.71}
\newcommand{\detCFFlashLiteImage}{0.72}
\newcommand{\detCFFlashLiteText}{0.79}
\newcommand{\detCFFlashText}{0.85}
\newcommand{\detCFGemmaCross}{0.80}
\newcommand{\detCFGemmaImage}{0.75}
\newcommand{\detCFGemmaText}{0.84}
\newcommand{\detCFInternVLCross}{0.76}
\newcommand{\detCFInternVLImage}{0.81}
\newcommand{\detCFInternVLText}{0.84}
\newcommand{\detCFMaxCross}{0.82}
\newcommand{\detCFMaxCrossModel}{Qwen2.5-VL-7B}
\newcommand{\detCFMaxImage}{0.86}
\newcommand{\detCFMaxImageModel}{Mistral-Small-3.2-24B}
\newcommand{\detCFMaxText}{0.89}
\newcommand{\detCFMaxTextModel}{Qwen2.5-VL-7B}
\newcommand{\detCFMeanCross}{0.75}
\newcommand{\detCFMeanImage}{0.78}
\newcommand{\detCFMeanText}{0.85}
\newcommand{\detCFMinCross}{0.65}
\newcommand{\detCFMinCrossModel}{Qwen3.8-27B}
\newcommand{\detCFMinImage}{0.72}
\newcommand{\detCFMinImageModel}{Gemini 3.1 Flash-Lite}
\newcommand{\detCFMinText}{0.79}
\newcommand{\detCFMinTextModel}{Gemini 3.1 Flash-Lite}
\newcommand{\detCFMistralCross}{0.81}
\newcommand{\detCFMistralImage}{0.86}
\newcommand{\detCFMistralText}{0.87}
\newcommand{\detCFModalLSApiCount}{7}
\newcommand{\detCFModalLSOpenCount}{0}
\newcommand{\detCFModalSEApiCount}{1}
\newcommand{\detCFModalSECount}{16}
\newcommand{\detCFModalSEOpenCount}{15}
\newcommand{\detCFOpenCells}{18}
\newcommand{\detCFOpenPool}{9}
\newcommand{\detCFProCross}{0.71}
\newcommand{\detCFProImage}{0.73}
\newcommand{\detCFProText}{0.87}
\newcommand{\detCFQwenThreeEightCross}{0.65}
\newcommand{\detCFQwenThreeEightImage}{0.74}
\newcommand{\detCFQwenThreeEightText}{0.81}
\newcommand{\detCFQwenThreeFiveCross}{0.71}
\newcommand{\detCFQwenThreeFiveImage}{0.82}
\newcommand{\detCFQwenThreeFiveText}{0.85}
\newcommand{\detCFQwenVLCross}{0.82}
\newcommand{\detCFQwenVLImage}{0.84}
\newcommand{\detCFQwenVLText}{0.89}
\newcommand{\detCFSourceSpreadMax}{0.16}
\newcommand{\detCFSourceSpreadMaxModel}{Gemini 3.1 Pro}
\newcommand{\detCFSourceSpreadMin}{0.06}
\newcommand{\detCFSpreadCross}{0.16}
\newcommand{\detCFSpreadImage}{0.14}
\newcommand{\detCFSpreadText}{0.09}
\newcommand{\detCFTextHighestCount}{9}
\newcommand{\detCFTextHighestExceptions}{none}
\newcommand{\detCFminusSEMax}{0.08}
\newcommand{\detCFminusSEMin}{-0.02}
\newcommand{\detCapGemma}{84.2}
\newcommand{\detCapInternVL}{84.8}
\newcommand{\detCapMax}{96.5}
\newcommand{\detCapMin}{84.2}
\newcommand{\detCapMistral}{93.5}
\newcommand{\detCapQwenThreeEight}{95.7}
\newcommand{\detCapQwenThreeFive}{96.5}
\newcommand{\detLenAboveBestCells}{Gemma 4 12B Image, Gemma 4 12B Text, Qwen3.5-9B Cross-modal, Gemini 3.1 Pro Image, Gemini 3.1 Flash-Lite Image}
\newcommand{\detLenAboveBestCount}{5}
\newcommand{\detLenAboveSECount}{8}
\newcommand{\detLenCells}{24}
\newcommand{\detLenFlashCross}{0.72}
\newcommand{\detLenFlashImage}{0.76}
\newcommand{\detLenFlashLiteCross}{0.72}
\newcommand{\detLenFlashLiteImage}{0.84}
\newcommand{\detLenFlashLiteText}{0.80}
\newcommand{\detLenFlashText}{0.71}
\newcommand{\detLenGemmaCross}{0.75}
\newcommand{\detLenGemmaImage}{0.87}
\newcommand{\detLenGemmaText}{0.86}
\newcommand{\detLenInternVLCross}{0.70}
\newcommand{\detLenInternVLImage}{0.50}
\newcommand{\detLenInternVLText}{0.65}
\newcommand{\detLenMinusBestFlashLiteImage}{0.11}
\newcommand{\detLenMinusBestFlashLiteImageHi}{0.17}
\newcommand{\detLenMinusBestFlashLiteImageLo}{0.05}
\newcommand{\detLenMinusSEGemmaImage}{0.10}
\newcommand{\detLenMinusSEGemmaImageHi}{0.16}
\newcommand{\detLenMinusSEGemmaImageLo}{0.04}
\newcommand{\detLenMinusSEProImage}{0.12}
\newcommand{\detLenMinusSEProImageHi}{0.19}
\newcommand{\detLenMinusSEProImageLo}{0.04}
\newcommand{\detLenMistralCross}{0.69}
\newcommand{\detLenMistralImage}{0.65}
\newcommand{\detLenMistralText}{0.67}
\newcommand{\detLenProCross}{0.66}
\newcommand{\detLenProImage}{0.78}
\newcommand{\detLenProText}{0.70}
\newcommand{\detLenQwenThreeEightCross}{0.65}
\newcommand{\detLenQwenThreeEightImage}{0.74}
\newcommand{\detLenQwenThreeEightText}{0.76}
\newcommand{\detLenQwenThreeFiveCross}{0.74}
\newcommand{\detLenQwenThreeFiveImage}{0.71}
\newcommand{\detLenQwenThreeFiveText}{0.76}
\newcommand{\detNApi}{3}
\newcommand{\detNCells}{27}
\newcommand{\detNClean}{150}
\newcommand{\detNModels}{9}
\newcommand{\detNOpen}{6}
\newcommand{\detNPosCross}{122}
\newcommand{\detNPosImage}{150}
\newcommand{\detNPosText}{150}
\newcommand{\detOptimismMax}{0.02}
\newcommand{\detOptimismMedian}{0.01}
\newcommand{\detQwenPipeOtherMax}{0.17}
\newcommand{\detQwenPipeSEMax}{0.04}
\newcommand{\detSEAboveChanceCount}{27}
\newcommand{\detSECILowMin}{0.59}
\newcommand{\detSECrossLowestCount}{8}
\newcommand{\detSECrossLowestExceptions}{Gemma 4 12B (Image lowest)}
\newcommand{\detSECrossLowestNearTieCount}{2}
\newcommand{\detSECrossLowestNearTieGap}{0.015}
\newcommand{\detSEFlashCross}{0.75}
\newcommand{\detSEFlashImage}{0.81}
\newcommand{\detSEFlashLiteCross}{0.71}
\newcommand{\detSEFlashLiteImage}{0.72}
\newcommand{\detSEFlashLiteText}{0.76}
\newcommand{\detSEFlashText}{0.83}
\newcommand{\detSEGemmaCross}{0.81}
\newcommand{\detSEGemmaImage}{0.77}
\newcommand{\detSEGemmaText}{0.84}
\newcommand{\detSEInternVLCross}{0.77}
\newcommand{\detSEInternVLImage}{0.82}
\newcommand{\detSEInternVLText}{0.85}
\newcommand{\detSEMaxCross}{0.81}
\newcommand{\detSEMaxCrossModel}{Mistral-Small-3.2-24B}
\newcommand{\detSEMaxImage}{0.87}
\newcommand{\detSEMaxImageModel}{Mistral-Small-3.2-24B}
\newcommand{\detSEMaxText}{0.89}
\newcommand{\detSEMaxTextModel}{Qwen2.5-VL-7B}
\newcommand{\detSEMeanCross}{0.74}
\newcommand{\detSEMeanImage}{0.79}
\newcommand{\detSEMeanText}{0.84}
\newcommand{\detSEMinCross}{0.66}
\newcommand{\detSEMinCrossModel}{Gemini 3.1 Pro}
\newcommand{\detSEMinImage}{0.67}
\newcommand{\detSEMinImageModel}{Gemini 3.1 Pro}
\newcommand{\detSEMinText}{0.76}
\newcommand{\detSEMinTextModel}{Gemini 3.1 Flash-Lite}
\newcommand{\detSEMistralCross}{0.81}
\newcommand{\detSEMistralImage}{0.87}
\newcommand{\detSEMistralText}{0.87}
\newcommand{\detSEProCross}{0.66}
\newcommand{\detSEProImage}{0.67}
\newcommand{\detSEProText}{0.79}
\newcommand{\detSEQwenThreeEightCross}{0.67}
\newcommand{\detSEQwenThreeEightImage}{0.76}
\newcommand{\detSEQwenThreeEightText}{0.83}
\newcommand{\detSEQwenThreeFiveCross}{0.73}
\newcommand{\detSEQwenThreeFiveImage}{0.82}
\newcommand{\detSEQwenThreeFiveText}{0.86}
\newcommand{\detSEQwenVLCross}{0.80}
\newcommand{\detSEQwenVLImage}{0.85}
\newcommand{\detSEQwenVLText}{0.89}
\newcommand{\detSESourceSpreadMax}{0.16}
\newcommand{\detSESourceSpreadMaxModel}{Qwen3.8-27B}
\newcommand{\detSESourceSpreadMin}{0.05}
\newcommand{\detSESpreadCross}{0.15}
\newcommand{\detSESpreadCrossApi}{0.09}
\newcommand{\detSESpreadCrossOpen}{0.15}
\newcommand{\detSESpreadImage}{0.20}
\newcommand{\detSESpreadImageApi}{0.15}
\newcommand{\detSESpreadImageOpen}{0.12}
\newcommand{\detSESpreadText}{0.13}
\newcommand{\detSESpreadTextApi}{0.07}
\newcommand{\detSESpreadTextOpen}{0.07}
\newcommand{\detSETextHighestCount}{8}
\newcommand{\detSETextHighestExceptions}{Mistral-Small-3.2-24B (Image highest)}
\newcommand{\detSFourAboveChanceCount}{27}
\newcommand{\detSFourFlashCross}{0.69}
\newcommand{\detSFourFlashImage}{0.77}
\newcommand{\detSFourFlashLiteCross}{0.68}
\newcommand{\detSFourFlashLiteImage}{0.71}
\newcommand{\detSFourFlashLiteText}{0.75}
\newcommand{\detSFourFlashText}{0.80}
\newcommand{\detSFourGemmaCross}{0.73}
\newcommand{\detSFourGemmaImage}{0.70}
\newcommand{\detSFourGemmaText}{0.78}
\newcommand{\detSFourInternVLCross}{0.70}
\newcommand{\detSFourInternVLImage}{0.82}
\newcommand{\detSFourInternVLText}{0.79}
\newcommand{\detSFourMaxCross}{0.75}
\newcommand{\detSFourMaxCrossModel}{Mistral-Small-3.2-24B}
\newcommand{\detSFourMaxImage}{0.83}
\newcommand{\detSFourMaxImageModel}{Mistral-Small-3.2-24B}
\newcommand{\detSFourMaxText}{0.83}
\newcommand{\detSFourMaxTextModel}{Mistral-Small-3.2-24B}
\newcommand{\detSFourMeanCross}{0.69}
\newcommand{\detSFourMeanImage}{0.75}
\newcommand{\detSFourMeanText}{0.79}
\newcommand{\detSFourMinCross}{0.65}
\newcommand{\detSFourMinCrossModel}{Gemini 3.1 Pro}
\newcommand{\detSFourMinImage}{0.67}
\newcommand{\detSFourMinImageModel}{Gemini 3.1 Pro}
\newcommand{\detSFourMinText}{0.75}
\newcommand{\detSFourMinTextModel}{Gemini 3.1 Flash-Lite}
\newcommand{\detSFourMistralCross}{0.75}
\newcommand{\detSFourMistralImage}{0.83}
\newcommand{\detSFourMistralText}{0.83}
\newcommand{\detSFourProCross}{0.65}
\newcommand{\detSFourProImage}{0.67}
\newcommand{\detSFourProText}{0.80}
\newcommand{\detSFourQwenThreeEightCross}{0.66}
\newcommand{\detSFourQwenThreeEightImage}{0.73}
\newcommand{\detSFourQwenThreeEightText}{0.80}
\newcommand{\detSFourQwenThreeFiveCross}{0.69}
\newcommand{\detSFourQwenThreeFiveImage}{0.77}
\newcommand{\detSFourQwenThreeFiveText}{0.81}
\newcommand{\detSFourQwenVLCross}{0.67}
\newcommand{\detSFourQwenVLImage}{0.76}
\newcommand{\detSFourQwenVLText}{0.77}
\newcommand{\detSFourSpreadCross}{0.10}
\newcommand{\detSFourSpreadImage}{0.16}
\newcommand{\detSFourSpreadText}{0.07}
\newcommand{\detVarNineLomoModelMax}{28}
\newcommand{\detVarNineLomoModelMin}{15}
\newcommand{\detVarNineLomoSourceMax}{65}
\newcommand{\detVarNineLomoSourceMin}{53}
\newcommand{\detVarNineModel}{25}
\newcommand{\detVarNineModelHi}{47}
\newcommand{\detVarNineModelLo}{14}
\newcommand{\detVarNineModelP}{0.04}
\newcommand{\detVarNineQwenRerunModel}{29}
\newcommand{\detVarNineQwenRerunModelP}{0.02}
\newcommand{\detVarNineQwenRerunResidual}{17}
\newcommand{\detVarNineQwenRerunSource}{54}
\newcommand{\detVarNineQwenRerunSourceP}{$<$0.001}
\newcommand{\detVarNineResidual}{18}
\newcommand{\detVarNineResidualHi}{31}
\newcommand{\detVarNineResidualLo}{12}
\newcommand{\detVarNineSEModel}{55}
\newcommand{\detVarNineSEModelHi}{71}
\newcommand{\detVarNineSEModelLo}{36}
\newcommand{\detVarNineSEModelP}{$<$0.001}
\newcommand{\detVarNineSEResidual}{12}
\newcommand{\detVarNineSEResidualHi}{23}
\newcommand{\detVarNineSEResidualLo}{8}
\newcommand{\detVarNineSESource}{33}
\newcommand{\detVarNineSESourceHi}{49}
\newcommand{\detVarNineSESourceLo}{15}
\newcommand{\detVarNineSESourceP}{$<$0.001}
\newcommand{\detVarNineSource}{57}
\newcommand{\detVarNineSourceHi}{68}
\newcommand{\detVarNineSourceLo}{32}
\newcommand{\detVarNineSourceP}{$<$0.001}
\newcommand{\detVarSixModel}{27}
\newcommand{\detVarSixModelHi}{49}
\newcommand{\detVarSixModelLo}{13}
\newcommand{\detVarSixModelP}{0.11}
\newcommand{\detVarSixResidual}{23}
\newcommand{\detVarSixResidualHi}{38}
\newcommand{\detVarSixResidualLo}{13}
\newcommand{\detVarSixSource}{50}
\newcommand{\detVarSixSourceHi}{66}
\newcommand{\detVarSixSourceLo}{27}
\newcommand{\detVarSixSourceP}{0.003}

\newcommand{\cTwoNImage}{8}
\newcommand{\cTwoNTextKept}{2}
\newcommand{\cTwoHCCcells}{1{,}902}
\newcommand{\cTwoHCCcellsPos}{693}
\newcommand{\cTwoHCCcellsNeg}{115}
\newcommand{\cTwoHCCcellsUndef}{18}
\newcommand{\cTwoHCCcellsLowN}{597}
\newcommand{\cTwoHCCcommittedCells}{1{,}837}
\newcommand{\cTwoHCCcommittedPos}{522}

\newcommand{\cTwoFlashMaxTokAnswers}{1{,}841}
\newcommand{\cTwoFlashPairs}{4{,}620}
\newcommand{\cTwoFlashMaxTokH}{775}
\newcommand{\cTwoFlashH}{969}

\newcommand{\cTwoFlashMaxTokHcommitted}{370}
\newcommand{\cTwoFlashCapHdiffers}{87}
\newcommand{\cTwoFlashCapH}{969}
\newcommand{\cTwoFlashFullH}{956}
\newcommand{\cTwoProCapHdiffers}{188}
\newcommand{\cTwoProCapH}{857}
\newcommand{\cTwoProFullH}{779}
\newcommand{\cTwoLiteCapHdiffers}{131}
\newcommand{\cTwoLiteCapH}{796}
\newcommand{\cTwoLiteFullH}{717}

\providecommand{\amb}[1]{\csname amb@#1\endcsname}
\expandafter\def\csname amb@npos\endcsname{60}
\expandafter\def\csname amb@nneg\endcsname{80}
\expandafter\def\csname amb@allamb\endcsname{0.60}
\expandafter\def\csname amb@budget-open-label\endcsname{8}
\expandafter\def\csname amb@budget-open-free\endcsname{512}
\expandafter\def\csname amb@cap-share-min\endcsname{61}
\expandafter\def\csname amb@cap-share-max\endcsname{100}
\expandafter\def\csname amb@ntot\endcsname{140}
\expandafter\def\csname amb@uqbest-min\endcsname{0.57}
\expandafter\def\csname amb@uqbest-max\endcsname{0.64}
\expandafter\def\csname amb@n-models\endcsname{9}
\expandafter\def\csname amb@n-uq-above-allamb\endcsname{0}
\expandafter\def\csname amb@qwen25-uqbest\endcsname{0.64}
\expandafter\def\csname amb@qwen25-uqbest3\endcsname{0.640}
\expandafter\def\csname amb@qwen25-uqlo\endcsname{0.57}
\expandafter\def\csname amb@qwen25-uqhi\endcsname{0.73}
\expandafter\def\csname amb@qwen25-uqtop\endcsname{MTE}
\expandafter\def\csname amb@qwen25-uqminusall\endcsname{+0.05}
\expandafter\def\csname amb@qwen25-uqminusall-lo\endcsname{-0.00}
\expandafter\def\csname amb@qwen25-uqminusall-hi\endcsname{+0.11}
\expandafter\def\csname amb@qwen25-zs\endcsname{0.63}
\expandafter\def\csname amb@qwen25-zs3\endcsname{0.628}
\expandafter\def\csname amb@qwen25-zs-flagged\endcsname{128}
\expandafter\def\csname amb@qwen25-zs-limited\endcsname{0}
\expandafter\def\csname amb@qwen25-zs-minusuq\endcsname{-0.03}
\expandafter\def\csname amb@qwen25-zs-minusuq-lo\endcsname{-0.09}
\expandafter\def\csname amb@qwen25-zs-minusuq-hi\endcsname{+0.03}
\expandafter\def\csname amb@qwen25-cot\endcsname{0.62}
\expandafter\def\csname amb@qwen25-cot3\endcsname{0.625}
\expandafter\def\csname amb@qwen25-cot-flagged\endcsname{132}
\expandafter\def\csname amb@qwen25-cot-limited\endcsname{0}
\expandafter\def\csname amb@qwen25-cot-minusuq\endcsname{-0.03}
\expandafter\def\csname amb@qwen25-cot-minusuq-lo\endcsname{-0.09}
\expandafter\def\csname amb@qwen25-cot-minusuq-hi\endcsname{+0.03}
\expandafter\def\csname amb@qwen25-ecot\endcsname{0.62}
\expandafter\def\csname amb@qwen25-ecot3\endcsname{0.616}
\expandafter\def\csname amb@qwen25-ecot-flagged\endcsname{125}
\expandafter\def\csname amb@qwen25-ecot-limited\endcsname{0}
\expandafter\def\csname amb@qwen25-ecot-minusuq\endcsname{-0.04}
\expandafter\def\csname amb@qwen25-ecot-minusuq-lo\endcsname{-0.10}
\expandafter\def\csname amb@qwen25-ecot-minusuq-hi\endcsname{+0.02}
\expandafter\def\csname amb@qwen25-cotfree\endcsname{0.57}
\expandafter\def\csname amb@qwen25-cotfree3\endcsname{0.569}
\expandafter\def\csname amb@qwen25-cotfree-flagged\endcsname{77}
\expandafter\def\csname amb@qwen25-cotfree-limited\endcsname{0}
\expandafter\def\csname amb@qwen25-cotfree-minusuq\endcsname{-0.08}
\expandafter\def\csname amb@qwen25-cotfree-minusuq-lo\endcsname{-0.18}
\expandafter\def\csname amb@qwen25-cotfree-minusuq-hi\endcsname{+0.01}
\expandafter\def\csname amb@qwen25-ecotfree\endcsname{0.60}
\expandafter\def\csname amb@qwen25-ecotfree3\endcsname{0.600}
\expandafter\def\csname amb@qwen25-ecotfree-flagged\endcsname{60}
\expandafter\def\csname amb@qwen25-ecotfree-limited\endcsname{0}
\expandafter\def\csname amb@qwen25-ecotfree-minusuq\endcsname{-0.05}
\expandafter\def\csname amb@qwen25-ecotfree-minusuq-lo\endcsname{-0.16}
\expandafter\def\csname amb@qwen25-ecotfree-minusuq-hi\endcsname{+0.05}
\expandafter\def\csname amb@gemma-uqbest\endcsname{0.61}
\expandafter\def\csname amb@gemma-uqbest3\endcsname{0.613}
\expandafter\def\csname amb@gemma-uqlo\endcsname{0.49}
\expandafter\def\csname amb@gemma-uqhi\endcsname{0.68}
\expandafter\def\csname amb@gemma-uqtop\endcsname{LUQ}
\expandafter\def\csname amb@gemma-uqminusall\endcsname{-0.01}
\expandafter\def\csname amb@gemma-uqminusall-lo\endcsname{-0.09}
\expandafter\def\csname amb@gemma-uqminusall-hi\endcsname{+0.06}
\expandafter\def\csname amb@gemma-zs\endcsname{0.74}
\expandafter\def\csname amb@gemma-zs3\endcsname{0.736}
\expandafter\def\csname amb@gemma-zs-flagged\endcsname{65}
\expandafter\def\csname amb@gemma-zs-limited\endcsname{0}
\expandafter\def\csname amb@gemma-zs-minusuq\endcsname{+0.14}
\expandafter\def\csname amb@gemma-zs-minusuq-lo\endcsname{+0.05}
\expandafter\def\csname amb@gemma-zs-minusuq-hi\endcsname{+0.25}
\expandafter\def\csname amb@gemma-cot\endcsname{0.70}
\expandafter\def\csname amb@gemma-cot3\endcsname{0.699}
\expandafter\def\csname amb@gemma-cot-flagged\endcsname{86}
\expandafter\def\csname amb@gemma-cot-limited\endcsname{6}
\expandafter\def\csname amb@gemma-cot-minusuq\endcsname{+0.11}
\expandafter\def\csname amb@gemma-cot-minusuq-lo\endcsname{+0.02}
\expandafter\def\csname amb@gemma-cot-minusuq-hi\endcsname{+0.20}
\expandafter\def\csname amb@gemma-ecot\endcsname{0.67}
\expandafter\def\csname amb@gemma-ecot3\endcsname{0.673}
\expandafter\def\csname amb@gemma-ecot-flagged\endcsname{50}
\expandafter\def\csname amb@gemma-ecot-limited\endcsname{1}
\expandafter\def\csname amb@gemma-ecot-minusuq\endcsname{+0.08}
\expandafter\def\csname amb@gemma-ecot-minusuq-lo\endcsname{-0.04}
\expandafter\def\csname amb@gemma-ecot-minusuq-hi\endcsname{+0.21}
\expandafter\def\csname amb@gemma-cotfree\endcsname{0.76}
\expandafter\def\csname amb@gemma-cotfree3\endcsname{0.756}
\expandafter\def\csname amb@gemma-cotfree-flagged\endcsname{67}
\expandafter\def\csname amb@gemma-cotfree-limited\endcsname{0}
\expandafter\def\csname amb@gemma-cotfree-minusuq\endcsname{+0.16}
\expandafter\def\csname amb@gemma-cotfree-minusuq-lo\endcsname{+0.07}
\expandafter\def\csname amb@gemma-cotfree-minusuq-hi\endcsname{+0.26}
\expandafter\def\csname amb@gemma-ecotfree\endcsname{0.79}
\expandafter\def\csname amb@gemma-ecotfree3\endcsname{0.790}
\expandafter\def\csname amb@gemma-ecotfree-flagged\endcsname{59}
\expandafter\def\csname amb@gemma-ecotfree-limited\endcsname{0}
\expandafter\def\csname amb@gemma-ecotfree-minusuq\endcsname{+0.20}
\expandafter\def\csname amb@gemma-ecotfree-minusuq-lo\endcsname{+0.09}
\expandafter\def\csname amb@gemma-ecotfree-minusuq-hi\endcsname{+0.30}
\expandafter\def\csname amb@flash-uqbest\endcsname{0.57}
\expandafter\def\csname amb@flash-uqbest3\endcsname{0.569}
\expandafter\def\csname amb@flash-uqlo\endcsname{0.48}
\expandafter\def\csname amb@flash-uqhi\endcsname{0.66}
\expandafter\def\csname amb@flash-uqtop\endcsname{LexSim}
\expandafter\def\csname amb@flash-uqminusall\endcsname{-0.02}
\expandafter\def\csname amb@flash-uqminusall-lo\endcsname{-0.08}
\expandafter\def\csname amb@flash-uqminusall-hi\endcsname{+0.03}
\expandafter\def\csname amb@flash-zs\endcsname{0.84}
\expandafter\def\csname amb@flash-zs3\endcsname{0.837}
\expandafter\def\csname amb@flash-zs-flagged\endcsname{69}
\expandafter\def\csname amb@flash-zs-limited\endcsname{0}
\expandafter\def\csname amb@flash-zs-minusuq\endcsname{+0.26}
\expandafter\def\csname amb@flash-zs-minusuq-lo\endcsname{+0.18}
\expandafter\def\csname amb@flash-zs-minusuq-hi\endcsname{+0.35}
\expandafter\def\csname amb@flash-cot\endcsname{0.81}
\expandafter\def\csname amb@flash-cot3\endcsname{0.806}
\expandafter\def\csname amb@flash-cot-flagged\endcsname{69}
\expandafter\def\csname amb@flash-cot-limited\endcsname{8}
\expandafter\def\csname amb@flash-cot-minusuq\endcsname{+0.23}
\expandafter\def\csname amb@flash-cot-minusuq-lo\endcsname{+0.14}
\expandafter\def\csname amb@flash-cot-minusuq-hi\endcsname{+0.32}
\expandafter\def\csname amb@flash-ecot\endcsname{0.82}
\expandafter\def\csname amb@flash-ecot3\endcsname{0.815}
\expandafter\def\csname amb@flash-ecot-flagged\endcsname{70}
\expandafter\def\csname amb@flash-ecot-limited\endcsname{1}
\expandafter\def\csname amb@flash-ecot-minusuq\endcsname{+0.24}
\expandafter\def\csname amb@flash-ecot-minusuq-lo\endcsname{+0.15}
\expandafter\def\csname amb@flash-ecot-minusuq-hi\endcsname{+0.33}
\expandafter\def\csname amb@flash-cotfree\endcsname{0.82}
\expandafter\def\csname amb@flash-cotfree3\endcsname{0.815}
\expandafter\def\csname amb@flash-cotfree-flagged\endcsname{70}
\expandafter\def\csname amb@flash-cotfree-limited\endcsname{0}
\expandafter\def\csname amb@flash-cotfree-minusuq\endcsname{+0.24}
\expandafter\def\csname amb@flash-cotfree-minusuq-lo\endcsname{+0.15}
\expandafter\def\csname amb@flash-cotfree-minusuq-hi\endcsname{+0.33}
\expandafter\def\csname amb@flash-ecotfree\endcsname{0.84}
\expandafter\def\csname amb@flash-ecotfree3\endcsname{0.841}
\expandafter\def\csname amb@flash-ecotfree-flagged\endcsname{66}
\expandafter\def\csname amb@flash-ecotfree-limited\endcsname{0}
\expandafter\def\csname amb@flash-ecotfree-minusuq\endcsname{+0.26}
\expandafter\def\csname amb@flash-ecotfree-minusuq-lo\endcsname{+0.17}
\expandafter\def\csname amb@flash-ecotfree-minusuq-hi\endcsname{+0.36}
\expandafter\def\csname amb@pro-uqbest\endcsname{0.59}
\expandafter\def\csname amb@pro-uqbest3\endcsname{0.585}
\expandafter\def\csname amb@pro-uqlo\endcsname{0.50}
\expandafter\def\csname amb@pro-uqhi\endcsname{0.68}
\expandafter\def\csname amb@pro-uqtop\endcsname{LexSim}
\expandafter\def\csname amb@pro-uqminusall\endcsname{-0.00}
\expandafter\def\csname amb@pro-uqminusall-lo\endcsname{-0.08}
\expandafter\def\csname amb@pro-uqminusall-hi\endcsname{+0.07}
\expandafter\def\csname amb@pro-zs\endcsname{0.80}
\expandafter\def\csname amb@pro-zs3\endcsname{0.800}
\expandafter\def\csname amb@pro-zs-flagged\endcsname{75}
\expandafter\def\csname amb@pro-zs-limited\endcsname{0}
\expandafter\def\csname amb@pro-zs-minusuq\endcsname{+0.20}
\expandafter\def\csname amb@pro-zs-minusuq-lo\endcsname{+0.12}
\expandafter\def\csname amb@pro-zs-minusuq-hi\endcsname{+0.30}
\expandafter\def\csname amb@pro-cot\endcsname{0.82}
\expandafter\def\csname amb@pro-cot3\endcsname{0.818}
\expandafter\def\csname amb@pro-cot-flagged\endcsname{72}
\expandafter\def\csname amb@pro-cot-limited\endcsname{0}
\expandafter\def\csname amb@pro-cot-minusuq\endcsname{+0.22}
\expandafter\def\csname amb@pro-cot-minusuq-lo\endcsname{+0.13}
\expandafter\def\csname amb@pro-cot-minusuq-hi\endcsname{+0.33}
\expandafter\def\csname amb@pro-ecot\endcsname{0.83}
\expandafter\def\csname amb@pro-ecot3\endcsname{0.833}
\expandafter\def\csname amb@pro-ecot-flagged\endcsname{72}
\expandafter\def\csname amb@pro-ecot-limited\endcsname{0}
\expandafter\def\csname amb@pro-ecot-minusuq\endcsname{+0.24}
\expandafter\def\csname amb@pro-ecot-minusuq-lo\endcsname{+0.14}
\expandafter\def\csname amb@pro-ecot-minusuq-hi\endcsname{+0.34}
\expandafter\def\csname amb@pro-cotfree\endcsname{0.78}
\expandafter\def\csname amb@pro-cotfree3\endcsname{0.779}
\expandafter\def\csname amb@pro-cotfree-flagged\endcsname{76}
\expandafter\def\csname amb@pro-cotfree-limited\endcsname{0}
\expandafter\def\csname amb@pro-cotfree-minusuq\endcsname{+0.18}
\expandafter\def\csname amb@pro-cotfree-minusuq-lo\endcsname{+0.09}
\expandafter\def\csname amb@pro-cotfree-minusuq-hi\endcsname{+0.28}
\expandafter\def\csname amb@pro-ecotfree\endcsname{0.81}
\expandafter\def\csname amb@pro-ecotfree3\endcsname{0.812}
\expandafter\def\csname amb@pro-ecotfree-flagged\endcsname{73}
\expandafter\def\csname amb@pro-ecotfree-limited\endcsname{0}
\expandafter\def\csname amb@pro-ecotfree-minusuq\endcsname{+0.22}
\expandafter\def\csname amb@pro-ecotfree-minusuq-lo\endcsname{+0.12}
\expandafter\def\csname amb@pro-ecotfree-minusuq-hi\endcsname{+0.31}
\expandafter\def\csname amb@internvl-uqbest\endcsname{0.62}
\expandafter\def\csname amb@internvl-uqbest3\endcsname{0.624}
\expandafter\def\csname amb@internvl-uqlo\endcsname{0.52}
\expandafter\def\csname amb@internvl-uqhi\endcsname{0.70}
\expandafter\def\csname amb@internvl-uqtop\endcsname{PPL}
\expandafter\def\csname amb@internvl-uqminusall\endcsname{+0.02}
\expandafter\def\csname amb@internvl-uqminusall-lo\endcsname{-0.08}
\expandafter\def\csname amb@internvl-uqminusall-hi\endcsname{+0.10}
\expandafter\def\csname amb@internvl-zs\endcsname{0.19}
\expandafter\def\csname amb@internvl-zs3\endcsname{0.192}
\expandafter\def\csname amb@internvl-zs-flagged\endcsname{13}
\expandafter\def\csname amb@internvl-zs-limited\endcsname{0}
\expandafter\def\csname amb@internvl-zs-minusuq\endcsname{-0.42}
\expandafter\def\csname amb@internvl-zs-minusuq-lo\endcsname{-0.58}
\expandafter\def\csname amb@internvl-zs-minusuq-hi\endcsname{-0.26}
\expandafter\def\csname amb@internvl-cot\endcsname{0.35}
\expandafter\def\csname amb@internvl-cot3\endcsname{0.352}
\expandafter\def\csname amb@internvl-cot-flagged\endcsname{31}
\expandafter\def\csname amb@internvl-cot-limited\endcsname{0}
\expandafter\def\csname amb@internvl-cot-minusuq\endcsname{-0.26}
\expandafter\def\csname amb@internvl-cot-minusuq-lo\endcsname{-0.42}
\expandafter\def\csname amb@internvl-cot-minusuq-hi\endcsname{-0.10}
\expandafter\def\csname amb@internvl-ecot\endcsname{0.54}
\expandafter\def\csname amb@internvl-ecot3\endcsname{0.542}
\expandafter\def\csname amb@internvl-ecot-flagged\endcsname{84}
\expandafter\def\csname amb@internvl-ecot-limited\endcsname{16}
\expandafter\def\csname amb@internvl-ecot-minusuq\endcsname{-0.07}
\expandafter\def\csname amb@internvl-ecot-minusuq-lo\endcsname{-0.19}
\expandafter\def\csname amb@internvl-ecot-minusuq-hi\endcsname{+0.03}
\expandafter\def\csname amb@internvl-cotfree\endcsname{0.64}
\expandafter\def\csname amb@internvl-cotfree3\endcsname{0.645}
\expandafter\def\csname amb@internvl-cotfree-flagged\endcsname{61}
\expandafter\def\csname amb@internvl-cotfree-limited\endcsname{0}
\expandafter\def\csname amb@internvl-cotfree-minusuq\endcsname{+0.03}
\expandafter\def\csname amb@internvl-cotfree-minusuq-lo\endcsname{-0.09}
\expandafter\def\csname amb@internvl-cotfree-minusuq-hi\endcsname{+0.14}
\expandafter\def\csname amb@internvl-ecotfree\endcsname{0.67}
\expandafter\def\csname amb@internvl-ecotfree3\endcsname{0.667}
\expandafter\def\csname amb@internvl-ecotfree-flagged\endcsname{45}
\expandafter\def\csname amb@internvl-ecotfree-limited\endcsname{0}
\expandafter\def\csname amb@internvl-ecotfree-minusuq\endcsname{+0.05}
\expandafter\def\csname amb@internvl-ecotfree-minusuq-lo\endcsname{-0.07}
\expandafter\def\csname amb@internvl-ecotfree-minusuq-hi\endcsname{+0.17}
\expandafter\def\csname amb@mistral-uqbest\endcsname{0.61}
\expandafter\def\csname amb@mistral-uqbest3\endcsname{0.612}
\expandafter\def\csname amb@mistral-uqlo\endcsname{0.52}
\expandafter\def\csname amb@mistral-uqhi\endcsname{0.70}
\expandafter\def\csname amb@mistral-uqtop\endcsname{SE}
\expandafter\def\csname amb@mistral-uqminusall\endcsname{+0.01}
\expandafter\def\csname amb@mistral-uqminusall-lo\endcsname{-0.05}
\expandafter\def\csname amb@mistral-uqminusall-hi\endcsname{+0.07}
\expandafter\def\csname amb@mistral-zs\endcsname{0.71}
\expandafter\def\csname amb@mistral-zs3\endcsname{0.706}
\expandafter\def\csname amb@mistral-zs-flagged\endcsname{59}
\expandafter\def\csname amb@mistral-zs-limited\endcsname{0}
\expandafter\def\csname amb@mistral-zs-minusuq\endcsname{+0.09}
\expandafter\def\csname amb@mistral-zs-minusuq-lo\endcsname{-0.02}
\expandafter\def\csname amb@mistral-zs-minusuq-hi\endcsname{+0.20}
\expandafter\def\csname amb@mistral-cot\endcsname{0.69}
\expandafter\def\csname amb@mistral-cot3\endcsname{0.689}
\expandafter\def\csname amb@mistral-cot-flagged\endcsname{62}
\expandafter\def\csname amb@mistral-cot-limited\endcsname{0}
\expandafter\def\csname amb@mistral-cot-minusuq\endcsname{+0.08}
\expandafter\def\csname amb@mistral-cot-minusuq-lo\endcsname{-0.03}
\expandafter\def\csname amb@mistral-cot-minusuq-hi\endcsname{+0.18}
\expandafter\def\csname amb@mistral-ecot\endcsname{0.72}
\expandafter\def\csname amb@mistral-ecot3\endcsname{0.719}
\expandafter\def\csname amb@mistral-ecot-flagged\endcsname{54}
\expandafter\def\csname amb@mistral-ecot-limited\endcsname{0}
\expandafter\def\csname amb@mistral-ecot-minusuq\endcsname{+0.11}
\expandafter\def\csname amb@mistral-ecot-minusuq-lo\endcsname{-0.01}
\expandafter\def\csname amb@mistral-ecot-minusuq-hi\endcsname{+0.22}
\expandafter\def\csname amb@mistral-cotfree\endcsname{0.63}
\expandafter\def\csname amb@mistral-cotfree3\endcsname{0.626}
\expandafter\def\csname amb@mistral-cotfree-flagged\endcsname{55}
\expandafter\def\csname amb@mistral-cotfree-limited\endcsname{0}
\expandafter\def\csname amb@mistral-cotfree-minusuq\endcsname{+0.01}
\expandafter\def\csname amb@mistral-cotfree-minusuq-lo\endcsname{-0.10}
\expandafter\def\csname amb@mistral-cotfree-minusuq-hi\endcsname{+0.13}
\expandafter\def\csname amb@mistral-ecotfree\endcsname{0.62}
\expandafter\def\csname amb@mistral-ecotfree3\endcsname{0.620}
\expandafter\def\csname amb@mistral-ecotfree-flagged\endcsname{40}
\expandafter\def\csname amb@mistral-ecotfree-limited\endcsname{0}
\expandafter\def\csname amb@mistral-ecotfree-minusuq\endcsname{+0.01}
\expandafter\def\csname amb@mistral-ecotfree-minusuq-lo\endcsname{-0.13}
\expandafter\def\csname amb@mistral-ecotfree-minusuq-hi\endcsname{+0.14}
\expandafter\def\csname amb@qwen35-uqbest\endcsname{0.58}
\expandafter\def\csname amb@qwen35-uqbest3\endcsname{0.585}
\expandafter\def\csname amb@qwen35-uqlo\endcsname{0.50}
\expandafter\def\csname amb@qwen35-uqhi\endcsname{0.69}
\expandafter\def\csname amb@qwen35-uqtop\endcsname{LexSim}
\expandafter\def\csname amb@qwen35-uqminusall\endcsname{+0.00}
\expandafter\def\csname amb@qwen35-uqminusall-lo\endcsname{-0.07}
\expandafter\def\csname amb@qwen35-uqminusall-hi\endcsname{+0.07}
\expandafter\def\csname amb@qwen35-zs\endcsname{0.14}
\expandafter\def\csname amb@qwen35-zs3\endcsname{0.139}
\expandafter\def\csname amb@qwen35-zs-flagged\endcsname{12}
\expandafter\def\csname amb@qwen35-zs-limited\endcsname{4}
\expandafter\def\csname amb@qwen35-zs-minusuq\endcsname{-0.46}
\expandafter\def\csname amb@qwen35-zs-minusuq-lo\endcsname{-0.59}
\expandafter\def\csname amb@qwen35-zs-minusuq-hi\endcsname{-0.33}
\expandafter\def\csname amb@qwen35-cot\endcsname{n/a}
\expandafter\def\csname amb@qwen35-cot3\endcsname{n/a}
\expandafter\def\csname amb@qwen35-cot-flagged\endcsname{136}
\expandafter\def\csname amb@qwen35-cot-limited\endcsname{140}
\expandafter\def\csname amb@qwen35-cot-minusuq\endcsname{n/a}
\expandafter\def\csname amb@qwen35-cot-minusuq-lo\endcsname{n/a}
\expandafter\def\csname amb@qwen35-cot-minusuq-hi\endcsname{n/a}
\expandafter\def\csname amb@qwen35-ecot\endcsname{n/a}
\expandafter\def\csname amb@qwen35-ecot3\endcsname{n/a}
\expandafter\def\csname amb@qwen35-ecot-flagged\endcsname{138}
\expandafter\def\csname amb@qwen35-ecot-limited\endcsname{139}
\expandafter\def\csname amb@qwen35-ecot-minusuq\endcsname{n/a}
\expandafter\def\csname amb@qwen35-ecot-minusuq-lo\endcsname{n/a}
\expandafter\def\csname amb@qwen35-ecot-minusuq-hi\endcsname{n/a}
\expandafter\def\csname amb@qwen35-cotfree\endcsname{0.55}
\expandafter\def\csname amb@qwen35-cotfree3\endcsname{0.547}
\expandafter\def\csname amb@qwen35-cotfree-flagged\endcsname{68}
\expandafter\def\csname amb@qwen35-cotfree-limited\endcsname{16}
\expandafter\def\csname amb@qwen35-cotfree-minusuq\endcsname{-0.05}
\expandafter\def\csname amb@qwen35-cotfree-minusuq-lo\endcsname{-0.17}
\expandafter\def\csname amb@qwen35-cotfree-minusuq-hi\endcsname{+0.06}
\expandafter\def\csname amb@qwen35-ecotfree\endcsname{0.48}
\expandafter\def\csname amb@qwen35-ecotfree3\endcsname{0.481}
\expandafter\def\csname amb@qwen35-ecotfree-flagged\endcsname{44}
\expandafter\def\csname amb@qwen35-ecotfree-limited\endcsname{15}
\expandafter\def\csname amb@qwen35-ecotfree-minusuq\endcsname{-0.12}
\expandafter\def\csname amb@qwen35-ecotfree-minusuq-lo\endcsname{-0.26}
\expandafter\def\csname amb@qwen35-ecotfree-minusuq-hi\endcsname{+0.03}
\expandafter\def\csname amb@qwen38-uqbest\endcsname{0.61}
\expandafter\def\csname amb@qwen38-uqbest3\endcsname{0.605}
\expandafter\def\csname amb@qwen38-uqlo\endcsname{0.53}
\expandafter\def\csname amb@qwen38-uqhi\endcsname{0.70}
\expandafter\def\csname amb@qwen38-uqtop\endcsname{PMI}
\expandafter\def\csname amb@qwen38-uqminusall\endcsname{+0.02}
\expandafter\def\csname amb@qwen38-uqminusall-lo\endcsname{-0.04}
\expandafter\def\csname amb@qwen38-uqminusall-hi\endcsname{+0.08}
\expandafter\def\csname amb@qwen38-zs\endcsname{0.48}
\expandafter\def\csname amb@qwen38-zs3\endcsname{0.477}
\expandafter\def\csname amb@qwen38-zs-flagged\endcsname{28}
\expandafter\def\csname amb@qwen38-zs-limited\endcsname{1}
\expandafter\def\csname amb@qwen38-zs-minusuq\endcsname{-0.14}
\expandafter\def\csname amb@qwen38-zs-minusuq-lo\endcsname{-0.28}
\expandafter\def\csname amb@qwen38-zs-minusuq-hi\endcsname{-0.00}
\expandafter\def\csname amb@qwen38-cot\endcsname{0.20}
\expandafter\def\csname amb@qwen38-cot3\endcsname{0.200}
\expandafter\def\csname amb@qwen38-cot-flagged\endcsname{10}
\expandafter\def\csname amb@qwen38-cot-limited\endcsname{0}
\expandafter\def\csname amb@qwen38-cot-minusuq\endcsname{-0.42}
\expandafter\def\csname amb@qwen38-cot-minusuq-lo\endcsname{-0.56}
\expandafter\def\csname amb@qwen38-cot-minusuq-hi\endcsname{-0.27}
\expandafter\def\csname amb@qwen38-ecot\endcsname{0.06}
\expandafter\def\csname amb@qwen38-ecot3\endcsname{0.065}
\expandafter\def\csname amb@qwen38-ecot-flagged\endcsname{2}
\expandafter\def\csname amb@qwen38-ecot-limited\endcsname{0}
\expandafter\def\csname amb@qwen38-ecot-minusuq\endcsname{-0.55}
\expandafter\def\csname amb@qwen38-ecot-minusuq-lo\endcsname{-0.66}
\expandafter\def\csname amb@qwen38-ecot-minusuq-hi\endcsname{-0.42}
\expandafter\def\csname amb@qwen38-cotfree\endcsname{0.69}
\expandafter\def\csname amb@qwen38-cotfree3\endcsname{0.692}
\expandafter\def\csname amb@qwen38-cotfree-flagged\endcsname{70}
\expandafter\def\csname amb@qwen38-cotfree-limited\endcsname{4}
\expandafter\def\csname amb@qwen38-cotfree-minusuq\endcsname{+0.08}
\expandafter\def\csname amb@qwen38-cotfree-minusuq-lo\endcsname{-0.02}
\expandafter\def\csname amb@qwen38-cotfree-minusuq-hi\endcsname{+0.18}
\expandafter\def\csname amb@qwen38-ecotfree\endcsname{0.78}
\expandafter\def\csname amb@qwen38-ecotfree3\endcsname{0.783}
\expandafter\def\csname amb@qwen38-ecotfree-flagged\endcsname{55}
\expandafter\def\csname amb@qwen38-ecotfree-limited\endcsname{5}
\expandafter\def\csname amb@qwen38-ecotfree-minusuq\endcsname{+0.17}
\expandafter\def\csname amb@qwen38-ecotfree-minusuq-lo\endcsname{+0.06}
\expandafter\def\csname amb@qwen38-ecotfree-minusuq-hi\endcsname{+0.28}
\expandafter\def\csname amb@flashlite-uqbest\endcsname{0.59}
\expandafter\def\csname amb@flashlite-uqbest3\endcsname{0.594}
\expandafter\def\csname amb@flashlite-uqlo\endcsname{0.50}
\expandafter\def\csname amb@flashlite-uqhi\endcsname{0.69}
\expandafter\def\csname amb@flashlite-uqtop\endcsname{DegMat}
\expandafter\def\csname amb@flashlite-uqminusall\endcsname{+0.00}
\expandafter\def\csname amb@flashlite-uqminusall-lo\endcsname{-0.07}
\expandafter\def\csname amb@flashlite-uqminusall-hi\endcsname{+0.08}
\expandafter\def\csname amb@flashlite-zs\endcsname{0.84}
\expandafter\def\csname amb@flashlite-zs3\endcsname{0.839}
\expandafter\def\csname amb@flashlite-zs-flagged\endcsname{64}
\expandafter\def\csname amb@flashlite-zs-limited\endcsname{0}
\expandafter\def\csname amb@flashlite-zs-minusuq\endcsname{+0.24}
\expandafter\def\csname amb@flashlite-zs-minusuq-lo\endcsname{+0.14}
\expandafter\def\csname amb@flashlite-zs-minusuq-hi\endcsname{+0.34}
\expandafter\def\csname amb@flashlite-cot\endcsname{0.84}
\expandafter\def\csname amb@flashlite-cot3\endcsname{0.839}
\expandafter\def\csname amb@flashlite-cot-flagged\endcsname{64}
\expandafter\def\csname amb@flashlite-cot-limited\endcsname{0}
\expandafter\def\csname amb@flashlite-cot-minusuq\endcsname{+0.24}
\expandafter\def\csname amb@flashlite-cot-minusuq-lo\endcsname{+0.14}
\expandafter\def\csname amb@flashlite-cot-minusuq-hi\endcsname{+0.34}
\expandafter\def\csname amb@flashlite-ecot\endcsname{0.75}
\expandafter\def\csname amb@flashlite-ecot3\endcsname{0.754}
\expandafter\def\csname amb@flashlite-ecot-flagged\endcsname{54}
\expandafter\def\csname amb@flashlite-ecot-limited\endcsname{0}
\expandafter\def\csname amb@flashlite-ecot-minusuq\endcsname{+0.15}
\expandafter\def\csname amb@flashlite-ecot-minusuq-lo\endcsname{+0.04}
\expandafter\def\csname amb@flashlite-ecot-minusuq-hi\endcsname{+0.28}
\expandafter\def\csname amb@flashlite-cotfree\endcsname{0.78}
\expandafter\def\csname amb@flashlite-cotfree3\endcsname{0.783}
\expandafter\def\csname amb@flashlite-cotfree-flagged\endcsname{60}
\expandafter\def\csname amb@flashlite-cotfree-limited\endcsname{0}
\expandafter\def\csname amb@flashlite-cotfree-minusuq\endcsname{+0.18}
\expandafter\def\csname amb@flashlite-cotfree-minusuq-lo\endcsname{+0.08}
\expandafter\def\csname amb@flashlite-cotfree-minusuq-hi\endcsname{+0.29}
\expandafter\def\csname amb@flashlite-ecotfree\endcsname{0.80}
\expandafter\def\csname amb@flashlite-ecotfree3\endcsname{0.800}
\expandafter\def\csname amb@flashlite-ecotfree-flagged\endcsname{55}
\expandafter\def\csname amb@flashlite-ecotfree-limited\endcsname{0}
\expandafter\def\csname amb@flashlite-ecotfree-minusuq\endcsname{+0.20}
\expandafter\def\csname amb@flashlite-ecotfree-minusuq-lo\endcsname{+0.09}
\expandafter\def\csname amb@flashlite-ecotfree-minusuq-hi\endcsname{+0.31}
\expandafter\def\csname amb@n-free-level-both-free-above-allamb\endcsname{0}
\expandafter\def\csname amb@n-free-ask-any\endcsname{5}
\expandafter\def\csname amb@models-free-ask-any\endcsname{Gemma 4 12B, Gemini 3.8 Flash, Gemini 3.1 Pro, Qwen3.8-27B, Gemini 3.1 Flash-Lite}
\expandafter\def\csname amb@n-free-level-both\endcsname{4}
\expandafter\def\csname amb@models-free-level-both\endcsname{Qwen2.5-VL-7B, InternVL3.5-8B, Mistral-Small-3.2-24B, Qwen3.5-9B}
\expandafter\def\csname amb@n-free-uq-any\endcsname{0}
\expandafter\def\csname amb@models-free-uq-any\endcsname{none}
\expandafter\def\csname amb@n-label-ask-any\endcsname{4}
\expandafter\def\csname amb@models-label-ask-any\endcsname{Gemma 4 12B, Gemini 3.8 Flash, Gemini 3.1 Pro, Gemini 3.1 Flash-Lite}
\expandafter\def\csname amb@n-label-uq-any\endcsname{3}
\expandafter\def\csname amb@models-label-uq-any\endcsname{InternVL3.5-8B, Qwen3.5-9B, Qwen3.8-27B}
\expandafter\def\csname amb@n-near-constant-any\endcsname{4}
\expandafter\def\csname amb@models-near-constant-any\endcsname{Qwen2.5-VL-7B, InternVL3.5-8B, Qwen3.5-9B, Qwen3.8-27B}
\expandafter\def\csname amb@auroc-msp-nabove\endcsname{4}
\expandafter\def\csname amb@auroc-msp-n\endcsname{6}
\expandafter\def\csname amb@auroc-msp-min\endcsname{0.57}
\expandafter\def\csname amb@auroc-msp-max\endcsname{0.69}
\expandafter\def\csname amb@auroc-ppl-nabove\endcsname{5}
\expandafter\def\csname amb@auroc-ppl-n\endcsname{6}
\expandafter\def\csname amb@auroc-ppl-min\endcsname{0.57}
\expandafter\def\csname amb@auroc-ppl-max\endcsname{0.74}
\expandafter\def\csname amb@auroc-mte-nabove\endcsname{5}
\expandafter\def\csname amb@auroc-mte-n\endcsname{6}
\expandafter\def\csname amb@auroc-mte-min\endcsname{0.56}
\expandafter\def\csname amb@auroc-mte-max\endcsname{0.73}
\expandafter\def\csname amb@auroc-pmi-nabove\endcsname{3}
\expandafter\def\csname amb@auroc-pmi-n\endcsname{6}
\expandafter\def\csname amb@auroc-pmi-min\endcsname{0.49}
\expandafter\def\csname amb@auroc-pmi-max\endcsname{0.63}
\expandafter\def\csname amb@auroc-ptrue-nabove\endcsname{0}
\expandafter\def\csname amb@auroc-ptrue-n\endcsname{6}
\expandafter\def\csname amb@auroc-ptrue-min\endcsname{0.36}
\expandafter\def\csname amb@auroc-ptrue-max\endcsname{0.49}
\expandafter\def\csname amb@auroc-se-nabove\endcsname{7}
\expandafter\def\csname amb@auroc-se-n\endcsname{9}
\expandafter\def\csname amb@auroc-se-min\endcsname{0.53}
\expandafter\def\csname amb@auroc-se-max\endcsname{0.71}
\expandafter\def\csname amb@auroc-lexsim-nabove\endcsname{8}
\expandafter\def\csname amb@auroc-lexsim-n\endcsname{9}
\expandafter\def\csname amb@auroc-lexsim-min\endcsname{0.54}
\expandafter\def\csname amb@auroc-lexsim-max\endcsname{0.66}
\expandafter\def\csname amb@auroc-degmat-nabove\endcsname{7}
\expandafter\def\csname amb@auroc-degmat-n\endcsname{9}
\expandafter\def\csname amb@auroc-degmat-min\endcsname{0.52}
\expandafter\def\csname amb@auroc-degmat-max\endcsname{0.65}
\expandafter\def\csname amb@auroc-luq-nabove\endcsname{5}
\expandafter\def\csname amb@auroc-luq-n\endcsname{9}
\expandafter\def\csname amb@auroc-luq-min\endcsname{0.52}
\expandafter\def\csname amb@auroc-luq-max\endcsname{0.65}
\expandafter\def\csname amb@gemini-thoughts-median-range\endcsname{97--175}

\newcommand{\rfHccAurocMed}{0.60}

\newcommand{\rfHccAurocUpMed}{0.69}
\newcommand{\rfHccCcAbove}{643}
\newcommand{\rfHccCcBelow}{105}
\newcommand{\rfHccCcMed}{0.15}
\newcommand{\rfHccFixedAuroc}{0.68}
\newcommand{\rfHccFixedMed}{0.30}
\newcommand{\rfHccInitWrongMed}{29}
\newcommand{\rfHccLibSameMed}{$-$0.15}

\newcommand{\rfHccPtrueAbove}{40}
\newcommand{\rfHccPtrueBelow}{21}

\newcommand{\rfHccSevenAbove}{597}
\newcommand{\rfHccSevenBelow}{38}
\newcommand{\rfHccSevenCells}{1,586}

\newcommand{\rfLenCapHi}{99}
\newcommand{\rfLenCapLo}{76}

\newcommand{\rfLenLiteTok}{0.57 [0.38, 0.70]}

\newcommand{\rfLenLowRefHi}{24}
\newcommand{\rfLenLowRefLo}{8}

\newcommand{\rfTrFlash}{0.66}

\newcommand{\rfTrFlashCells}{160}
\newcommand{\rfTrFlashCut}{775}
\newcommand{\rfTrFlashCutPct}{80}
\newcommand{\rfTrFlashFull}{0.68}

\newcommand{\rfTrFlashNoapi}{0.50}
\newcommand{\rfTrFlashNoapiAbove}{28}
\newcommand{\rfTrFlashNoapiBelow}{18}
\newcommand{\rfTrFlashNoapiDefined}{116}
\newcommand{\rfTrFlashTrans}{969}
\newcommand{\rfTrLite}{0.55}

\newcommand{\rfTrLiteFull}{0.54}

\newcommand{\rfTrLiteNoapi}{0.55}

\newcommand{\rfTrPro}{0.63}

\newcommand{\rfTrProFull}{0.62}

\newcommand{\rfTrProNoapi}{0.62}

\newcommand{\rfUrrBothCross}{7}
\newcommand{\rfUrrBothImage}{95}
\newcommand{\rfUrrBothText}{51}
\newcommand{\rfUrrFwdCross}{14}
\newcommand{\rfUrrFwdImage}{112}
\newcommand{\rfUrrFwdText}{61}
\newcommand{\rfUrrRevCross}{10}

\newcommand{\rfUrrRevText}{62}

\title{VLM-UQBench: A Benchmark for Modality-Specific and Cross-Modality Uncertainties in
Vision Language Models}

\author{
 \textbf{Chenyu Wang\textsuperscript{1}\thanks{Equal contribution.}},
 \textbf{Tianle Chen\textsuperscript{1}\footnotemark[1]},
 \textbf{Yuxin Wu\textsuperscript{2}},
 \textbf{Qijia He\textsuperscript{3}},
\\
 \textbf{Mingyu Chen\textsuperscript{1}},
 \textbf{H.~M.~Sabbir Ahmad\textsuperscript{1}},
 \textbf{Kayhan Batmanghelich\textsuperscript{1}},
 \textbf{Wenchao Li\textsuperscript{1}}
\\
\\
 \textsuperscript{1}Boston University,
 \textsuperscript{2}Amazon,
 \textsuperscript{3}University of Washington
\\
 \small{\texttt{\{chyuwang, tianle, sabbir92, batman, wenchao\}@bu.edu}, \texttt{yuxinwu139@gmail.com}, \texttt{heqj3@uw.edu}}
\\
 \small{Code: \url{https://github.com/cskyl/vlm-uqbench} \quad Project page: \url{https://cskyl.github.io/vlm-uqbench/}}
}

\renewcommand{\topfraction}{0.85}
\renewcommand{\dbltopfraction}{0.85}
\renewcommand{\textfraction}{0.1}
\begin{document}
\maketitle
\addtocontents{toc}{\protect\setcounter{tocdepth}{-1}}
\begin{abstract}
Uncertainty quantification (UQ) is essential for ensuring the safety and reliability of vision--language models (VLMs). A central challenge is to identify the source of uncertainty, whether it arises from the image, the text, or their interaction. We introduce VLM-UQBench, a benchmark for modality-specific and cross-modal data uncertainty in VLMs. It combines human-curated VizWiz subsets covering clean, image-, text-, and cross-modal uncertainty with a scalable perturbation pipeline comprising visual, textual, and cross-modal perturbations. We further introduce two metrics to assess how UQ scores respond to these perturbations and how their changes correlate with hallucinations. Using this framework, we systematically benchmark both LLM-based and VLM-specific UQ methods across nine VLMs and three datasets. Empirically, we find that (i) UQ performance varies substantially across uncertainty sources and VLMs; (ii) modality-specific uncertainty frequently co-occurs with hallucinations, yet changes in UQ scores only weakly track these failures; and (iii) we uncover response-length confounding in UQ evaluation, where seemingly effective uncertainty detectors may rely on simple answer-length cues. These findings highlight limitations in both current UQ methods and their evaluation, calling for more robust, modality-aware UQ approaches and more rigorous evaluation protocols that go beyond detection performance to assess the reliability of uncertainty signals.
\end{abstract}

\section{Introduction}
\label{sec:intro}

\begin{figure*}[t]
  \centering
    \vspace{-0.4em} %
  \includegraphics[width=\textwidth]{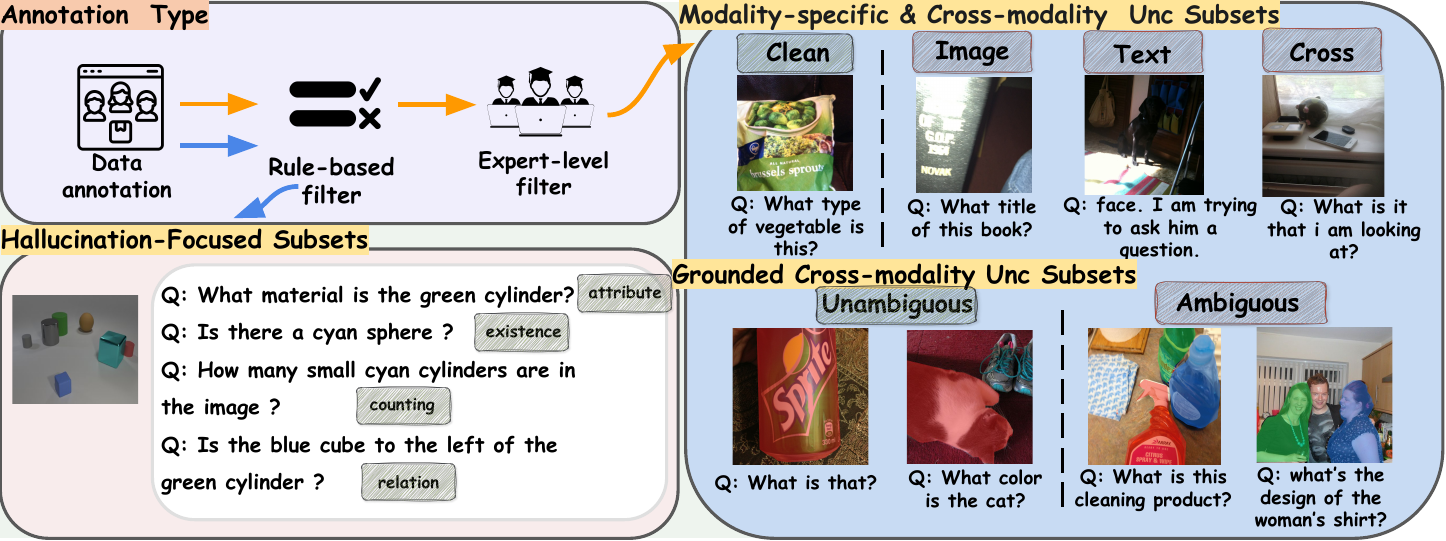}
  
  \caption{\textbf{Benchmark composition and curation workflow.}
We integrate diverse data sources through a three-stage pipeline of \emph{human annotation}, \emph{rule-based filtering}, and \emph{expert curation} (upper left), yielding four VizWiz-based subsets with modality-specific uncertainty labels (Clean, Image, Text, Cross; top right).  
Hallucination-focused subsets are built from CLEVR scene graphs using compositional templates and rule-based generation, enabling controlled evaluation of attribute, existence, counting, and relation hallucinations (bottom left).
  }

  \label{fig:workflow_overview}
\end{figure*}

Uncertainty quantification (UQ) is becoming a core requirement for vision–language models (VLMs) deployed in downstream tasks such as assistive question answering, everyday visual assistants, robotics, and medical or scientific image analysis~\cite{venkataramanan2025probabilistic, huang2024review, ren2023robots, wang2025semantic}. In these settings, VLMs must not only produce answers but also decide when to abstain, ask for clarification, or escalate to a human, since hallucinations can be costly or harmful. Effective UQ must therefore indicate both how uncertain the model is and where that uncertainty arises—for example, whether it is caused by a low-quality or ambiguous image, an under-specified or conflicting text query, or a mismatch between the two. Such source-aware UQ is crucial for taking the appropriate downstream action (retake the image, rephrase the question, or defer), yet current VLM systems provide little structured support for this modality-specific view of uncertainty.

Most existing work on UQ adopts the classical epistemic--aleatoric decomposition, attributing uncertainty either to the model (epistemic) or to the data (aleatoric). Recent work in Natural Language Generation (NLG) and LLMs questions whether this view is expressive enough for modern generative systems. Recent work catalogues sources of uncertainty in NLG and proposes a two-dimensional taxonomy that separates where uncertainty comes from and how it manifests, arguing that this is more informative than the aleatoric/epistemic dichotomy~\cite{baan2023uncertainty}. In a similar spirit, subsequent work reviews the standard distinction for LLMs and shows that it misses key sources of uncertainty introduced by large-scale pre-training, alignment, and user interaction, instead distinguishing operational uncertainty (along the model lifecycle) from output uncertainty (in generated content)~\cite{beigi2024rethinking}. Further analysis of reducibility-based definitions, where epistemic uncertainty is treated as reducible and aleatoric as irreducible, shows that for interactive agents the same situation can often be made more or less “reducible” by acquiring additional information or taking actions, making a strict binary split unstable~\cite{kirchhof2025position}. Taken together, these works suggest that the classical epistemic/aleatoric view is too coarse for modern generative models: we need uncertainty descriptions that are both source-aware and explicit about reducibility, which is especially important for VLMs, where uncertainty may originate from the image, the text, or their interaction, and where model architectures and training paradigms are even more complex.

In parallel with these conceptual critiques, several recent works argue that progress in UQ requires matching progress in benchmarks and evaluation protocols, and explicitly call for more comprehensive, task-diverse benchmarks that go beyond a single scalar confidence score and standard QA datasets~\cite{beigi2024rethinking,kirchhof2025position}. In the VLM setting, existing datasets make only partial steps in this direction. UNK-VQA~\cite{guo2024unk} and FocusAmbiguity~\cite{chen2025acknowledging} provide useful labels for answerability and grounded ambiguity in open-ended and free-form VQA, but the uncertainty types they cover are limited. MM-UPD~\cite{miyai2025unsolvable} offers more fine-grained unsolvability categories, but is restricted to multiple-choice VQA. Besides, these benchmarks are all created at a group level, without instance-wise contrastive structure that would support controlled evaluation. MuirBench~\cite{wang2024muirbench} moves toward such contrastive design by constructing pairwise instances, where each standard example is paired with an unanswerable variant, but it is still limited to multiple-choice VQA and focuses only on answerability versus unanswerability. CertainlyUncertain~\cite{chandu2024certainly} provides fine-grained, instance-wise annotations for free-form VQA, but its annotations are defined within the aleatoric--epistemic dichotomy, which is not well aligned with the modality-specific uncertainty needs of VLMs. Moreover, most of these datasets were originally developed to evaluate model performance, robustness, or answerability, rather than to directly probe UQ methods. A dedicated benchmark for UQ in VLMs should instead offer fine-grained, instance-wise structure across multiple tasks, and explicitly support modality-aware uncertainty analysis that can separate image, text, and cross-modal sources.

To address these gaps, we introduce VLM-UQBench, a benchmark specifically designed for modality-specific uncertainty in visual question answering. At its core, VLM-UQBench contains four subsets derived from VizWiz~\cite{gurari2018vizwiz}: a clean set, an image-uncertainty set, a text-uncertainty set, and a cross-modality uncertainty set. Each example is selected from human annotations of why answers disagree and checked by experts, giving labels for whether uncertainty comes from the image, the text, or their interaction. These subsets offer rich, modality-aware annotations that go beyond coarse epistemic/aleatoric tags.

On top of the annotated subsets, VLM-UQBench includes a scalable perturbation pipeline and hallucination-focused subsets. The perturbation pipeline introduces instance-wise, fine-grained uncertainty through targeted visual, textual, and cross-modal edits, allowing us to construct contrastive pairs and probe how UQ methods respond to controlled, modality-specific perturbations without additional human labels. The hallucination subsets, built from synthetic scenes with full scene graphs, further enable systematic evaluation of how different UQ methods relate to hallucination behaviour under controlled visual and textual changes.

\noindent\textbf{The contributions of this paper are fourfold:}
\begin{itemize}[leftmargin=1.5em,itemsep=1pt,topsep=2pt,parsep=0pt]  %
    \item \textbf{Modality-specific benchmark.}
    We formalize and release \textbf{VLM-UQBench}, which, as described above, provides modality-specific uncertainty labels for image, text, and cross-modal sources in free-form VQA, built from real-world data.

    \item \textbf{Scalable pipeline and paired metrics.}
    We design a scalable perturbation pipeline that injects controllable uncertainty through seven image edits, five text edits, and three cross-modal perturbations, and introduce two metrics, \emph{Uncertainty Reflection Rate (URR)} and \emph{Hallucination Consistency Coefficient (HCC)}, to evaluate how UQ scores respond to these perturbations and relate to hallucinations introduced.

    \item \textbf{Systematic study of UQ in VLMs.}
    We benchmark nine UQ methods across nine VLMs, together with five VLM-specific UQ methods, on three datasets and their modality-targeted subsets. We find (i) UQ performance that depends on the uncertainty source, (ii) a further dependence on the VLM, and (iii) a weak link between changes in UQ scores and the hallucinations induced by modality-specific uncertainty.

    \item \textbf{Answer-length confounding.}
    We uncover response-length confounding in UQ evaluation, where seemingly effective uncertainty detectors may rely on simple answer-length cues; the confound depends on the model and the uncertainty source.

\end{itemize}

\section{Related Work}
\label{sec:related_work}
\subsection{UQ for LLM/VLLM}
Uncertainty estimation methods for LLMs and VLMs are often categorized into white-box approaches, which require access to logits or hidden states, and black-box approaches, which operate only on generated samples~\cite{shorinwa2025survey}. White-box methods include information-based scores such as Perplexity, Maximum Sequence Probability, Mean Token Entropy~\cite{fomicheva2020unsupervised}, and Pointwise Mutual Information (PMI)~\cite{takayama2019relevant}, as well as Semantic Entropy, which clusters semantically equivalent outputs~\cite{kuhn2023semantic}, and reflexive methods like p(True) that elicit explicit confidence judgments~\cite{kadavath2022language}. Black-box measures instead exploit output diversity, using graph-based statistics such as DegMat or spectral Laplacians~\cite{lin2024generating} and simple proxies like Lexical Similarity~\cite{fomicheva2020unsupervised}. These methods have been effective for hallucination detection and selective prediction in LLMs and VLMs~\cite{zhang2024vl,farquhar2024detecting,srinivasan2024selective}, but they treat uncertainty as a single scalar and offer little insight into whether it arises from the image, the text, or their multimodal interaction~\cite{kuhn2023semantic,srinivasan2024selective}. VLM-specific methods additionally use the image, by modifying the input and checking answer consistency~\cite{zhang2024vl,bhattacharya2025festa}, masking salient image regions~\cite{park2026vauq}, or reading visual tokens and hidden states~\cite{fang2025dropout,khanmohammadi2026grounded}. We address this by systematically evaluating nine LLM UQ and five VLM-specific methods on VLM-UQBench, which offers labeled modality-specific and cross-modal uncertainty and contrastive samples for fine-grained evaluation.
\subsection{UQ Benchmark}
 Benchmarks for UQ in language and vision-language models mainly test whether uncertainty correlates with output quality, hallucination, or ambiguity. For LLMs, selective QA and generation tasks on datasets such as CoQA~\cite{reddy2019coqa}, TriviaQA~\cite{joshi2017triviaqa}, MMLU~\cite{hendrycks2020measuring}, and GSM8k~\cite{cobbe2021training} underpin benchmarks like LM-Polygraph~\cite{shelmanovvashurin2025}, while TruthfulQA~\cite{lin2021truthfulqa}, AmbigQA~\cite{min2020ambigqa}, and AmbigEnt~\cite{hou2023decomposing} focus on hallucination and aleatoric ambiguity via claim verification and multi-answer annotations. In the VLM setting, the \textit{Certainly Uncertain} benchmark~\cite{chandu2024certainly} is among the first to provide multimodal epistemic versus aleatoric labels in VQA, but it is restricted to this two-way categorization and does not capture modality-specific uncertainty transfer between image, text, and their interaction. Our VLM-UQBench addresses this gap by introducing explicit modality-specific uncertainty (image, text, cross-modal) together with a scalable perturbation pipeline for controlled sample-wise uncertainty injection (Table~\ref{tab:uq_benchmark_main}).

\section{VLM-UQBench}
\label{sec:benchmark}
\begin{figure*}[t]
  \centering
  \vspace{-0.6em} %
  \includegraphics[width=0.9\textwidth]{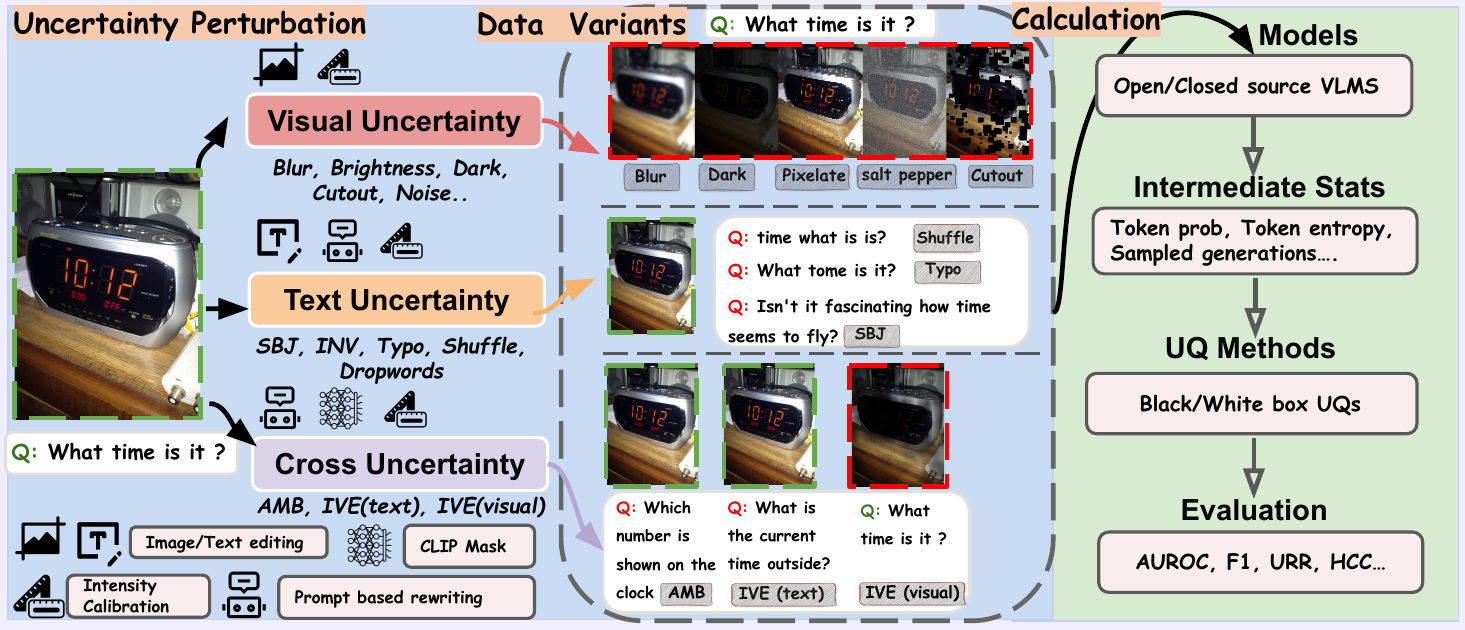}
  \vspace{-0.6em} %
  \caption{\textbf{Overview of our pipeline.}
A clean image–question pair is processed through three stages: \emph{uncertainty perturbation}, 
\emph{variant generation}, and \emph{uncertainty calculation}. In the perturbation stage (left), we inject 
\emph{visual}, \emph{textual}, and \emph{cross-modal} uncertainty (e.g., blur, brightness, and occlusion for images; subjective or invalid rewrites, typos, and shuffles 
for text; AMB- and IVE-based edits for cross-modality). Perturbation intensity is chosen to avoid being too weak (no effect) 
or too strong (catastrophic failure); we manually select these levels using a visualization tool, 
as detailed in Appendix~\ref{app:calibration}. In the variant-generation stage (middle), each original clean pair is expanded into a set of perturbed counterparts. In the calculation stage (right), all variants are fed into 
VLMs, from which we collect intermediate statistics (such as token probabilities, entropy, and 
sampled generations). Different UQ methods use these statistics to compute
uncertainty scores, which are evaluated with AUROC, URR, and HCC to study modality sensitivity and the
relationship between uncertainty and hallucinations.}
  \label{fig:pipeline}
  \vspace{-0.5em} %
\end{figure*}
\begin{table*}[t]
\centering
\scriptsize
\setlength{\tabcolsep}{5pt}
\renewcommand{\arraystretch}{0.95}
\caption{
Comparison of existing uncertainty benchmarks and VLM-UQBench on task modality, annotation type, and contrast granularity.
}
\vspace{0.4em}
\resizebox{\textwidth}{!}{
\begin{tabular}{@{}l l l c@{}}
\toprule
\textbf{Benchmark} & \textbf{Task / Modality} & \textbf{Annotation Type} & \textbf{Contrast Granularity} \\
\midrule
\textbf{AmbigQA}~\cite{min2020ambigqa} & Open-domain QA & Ambiguity & Group \\
\textbf{AbstentionBench}~\cite{kirichenko2025abstentionbench} & Open-domain QA & Abstention Scenarios & Group \\
\textbf{UNK-VQA}~\cite{guo2024unk} & Open-ended VQA & Answerability & Group \\
\textbf{VQ-FocusAmbiguity}~\cite{chen2025acknowledging} & Free-form VQA & Ambiguity (Grounded) & Group \\
\textbf{MM-UPD}~\cite{miyai2025unsolvable} & Multiple-choice VQA & Unsolvability Categories & Group \\
\textbf{MuirBench}~\cite{wang2024muirbench} & Multiple-choice VQA & Answerability & Sample \\
\textbf{CertainlyUncertain}~\cite{chandu2024certainly} & Free-form VQA & Aleatoric--Epistemic Uncertainty & Sample \\
\midrule
\rowcolor{gray!10}
\textbf{VLM-UQBench (Ours)} & Free-form VQA & \textit{Modality-specific \& Cross-Modality} Uncertainty & Sample \\
\bottomrule
\end{tabular}
}
\label{tab:uq_benchmark_main}
\end{table*}
VLM-UQBench combines three components into a unified benchmark for uncertainty in VLMs:
(i) human-annotated modality-specific subsets from VizWiz (clean, image, text, cross-modal),
(ii) hallucination-focused subsets constructed from CLEVR scene graphs, and
(iii) a synthetic perturbation pipeline that injects modality-specific uncertainty into standard VQA datasets.
Table~\ref{tab:uq_benchmark_main} summarizes how VLM-UQBench relates to existing uncertainty and ambiguity benchmarks.
\subsection{Benchmark Overview}
\label{sec:data_sources}
\noindent\textbf{Human-Annotated Modality Subsets (VizWiz).} 
We build the core modality-specific component of VLM-UQBench from the VizWiz dataset, which contains images captured by blind users, natural language questions, and ten crowd-sourced answers per question. Crucially, VizWiz provides annotations of plausible reasons for answer disagreement~\cite{bhattacharya2019does}, including low-quality images, insufficient evidence, invalid or ambiguous questions, and subjective judgments.

For our benchmark, we focus on input-modality factors and group these labels into three categories:
\emph{visual uncertainty} (e.g., low-quality image artifacts such as blur, occlusion, or poor lighting),
\emph{textual uncertainty} (e.g., invalid or subjective questions that are vague or incoherent),
and \emph{cross-modality uncertainty} (e.g., ambiguous references or insufficient visual evidence given an otherwise clear question and image).
Details of the full taxonomy and mapping are provided in Appendix~\ref{app:vizwiz}.

We then apply a filtering rule to construct non-overlapping subsets and have three domain-expert PhD annotators curate high-quality examples, yielding 150 samples for each modality-specific subset (\nCrossUnique{} unique for the cross-modality subset), plus a clean subset with little modality-induced uncertainty. These curated splits provide clear, human-labeled sources of visual, textual, and cross-modal uncertainty for evaluating UQ methods in VLMs.

\noindent\textbf{Hallucination-Focused Subset (CLEVR-Hallucination).} 
To analyze how modality-specific uncertainty contributes to different types of hallucination in VQA, we construct controlled evaluation sets using the CLEVR dataset~\cite{johnson2017clevr}.
Scene graphs and compositional templates are leveraged to generate four hallucination categories: attribute, existence, counting, and relation.
Each question is grounded in object-level metadata, enabling reliable assessment.
We keep the 462 questions whose answers we verified against the scene graph; further dataset generation details are provided in Appendix~\ref{app:clevr}.

\subsection{Generalizing to Datasets Without Annotations: Synthetic Perturbation Pipeline}

\label{sec:perturbation}
While the modality-specific and cross-modality subsets provide strong annotated uncertainty signals, they are costly to scale. To extend evaluation to larger and more diverse VQA datasets, we introduce a synthetic perturbation pipeline that automatically converts standard datasets into uncertainty-aware benchmarks without additional labels. The pipeline applies aligned perturbations to images, text, or both, corresponding to visual ($U_v$), textual ($U_t$), and cross-modal ($U_x$) uncertainty, with tunable intensity for controlled evaluation.

The formal definition of the pipeline is given in Appendix~\ref{app:pipeline}.
Implementation details and augmentation parameters are provided in Appendix~\ref{app:augmentation}.

\subsection{Evaluation}
\label{sec:evaluation}
We employ both conventional classification metrics and tailored correlation measures 
to evaluate UQ performance across modality-specific groups
(image, text, and cross-modality) in VLMs. \textbf{AUROC} assesses how well each UQ method 
distinguishes uncertain samples from clean ones within each subset. Beyond these aggregate metrics, 
we introduce two measures that explicitly link UQ to modality-targeted perturbations and hallucination behavior.

\paragraph{Uncertainty Reflection Rate (URR).}
To test whether a UQ method responds appropriately to injected uncertainty, 
we define the \emph{Uncertainty Reflection Rate} for a given UQ method $u$ and a fixed perturbation
operator $T^{(k)}$ with intensity $\lambda$ (as introduced in Appendix~\ref{app:pipeline}).
Given a clean input $x_i$ and its perturbed version $x_i^{\text{pert}} = T^{(k)}(x_i; \lambda)$,
let $U^{\text{clean}}_i = u(x_i)$ and $U^{\text{pert}}_i = u(x_i^{\text{pert}})$ be the corresponding uncertainty scores.
We set $\delta_i = 1$ if $U^{\text{pert}}_i > U^{\text{clean}}_i$ and $\delta_i = 0$ otherwise, and define $\mathrm{URR}(u, T^{(k)}, \lambda) = \frac{1}{n} \sum_{i=1}^{n} \delta_i$.

A higher $\mathrm{URR}$ indicates that method $u$ more consistently increases its uncertainty under this perturbation,
showing stronger modality-specific sensitivity. Because re-sampling alone changes sampling-based scores, we compare URR with the \emph{sampling-noise level}, the URR between two halves of the answers sampled for the same clean input (Appendix~\ref{app:stats}), rather than with 0.5.

\paragraph{Hallucination Consistency Coefficient (HCC).}
To evaluate whether \emph{changes} in uncertainty track hallucination behavior, 
we use the \emph{Hallucination Consistency Coefficient} for a UQ method $u$ under a fixed perturbation setting.
For each sample $i$, let $U^{\text{clean}}_i = u(x_i)$, $U^{\text{pert}}_i = u(x_i^{\text{pert}})$, 
and $\Delta U_i = U^{\text{pert}}_i - U^{\text{clean}}_i$, and let $H_i \in \{0,1\}$ be the hallucination label (1 = correct on the clean input and wrong on the perturbed one); pairs whose perturbed answer is an abstention are left out.
We compute the point-biserial correlation between $\Delta U$ and $H$, which in closed form is
\begin{equation}
\label{eq:hcc-delta-group}
\mathrm{HCC}(\Delta u) = 
\frac{\overline{\Delta U}_1 - \overline{\Delta U}_0}{s_{\Delta U}}
\sqrt{\frac{n_1 n_0}{n^2}},
\end{equation}
where $\overline{\Delta U}_1$ and $\overline{\Delta U}_0$ are the mean uncertainty changes 
for hallucinated and non-hallucinated samples, $s_{\Delta U}$ is the standard deviation of all $\Delta U_i$ values,
$n_1$ and $n_0$ are their counts, and $n = n_1 + n_0$.
A positive $\mathrm{HCC}(\Delta u)$ indicates that hallucinated cases tend to exhibit larger uncertainty increases 
under perturbation, suggesting the UQ reliably reflects hallucination sensitivity.

\section{Experiments}
\label{sec:setup}
\definecolor{ctrlabove}{RGB}{0,115,90}\definecolor{ctrlbelow}{RGB}{200,50,0}\definecolor{vlmbg}{RGB}{229,240,250}
\begin{table*}[t]
\centering
\small
\setlength{\tabcolsep}{7pt}
\renewcommand{\arraystretch}{1.0}
\caption{\textbf{Detecting each uncertainty source with UQ (AUROC$\uparrow$; chance 0.50).} Each column compares the VizWiz items of one source (Image, Text, Cross-modal) with the same clean items. Shaded rows: VLM-specific UQ; unshaded: LLM UQ. Indented rows are baselines of the VLM-specific method above them; green or red marks a method above or below its baseline(s). ``--'': not applicable (no white-box access to closed-source models). Bold: highest value of the column within each block. $^{\ast}$Question-only P(True) prompt. Gemma 4 12B and Gemini 3.1 Pro: Table~\ref{tab:det_merged_app}; LLaVA-1.5-7B: Appendix~\ref{app:vlmuq}.}
\label{tab:det_llmuq_main}\label{tab:slim-native-detection-main}
\begin{tabular}{lcccccc}
\toprule
 & \multicolumn{3}{c}{\textbf{Open-source}} & \multicolumn{3}{c}{\textbf{Closed-source}} \\
\cmidrule(lr){2-4} \cmidrule(lr){5-7}
 & \multicolumn{3}{c}{Qwen2.5-VL-7B} & \multicolumn{3}{c}{Gemini 3.8 Flash} \\
\cmidrule(lr){2-4} \cmidrule(lr){5-7}
\textbf{Method} & Image & Text & Cross & Image & Text & Cross \\
\midrule
\rowcolor{black!6}\multicolumn{7}{l}{\textbf{Black-box: sampled answers}} \\
\multicolumn{7}{l}{\textit{LLM UQ}} \\
Semantic Entropy & \textbf{0.85} & \textbf{0.89} & \textbf{0.80} & 0.81 & 0.83 & \textbf{0.75} \\
Lexical Similarity & \textbf{0.85} & 0.87 & 0.76 & \textbf{0.82} & \textbf{0.85} & 0.72 \\
DegMat & 0.73 & 0.70 & 0.60 & 0.80 & 0.81 & 0.70 \\
LUQ & 0.61 & 0.63 & 0.54 & 0.66 & 0.70 & 0.57 \\
\addlinespace[2pt]
\rowcolor{vlmbg}\multicolumn{7}{l}{\textit{VLM-specific UQ}} \\
\rowcolor{vlmbg}VL-Uncertainty & \textcolor{ctrlbelow}{0.61} & \textcolor{ctrlbelow}{0.64} & \textcolor{ctrlbelow}{0.63} & 0.51 & 0.62 & 0.62 \\
\rowcolor{vlmbg}\quad SE, same sampler & \textbf{0.65} & \textbf{0.67} & \textbf{0.63} & \textbf{0.59} & \textbf{0.70} & \textbf{0.66} \\
\midrule
\rowcolor{black!6}\multicolumn{7}{l}{\textbf{White-box: logits, attention or hidden states}} \\
\multicolumn{7}{l}{\textit{LLM UQ}} \\
MSP & \textbf{0.84} & \textbf{0.88} & \textbf{0.82} & {\color{black!35}--} & {\color{black!35}--} & {\color{black!35}--} \\
Perplexity & 0.80 & 0.81 & 0.77 & {\color{black!35}--} & {\color{black!35}--} & {\color{black!35}--} \\
Mean token entropy & 0.81 & 0.81 & 0.78 & {\color{black!35}--} & {\color{black!35}--} & {\color{black!35}--} \\
PMI & 0.79 & 0.82 & 0.78 & {\color{black!35}--} & {\color{black!35}--} & {\color{black!35}--} \\
P(True) & 0.27$^{\ast}$ & 0.26$^{\ast}$ & 0.43$^{\ast}$ & {\color{black!35}--} & {\color{black!35}--} & {\color{black!35}--} \\
\addlinespace[2pt]
\rowcolor{vlmbg}\multicolumn{7}{l}{\textit{VLM-specific UQ}} \\
\rowcolor{vlmbg}VAUQ & \textcolor{ctrlabove}{0.80} & \textcolor{ctrlabove}{0.82} & \textcolor{ctrlabove}{0.76} & {\color{black!35}--} & {\color{black!35}--} & {\color{black!35}--} \\
\rowcolor{vlmbg}\quad unmasked entropy & 0.78 & 0.81 & 0.74 & {\color{black!35}--} & {\color{black!35}--} & {\color{black!35}--} \\
\rowcolor{vlmbg}\quad random mask & 0.75 & 0.78 & 0.74 & {\color{black!35}--} & {\color{black!35}--} & {\color{black!35}--} \\
\rowcolor{vlmbg}Response SVAR & 0.83 & 0.82 & 0.68 & {\color{black!35}--} & {\color{black!35}--} & {\color{black!35}--} \\
\rowcolor{vlmbg}Dropout Dec. (token entropy) & 0.81 & 0.70 & \textbf{0.79} & {\color{black!35}--} & {\color{black!35}--} & {\color{black!35}--} \\
\rowcolor{vlmbg}Dropout Dec. (token disagreement) & 0.23 & 0.32 & 0.35 & {\color{black!35}--} & {\color{black!35}--} & {\color{black!35}--} \\
\rowcolor{vlmbg}BICR & \textcolor{ctrlabove}{\textbf{0.90}} & \textcolor{ctrlabove}{\textbf{0.89}} & \textcolor{ctrlabove}{0.54} & {\color{black!35}--} & {\color{black!35}--} & {\color{black!35}--} \\
\rowcolor{vlmbg}\quad no ranking term & 0.86 & 0.87 & 0.51 & {\color{black!35}--} & {\color{black!35}--} & {\color{black!35}--} \\
\bottomrule
\end{tabular}
\end{table*}

We evaluate nine UQ methods covering both white-box and black-box approaches, and five VLM-specific UQ methods.  
Our experiments span three datasets and nine models to assess modality-specific uncertainty reflection and hallucination behavior under controlled conditions.

\subsection{Experiment Setup}
\paragraph{Datasets.}
We evaluate across datasets spanning real-world and synthetic settings.
The human-annotated subsets from \textbf{VizWiz} provide modality-specific and cross-modality uncertainty labels for real images.
The \textbf{VizWiz-Clean}, \textbf{VQAv2}, and \textbf{CLEVR} datasets serve as bases for our perturbation pipeline, enabling controlled injection of fine-grained visual, textual, and cross-modal uncertainty for large-scale evaluation.
Finally, \textbf{CLEVR-Hallucination} offers scene-graph–based control for analyzing how modality uncertainty contributes to different hallucination types.
\paragraph{Models.}
We benchmark nine vision–language models spanning different architectures and access regimes. Four main models answer every subset and perturbation: two open-source models, \textbf{Gemma 4 12B}~\cite{gemma4} and \textbf{Qwen2.5-VL-7B}~\cite{wang2024qwen2,bai2025qwen25vl}, and two closed-source models, \textbf{Gemini 3.8 Flash}~\cite{gemini38flash} and \textbf{Gemini 3.1 Pro}~\cite{gemini31pro}. We further evaluate Qwen3.5-9B~\cite{qwen35}, Qwen3.8-27B~\cite{qwen38}, InternVL3.5-8B~\cite{internvl35}, Mistral-Small-3.2-24B~\cite{mistralsmall32} and Gemini 3.1 Flash-Lite~\cite{gemini31flashlite}; InternVL3.5-8B and Mistral-Small-3.2-24B answer only the VizWiz items (Appendix~\ref{app:generation}). All models are evaluated under two decoding regimes: 
(1) \emph{greedy decoding} with temperature $=0$ for deterministic inference, and 
(2) \emph{sampling-based decoding} with temperature $=0.7$ for diversity-driven UQ, where $N{=}10$ samples are drawn per input. 
All other decoding hyperparameters follow the LM-Polygraph protocol~\cite{shelmanovvashurin2025}.
\paragraph{Uncertainty Quantification Methods.}
We evaluate nine UQ methods spanning white-box and black-box approaches.
\textbf{White-box} methods (\emph{Mean Token Entropy}, \emph{Perplexity}, \emph{Maximum Sequence Probability} (MSP), \emph{PMI}, \emph{p(True)}) compute uncertainty directly from model logits or token probabilities, with p(True) eliciting reflexive confidence scores.
\textbf{Black-box} methods (\emph{Semantic Entropy}, \emph{Lexical Similarity}, \emph{DegMat}, \emph{LUQ}) estimate uncertainty from the variability and structure of $N{=}10$ sampled responses, with Semantic Entropy additionally clustering semantically similar outputs.
For the closed-source Gemini models, which do not expose token probabilities, only Semantic Entropy (from sample frequencies), Lexical Similarity, DegMat and LUQ are computed.
We further evaluate five \textbf{VLM-specific} UQ methods, VL-Uncertainty~\cite{zhang2024vl}, VAUQ and Response SVAR~\cite{park2026vauq}, Dropout Decoding~\cite{fang2025dropout} and BICR~\cite{khanmohammadi2026grounded}, on Qwen2.5-VL-7B and LLaVA-1.5-7B~\cite{liu2023visual,liu2024improved}, with all five also on Gemma 4 12B and VL-Uncertainty also on the closed-source Gemini models, each compared with a simpler baseline where one is defined (Appendix~\ref{app:vlmuq}).
\subsection{Experiment Results}

\begin{figure}[t]
  \centering
  \includegraphics[width=\columnwidth]{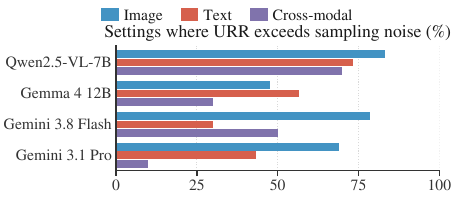}
  \caption{\textbf{Uncertainty responses to visual, textual, and cross-modal perturbations.} Share of perturbation
  settings in which the URR of three sampling-based scores (Lexical Similarity, DegMat, LUQ) exceeds sampling noise, i.e., the URR between two halves of the
  answers sampled for the same clean input (Appendix~\ref{app:res_urr}).}
  \label{fig:urr_summary}
\end{figure}

\begin{figure*}[t]
  \centering
  \includegraphics[width=\textwidth]{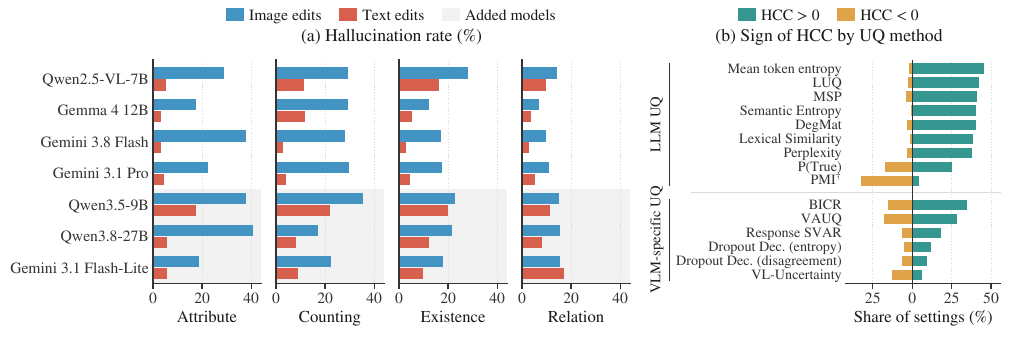}
  \caption{\textbf{Hallucinations and uncertainty changes under CLEVR perturbations.} \textbf{(a)} Hallucination rate: share of pairs whose
  correct answer becomes wrong after image or text perturbations; shaded rows: Qwen3.5-9B, Qwen3.8-27B and Gemini 3.1 Flash-Lite. \textbf{(b)} For each LLM UQ score and VLM-specific method (Qwen2.5-VL-7B and LLaVA-1.5-7B; Appendix~\ref{app:native-perturbation-matrices}), share of perturbation settings whose HCC is significantly above
  zero (right) or below zero (left). $^{\dagger}$PMI as implemented in LM-Polygraph. For Gemini 3.8 Flash, about \rfTrFlashCutPct\% of the transitions involve answers cut by the API output budget. Details: Appendix~\ref{app:res_hcc}.}
  \label{fig:claim2_hallucination_hcc}
\end{figure*}

\paragraph{UQ performance depends on the uncertainty source and the VLM.}
Table~\ref{tab:det_llmuq_main} reveals two key trends: UQ performance depends on the source of uncertainty, and it also depends on the model. On the curated VizWiz subsets, Semantic Entropy detects textual uncertainty best on most VLMs, while cross-modal uncertainty is the most challenging, with the lowest AUROC on 8 of the 9 VLMs (Table~\ref{tab:det_llmuq_full_auroc}). VLM-specific UQ methods improve over their baselines only in some settings and remain source-dependent: BICR, a supervised probe, reaches AUROC 0.90 on image and 0.89 on text uncertainty for Qwen2.5-VL-7B but 0.54 on cross-modal uncertainty, while VL-Uncertainty mostly falls below its Semantic Entropy baseline. Both matter: the source accounts for \detVarNineSource\% and the model for \detVarNineModel\% of the variation of the mean sampling-based AUROC over the nine VLMs, whereas for Semantic Entropy alone the model share is larger (\detVarNineSEModel\% vs.\ \detVarNineSESource\%; Table~\ref{tab:det_llmuq_variance}).

\paragraph{UQ scores respond inconsistently to perturbations.}
We analyze UQ sensitivity using our perturbation pipeline on VizWiz, VQAv2 and CLEVR for the four main models (Figure~\ref{fig:urr_summary}; full results in Appendix~\ref{app:res_urr}), where we observe that UQ sensitivity is highly inconsistent, varying substantially across datasets, models, and perturbation types. Among all perturbation types, UQ methods are overall most effective at detecting SBJ textual perturbations and image noise. In sharp contrast, they are worse at capturing cross-modal perturbations (16 of 40 settings above the sampling-noise level, against 117 of 168 visual and 61 of 120 textual settings), possibly because the GPT-4o rewrites, which were not human-validated, create little ambiguity. Answer-preserving edits, in contrast, leave URR within 0.02 of the sampling-noise level (Appendix~\ref{app:ext_benign}). When the ambiguity is exact by construction, sampled uncertainty does respond: editing CLEVR questions so that several objects fit the description raises URR by 0.28 more than an answer-preserving edit on Gemma 4 12B and Qwen2.5-VL-7B (Appendix~\ref{app:ext_clevr_amb}). %

\paragraph{Perturbations induce hallucinations that UQ scores track only weakly.}
Evaluating the Hallucination-Focused Subsets, we find two key results. First, as shown in Figure~\ref{fig:claim2_hallucination_hcc}(a), modality-specific uncertainty induces distinct hallucination types, mostly affecting attribute and counting questions. Visual perturbations are the most damaging; salt-and-pepper noise alone causes a 40.9\% hallucination rate on attribute tasks for Qwen2.5-VL-7B. Second, changes in UQ scores only weakly track these perturbation-induced hallucinations. Figure~\ref{fig:claim2_hallucination_hcc}(b) shows the Hallucination Consistency Coefficient (HCC($\Delta U$)) is weak and depends on the UQ method: only 693 of 1{,}902 settings have a significantly positive HCC, and the VLM-specific UQ methods show the same pattern. For no perturbation family is HCC significantly positive in more than half of the settings (24--46\%). The score change separates hallucinated pairs with a median AUROC of only \rfHccAurocMed{} (Appendix~\ref{app:res_hcc}). Further HCC results are provided in Appendix~\ref{app:res_hcc}.

\paragraph{Seemingly effective detectors can rely on answer length.}
Answer length alone, which is not a UQ score, separates Image from Clean items with AUROC \detLenGemmaImage, \detLenProImage{} and \detLenFlashLiteImage{} on Gemma 4 12B, Gemini 3.1 Pro and Gemini 3.1 Flash-Lite, above Semantic Entropy (\detSEGemmaImage, \detSEProImage{} and \detSEFlashLiteImage), and it beats Semantic Entropy in \detLenAboveSECount{} of the \detLenCells{} model--source settings with stored answers (Table~\ref{tab:det_llmuq_full_auroc}). The confound depends on the model and the source (Table~\ref{tab:length_main}): after controlling for length, Image detection drops to 0.67 on Gemma 4 12B and 0.51 on Gemini 3.1 Pro, where 71\% and 46\% of Image answers are refusals, while Text detection stays above 0.60 on all nine VLMs (Appendix~\ref{app:ext_length}).
\begin{table}[t]
\centering
\scriptsize
\setlength{\tabcolsep}{2.6pt}
\renewcommand{\arraystretch}{1.0}
\caption{\textbf{Detection beyond answer length (AUROC$\uparrow$).} \emph{Len.}: answer length alone; \emph{Ctrl.}: the
best sampling-based UQ score after controlling for answer length. \emph{Ref.}: refusals among Image / Clean greedy
answers (\%). $^{\ast}$Rerun with the sampler of the other models. Details:
Table~\ref{tab:ext_length}.}
\label{tab:length_main}
\resizebox{\ifdim\width>\columnwidth\columnwidth\else\width\fi}{!}{%
\begin{tabular}{@{}lccccccc@{}}
\toprule
 & \multicolumn{2}{c}{\textbf{Image}} & \multicolumn{2}{c}{\textbf{Text}} & \multicolumn{2}{c}{\textbf{Cross-modal}} & \\
\cmidrule(lr){2-3} \cmidrule(lr){4-5} \cmidrule(lr){6-7}
\textbf{Model} & Len. & Ctrl. & Len. & Ctrl. & Len. & Ctrl. & Ref. \\
\midrule
Gemma 4 12B & 0.87 & 0.67 & 0.86 & 0.77 & 0.75 & 0.77 & 71/3 \\
Gemini 3.8 Flash & 0.76 & 0.73 & 0.71 & 0.81 & 0.72 & 0.65 & 51/1 \\
Gemini 3.1 Pro & 0.78 & 0.51 & 0.70 & 0.78 & 0.66 & 0.62 & 46/2 \\
\midrule
InternVL3.5-8B & 0.50 & 0.85 & 0.65 & 0.84 & 0.70 & 0.74 & 8/1 \\
Mistral-Small-3.2-24B & 0.65 & 0.82 & 0.67 & 0.82 & 0.69 & 0.77 & 19/2 \\
Qwen3.5-9B & 0.71 & 0.81 & 0.76 & 0.85 & 0.74 & 0.70 & 19/0 \\
Qwen3.8-27B & 0.74 & 0.72 & 0.76 & 0.80 & 0.65 & 0.65 & 24/0 \\
Gemini 3.1 Flash-Lite & 0.84 & 0.62 & 0.80 & 0.72 & 0.72 & 0.68 & 63/3 \\
\midrule
Qwen2.5-VL-7B$^{\ast}$ & 0.75 & 0.79 & 0.73 & 0.82 & 0.73 & 0.67 & 27/2 \\
\bottomrule
\end{tabular}}
\end{table}

\subsection{Discussion}
Our results reveal that current UQ methods are not yet robust in the complex multimodal setting of VLMs. This is underscored by three key findings: (i) UQ performance depends on both the uncertainty source (e.g., Semantic Entropy detects textual uncertainty best and cross-modal uncertainty worst on most VLMs) and the VLM, with the best UQ method shifting across VLMs (mostly Semantic Entropy on open-weight models, Lexical Similarity on Gemini); (ii) modality-specific uncertainty induces hallucinations, yet changes in UQ scores track them only weakly; and (iii) seemingly effective detectors may rely on answer length, which alone rivals the UQ scores on degraded images for models that often refuse them. A length-only baseline is thus a simple check for UQ evaluation. Telling the sources apart, beyond detecting each against clean inputs, remains a natural next step.

Furthermore, our findings expose a disconnect between changes in UQ scores and hallucination risk. Modality-specific uncertainty induces hallucinations, yet the corresponding changes in UQ scores are weak and inconsistent across models. The score of the perturbed input alone ranks these hallucinations better (median AUROC \rfHccAurocUpMed{} vs.\ \rfHccAurocMed{} for the change; Table~\ref{tab:hcc_cc}), so current scores reflect the uncertainty of an input more than the change a perturbation causes. Detecting uncertain inputs and tracking the hallucinations they cause are thus distinct properties, and VLM-UQBench evaluates UQ methods on both.

\section{Conclusion}
\label{sec:conclusion}
We introduced VLM-UQBench, a modality-specific and cross-modality benchmark for UQ in VLMs. Our evaluation of nine UQ methods across nine VLMs, together with five VLM-specific UQ methods, on multiple datasets reveals that current approaches are not yet robust in the VLM setting. Their performance depends on both the uncertainty source and the model, seemingly effective detectors may rely on simple answer-length cues, and their changes correlate only weakly with perturbation-induced hallucinations in our controlled CLEVR setting. Our framework, with its modality-specific uncertainty subsets, scalable pipeline, and URR/HCC metrics, provides tools to probe these failures and begin to bridge this gap. We hope VLM-UQBench will spur more robust, modality-aware UQ methods for VLMs. 
\label{end:main-text}

\clearpage
\section*{Limitations}

\paragraph{What the benchmark measures.}
VLM-UQBench tests how UQ scores respond to data (aleatoric) uncertainty from the image, the question or their mismatch, not the model's epistemic uncertainty. Each source is compared with clean inputs, so the results concern detecting each source; attributing an uncertain input to its source is not evaluated. A hallucination is operationally a perturbation-induced correct-to-wrong transition on scene-audited CLEVR items (Appendix~\ref{app:stats}), and HCC is a correlation, not a causal account. The points below mark the conditions under which the findings hold.

\paragraph{Data.}
The human-labelled core is small: 572 unique VizWiz items, with 150 Clean, Image and Text items each and 122 unique Cross-modal items. Labels apply source-specific vote rules to existing crowd disagreement reasons~\cite{bhattacharya2019does} before experts remove erroneous or unclear pairs, so they inherit annotator subjectivity, and Cross-modal holds 113 ambiguous-reference and 9 evidence-absent items (Appendix~\ref{app:vizwiz}). On a separate VizWiz pool, in a comparison matched after seeing the data, wording-only disagreement gives AUROC 0.42--0.51, so we find no evidence that wording explains the Cross-modal results (Appendix~\ref{app:ext_wording}). AUROC half-widths of 0.04--0.07 (Appendix~\ref{app:stats}) do not support ranking close methods; small leads can change across sample draws (Appendix~\ref{app:ext_retest}). All VizWiz items come from its training split, which current VLMs may have seen.

\paragraph{Perturbations.}
Visual strengths were chosen by inspecting batches of 150 images, and no human validity rate was measured for the perturbed items (Appendix~\ref{app:calibration}). Two controls partly address this, although neither measures that validity: answer-preserving edits keep URR within 0.02 of the sampling-noise level (Appendix~\ref{app:ext_benign}), and exact CLEVR ambiguity raises URR by 0.28 over an answer-preserving control on Gemma 4 12B and Qwen2.5-VL-7B (Appendix~\ref{app:ext_clevr_amb}). The GPT-4o cross-modal rewrites were not human-validated, so their weak URR may partly reflect rewrites that create little ambiguity. In Figure~\ref{fig:claim2_hallucination_hcc}, question edits that can change the correct answer are not counted as hallucinations, but solarization, which can change colour answers, is (Appendix~\ref{app:res_hcc}).

\paragraph{Generation and scoring.}
The 20-token cap truncates 84--97\% of the VizWiz greedy answers of five open models, so their scores are computed on answer prefixes (Appendix~\ref{app:generation}). The length confound depends on the answer prompt: on four models, a short-answer instruction cuts Image refusals from 27--71\% to 11--43\% and removes the length advantage on Gemma 4 12B's Image items, but lowers the best Cross-modal score by 0.06--0.07 (Appendix~\ref{app:ext_prompt}); Text detection stays above 0.60 after length control on all nine VLMs (Appendix~\ref{app:ext_length}). About 80\% of Gemini 3.8 Flash's CLEVR transitions involve answers cut by its output budget; excluding them lowers its median score-change AUROC from 0.66 to 0.50 (Appendix~\ref{app:res_hcc}). Qwen2.5-VL-7B keeps answers from an earlier sampler; the shared sampler moves its other sampling-based scores by up to 0.17 AUROC but leaves Semantic Entropy its best (Appendix~\ref{app:ext_retest}). The VLM-specific methods, two in partial form, are evaluated mainly on two backbones in a separate run with their own generation settings and compared only with their own baselines; their settings were fixed before these runs, although the benchmark data had been used in earlier development (Appendix~\ref{app:vlmuq}). The CLEVR scorer and the answer-equivalence judgments are not validated against human ratings.

\paragraph{HCC.}
Per-setting intervals are pointwise, and 597 of the 1,902 settings have fewer than 10 transitions; the median AUROC of the score change (0.60), which needs no per-setting test, shows the same weak link. On initially correct pairs, the library PMI accounts for 46 of the 105 settings with an interval below zero; a text-only PMI, stored for three models, has median HCC 0.30 (Table~\ref{tab:hcc_cc}). The perturbed-input score ranks transitions better than its change (median AUROC 0.69 vs.\ 0.60), so weak HCC concerns score changes more than score levels.

\paragraph{Future directions.}
These failures motivate new directions for UQ design. Our VLM-UQBench framework, which combines annotated datasets, a scalable perturbation pipeline, and targeted evaluation metrics, provides a basis for developing and validating more robust, modality-aware UQ methods. However, perturbation strength must be carefully calibrated (Appendix~\ref{app:calibration}).

\bibliography{main}

\appendix
\raggedbottom
\setcounter{topnumber}{4}\setcounter{bottomnumber}{3}\setcounter{totalnumber}{6}\setcounter{dbltopnumber}{4}
\renewcommand{\topfraction}{0.9}\renewcommand{\bottomfraction}{0.8}\renewcommand{\textfraction}{0.07}
\renewcommand{\floatpagefraction}{0.8}\renewcommand{\dbltopfraction}{0.9}\renewcommand{\dblfloatpagefraction}{0.8}
\clearpage
{\renewcommand{\contentsname}{Appendix Contents}\tableofcontents}
\addtocontents{toc}{\protect\setcounter{tocdepth}{2}}
\par\medskip\noindent\textbf{Use of AI assistants.} AI assistants were only used to polish the writing and table format.\par
\section{Benchmark Construction}
\label{app:data_create}\label{app:data}
VLM-UQBench draws on two sources: VizWiz, whose annotations label the source of uncertainty (Appendix~\ref{app:vizwiz}), and CLEVR, whose scene graphs give the answers of the hallucination subsets (Appendix~\ref{app:clevr}).
Table~\ref{tab:datasets} lists every subset with its item count and the perturbation families applied to the clean bases (Appendix~\ref{app:augmentation}); Table~\ref{tab:uq_benchmark_main} compares VLM-UQBench with other uncertainty and ambiguity benchmarks.
Other recent benchmarks evaluate specific abilities of VLMs, such as camera movement understanding~\citep{tan2026camera}, IQ-style visual reasoning~\citep{khezresmaeilzadeh2026vriq}, engineering problems~\citep{li2025eeebench}, converting figures into vector graphics~\citep{he2026vfig} and developmentally grounded cognition~\citep{wang2026babyvlm}; VLM-UQBench instead labels where the uncertainty of an input comes from.

\paragraph{Subset names.}
We call the three uncertain VizWiz subsets \emph{Image}, \emph{Text} (invalid or subjective questions) and \emph{Cross-modal} (pairs whose image does not settle the question). Cross-modal is an operational label for two annotator reasons (an ambiguous reference or evidence absent from the image), not for every kind of image--question uncertainty.

\begin{table}[t]
\centering
\scriptsize
\setlength{\tabcolsep}{3pt}
\renewcommand{\arraystretch}{1.05}
\caption{\textbf{Benchmark composition.} Rows are evaluation subsets; \emph{What varies} names the source of uncertainty that separates the subset from its clean counterpart; \emph{Items} counts question--image pairs. The four VizWiz subsets are disjoint and cover 572 unique items; the Cross-modal file lists 150 rows of which 122 are unique, and all Cross-modal results use the unique items. Perturbation families are applied to the three clean bases at the listed number of strengths.}
\label{tab:datasets}
\resizebox{\columnwidth}{!}{%
\begin{tabular}{@{}llp{2.6cm}l@{}}
\toprule
\textbf{Subset} & \textbf{Source} & \textbf{What varies} & \textbf{Items} \\
\midrule
\multicolumn{4}{@{}l}{\textit{Human-labelled sources (VizWiz photos from blind users; five crowd reasons per item)}} \\
Clean & VizWiz & low support for selected uncertainty reasons & 150 \\
Image & VizWiz & low image quality & 150 \\
Text & VizWiz & invalid or subjective question & 150 \\
Cross-modal & VizWiz & image does not answer the question & 150 (122 unique) \\
\addlinespace
\multicolumn{4}{@{}l}{\textit{Clean bases for controlled perturbations}} \\
Clean & VQAv2 & none & 150 \\
Attr.\ / count.\ / exist.\ / rel.\ & CLEVR & none (scene-graph answer) & 150 / 150 / 150 / 150 \\
\addlinespace
\multicolumn{4}{@{}l}{\textit{Perturbation families applied to each clean base (levels in parentheses)}} \\
\multicolumn{4}{@{}p{7.2cm}@{}}{Visual (7): blur (3), brightness (2), cutout (2), gaussian noise (3), pixelate (2), salt-and-pepper (2), solarize (2)} \\
\multicolumn{4}{@{}p{7.2cm}@{}}{Text (5): drop words (2), invalid rewrite (1), subjective rewrite (1), shuffle words (2), typos (2)} \\
\multicolumn{4}{@{}p{7.2cm}@{}}{Cross-modal (3): ambiguous rewrite (1), evidence removed (question) (1), evidence removed (image) (1)} \\
\bottomrule
\end{tabular}}
\end{table}

\subsection{Source Subsets from VizWiz}
\label{app:vizwiz}
In the disagreement annotations of VizWiz~\cite{bhattacharya2019does}, five crowd annotators mark, for each question--image pair, one or more reasons for answer disagreement from a fixed taxonomy: Low Quality Image (LQI), Answer Not Present in Image (IVE), Invalid Question (INV), Difficult Question (DFF), Ambiguous Question (AMB), Subjective Question (SBJ), Synonymous Answers (SYN), Granularity Mismatch (GRN), Spam (SPM), and Other (OTH). We define the subsets from the per-reason vote counts (0--5), treating IVE, INV, DFF, AMB, SBJ and SYN as question-related reasons.
A pair is \emph{Clean} if it has at most one LQI vote and no question-related reason receives more than two votes.
It belongs to \emph{Image} (degraded image, clear question) if it has at least four LQI votes and no question-related reason receives more than two votes; to \emph{Text} (reliable image, noisy question) if it has at most one LQI vote, at least three annotators select INV or SBJ, and neither AMB nor IVE is unanimous; and to \emph{Cross-modal} (question--image mismatch) if all five annotators select AMB or all five select IVE.

On the 19,176 pairs of the VizWiz training split, these rules give disjoint candidate pools of 1,473 Clean, 842 Image, 721 Text and 3,694 Cross-modal pairs. %
Three expert annotators with research experience in vision and language then remove pairs with annotation errors or an unclear category. The instruction given to the annotators was to remove a pair if its question, image or crowd labels contain an error, or if it does not clearly show the uncertainty of its subset (a degraded image for Image, an invalid or subjective question for Text, an ambiguous reference or missing visual evidence for Cross-modal); the task involved only viewing public VizWiz images and questions and posed no foreseeable risk to the annotators. The three annotators took part as volunteers and received no payment. We collect no new data from people: VizWiz, its answer-difference annotations, VQAv2 and CLEVR are public datasets used under their terms of use, and consent for the VizWiz images and questions, which blind users submitted through a mobile app, was handled by the original dataset creators~\cite{gurari2018vizwiz}. %

Each subset has 150 rows. The Cross-modal subset contains 28 duplicates, so we report Cross-modal results on its 122 unique items; the four subsets cover 572 unique items, and no item appears in two subsets. %
The mean vote counts agree with the intended sources: 4.39 of 5 LQI votes in Image, 3.35 INV votes in Text and 4.67 AMB votes in Cross-modal. %
The votes are also concentrated within items: all 150 Image items carry at least four LQI votes (59 unanimous), 55 of the 150 Text items carry at least four INV or SBJ votes (13 unanimous), every unique Cross-modal item has one unanimous reason (AMB for 113, IVE for 9), and 130 of the 150 Clean items carry no LQI vote. %
The Cross-modal rule does not bound the wording reasons (SYN, GRN), and the Cross-modal subset differs from Clean in wording as well as in source: 84\% of its unique items (102 of 122) carry three or more SYN votes against 0\% of Clean items, whereas GRN votes are common in both (66.7\% of Clean items carry three or more). %
Appendix~\ref{app:ext_wording} tests whether this wording difference explains the Cross-modal results.

\subsection{Hallucination Subsets from CLEVR}
\label{app:clevr}
\paragraph{Construction.}
A CLEVR scene graph lists every object with its color, size, material and shape, together with every pairwise spatial relation (left, right, front, behind); we use it as the oracle.
For each of four question types, we select objects, attributes or relations of a sampled scene, fill type-specific templates and compute the answer from the scene graph alone, sampling scenes until each type reaches a fixed number of questions.
\textbf{Attribute} questions ask for an intrinsic property of one object; \textbf{existence} questions ask whether an object with given attributes is present; \textbf{relation} questions ask about spatial relationships between objects or about objects in a given direction; \textbf{counting} questions ask how many objects satisfy attribute constraints.
Spatial and geometric relations are also a focus of recent VLM evaluations~\citep{wang2025site,li2026jigshape}.
A validity check rejects ill-formed or ambiguous queries (e.g., a ``negative'' existence question must refer to an absent object type); the cases it misses are caught by the label audit below.
We count an answer as a hallucination of the question's type when its yes/no, number or attribute contradicts the scene-graph oracle.

\paragraph{Label audit.}
Rerunning the deterministic construction reproduces all 600 question--answer pairs of the benchmark. %
Checking their labels against the scene graphs flags 138 items: 21 existence answers that contradict the scene, 23 attribute and 70 relation questions whose referent is not unique, and 24 relation questions with more than one correct answer; no counting item is flagged. %
We remove the flagged items and do not change any CLEVR label. The hallucination results use the remaining \emph{scene-audited} items (150 per type minus the flagged ones): 127 attribute, 150 counting, 129 existence and 56 relation questions. %
We judge correctness with the task-aware scorer of Appendix~\ref{app:corrections}, which records an abstention as a third outcome.
\section{Experimental Protocol}
\label{app:augmentation}\label{app:perturb}
We describe how perturbed inputs are produced (Appendix~\ref{app:pipeline}), how the models decode them (Appendix~\ref{app:generation}), how the LLM UQ scores are computed (Appendix~\ref{app:scores}), how detection, URR and HCC are evaluated (Appendix~\ref{app:stats}), and how CLEVR answers are scored (Appendix~\ref{app:corrections}). VLM-specific UQ methods follow their own protocol (Appendix~\ref{app:vlmuq}).

\subsection{Perturbation Pipeline}
\label{app:pipeline}
A perturbation of family $k$ at strength $\lambda$ maps a clean image--question pair $(x_v, x_t)$ to a perturbed pair $(\tilde{x}_v, \tilde{x}_t) = T^{(k)}(x_v, x_t; \lambda)$.
Perturbations are applied to three clean bases: the VizWiz clean subset, 150 VQAv2~\citep{goyal2017making} items and the four CLEVR subsets. %
Visual perturbations change pixels, question perturbations change the question text, and cross-modal perturbations change how question and image fit together; Table~\ref{tab:datasets} lists the 15 families and their number of strength levels. %

\paragraph{Image perturbations (7 families).}\label{app:visual_pert}
Each family has one strength parameter (Table~\ref{tab:strengths}), given in parentheses: Gaussian \textbf{blur}~\cite{hendrycks2019benchmarking} (radius); zero-mean per-pixel \textbf{Gaussian noise}~\cite{hendrycks2019benchmarking} (standard deviation); \textbf{brightness} scaling, darker or brighter, following the photometric distortions of \citet{howard2013some} (factor); \textbf{pixelate}, downsampling to a coarse grid and resizing back with nearest-neighbour interpolation (grid size); \textbf{solarize}, inverting pixel values above a threshold~\cite{cubuk2019autoaugment} (threshold); \textbf{cutout}, setting a share of pixels scattered over the image to black in the spirit of Cutout~\cite{devries2017improved} (that share); and \textbf{salt-and-pepper}, setting a fraction of pixels to black or white~\cite{devries2017improved} (that fraction). %

\paragraph{Text perturbations (5 families).}\label{app:text_pert}
Two families rewrite the question with GPT-4o~\cite{hurst2024gpt} (temperature 0.7, retried until a rewrite is returned) while keeping it close to the original: %
the \textbf{invalid rewrite} makes the question slightly incoherent or under-specified (prompt: ``rewrite the question into a natural but non-visual or semantically invalid form''), and the \textbf{subjective rewrite} asks for an opinion about the same visual content (prompt: ``make the question subjective --- add mild opinion or emotion while keeping the same visual topic'', with two examples).
Three families change the surface form at random:
\textbf{typos} replace characters by neighbouring keyboard keys with a fixed probability, \textbf{drop words} delete tokens with a fixed probability, and \textbf{shuffle words} swap short phrases (split at punctuation or conjunctions) a fixed number of times.
Every perturbed question ends with a question mark and is stored alongside the original.

\paragraph{Cross-modal perturbations (3 families).}\label{app:cross_pert}
Two families rewrite the question given the image, using GPT-4o with a single system prompt that asks for a silent analysis of the relevant visible objects and a strict-JSON answer containing both variants or null.
The \textbf{ambiguous rewrite} minimally changes the question so that it can refer to several visible regions or objects (an attribute is dropped, a relation is broadened, or similar objects are merged); it is produced only when at least two visible candidates exist.
The \textbf{evidence removed (question)} rewrite asks about a closely related object or attribute that would settle the original question but is hidden, cropped out or otherwise not verifiable in the photo, for one of three reasons: a missing interaction target, an implied but absent associate, or out of frame.
The \textbf{evidence removed (image)} family keeps the question and occludes the image region it refers to.
We locate that region with question-conditioned CLIP attention~\cite{yu2024api}: each image patch's attention to the question tokens is averaged over tokens, normalised to $[0,1]$ and smoothed with a Gaussian kernel into a mask $M$, and the perturbed image is $\tilde{I} = (1 - \lambda M) \odot I$ with $\lambda = 1$, i.e.\ the raw mask. %
Related studies perturb the agreement between modalities with misaligned audio-visual pairs~\citep{chen2025modalities} or with text added to the image or the speech~\citep{chen2026typographic}; our cross-modal families instead change how the question refers to the image or remove the evidence it asks about.
These three families are the Cross-modal perturbations used throughout the paper; Appendix~\ref{app:ext_clevr_amb} adds a CLEVR control in which the ambiguity of the rewritten question is known exactly.

\paragraph{Choosing strengths.}\label{app:calibration}\label{sec:calibration}
We chose visual strengths so that artefacts are visible while content stays recognizable wherever that is the intended edit.
We inspected each candidate strength in a viewer showing a random batch of 150 images from each base next to the originals, aiming to keep objects and attributes recognizable except in the evidence-removal families; no human validity rate was measured for the perturbed items. %
Nearby strengths inside this range form the additional levels in Table~\ref{tab:datasets}.
The prompt-based rewrites have no numeric strength; their effect was tuned through the prompts and the sampling temperature.
Table~\ref{tab:strengths} lists the strength levels of every family. %
The visual levels in Table~\ref{tab:strengths} are those of the benchmark images, checked against the named operator; the question-edit levels are measured on the edited questions. %
The hallucination analysis on CLEVR (Appendix~\ref{app:res_hcc}) uses one representative strength per family: 10 for blur, 0.2 for darkening, 4 for brightening, 0.2 for cutout, 50 for Gaussian noise, 5 for pixelation, 0.2 for salt-and-pepper noise and 1 for solarization; the minor level of drop words and shuffle words and the strong level of typos; and the two rewrites. %

\begin{table}[t]
\centering
\scriptsize
\setlength{\tabcolsep}{3pt}
\renewcommand{\arraystretch}{1.05}
\caption{\textbf{Perturbation strengths.} Rows are perturbation families; the two right columns give the strength levels applied to the VizWiz clean subset and to VQAv2 and the four CLEVR subsets, whose levels are identical. Visual levels are the level labels of the benchmark images; the meaning of each label was checked on these images: on VizWiz and CLEVR, blur and pixelate are reproduced pixel for pixel and Gaussian noise to its residual standard deviation; on every base, brightness and solarize match the Pillow operator at the labelled value and cutout and salt-and-pepper change the labelled share of pixels. Darken / brighten lists the two factors. For typos and dropped words, the table gives the share of letters or words changed at the minor / strong level, measured on the edited questions (VQAv2; CLEVR after the semicolon); for shuffle words, the number of phrase swaps is not recorded.}
\label{tab:strengths}
\resizebox{\columnwidth}{!}{%
\begin{tabular}{@{}llcc@{}}
\toprule
\textbf{Family} & \textbf{Strength parameter} & \textbf{VizWiz} & \textbf{VQAv2, CLEVR} \\
\midrule
Blur & Gaussian blur radius (px) & 5, 10, 15 & 5, 10, 15 \\
Gaussian noise & std.\ of added noise (0--255) & 100, 200, 250 & 50, 100, 150 \\
Darken / brighten & brightness factor & 0.2 / 4 & 0.2 / 4 \\
Pixelate & block size (px) & 10, 15 & 5, 10 \\
Cutout & share of pixels set to black & 0.2, 0.4 & 0.2, 0.4 \\
Salt-and-pepper & share of pixels set to black or white & 0.3, 0.6 & 0.2, 0.3 \\
Solarize & inversion threshold (0--255) & 1, 2 & 1, 2 \\
Typos & share of letters substituted (measured) & 0.05 / 0.16 & 0.05 / 0.15; 0.05 / 0.15 \\
Drop words & share of words dropped (measured) & 0.17 / 0.25 & 0.18 / 0.25; 0.15--0.17 / 0.24--0.26 \\
Shuffle words & number of phrase swaps & \multicolumn{2}{c}{not recorded (minor / strong)} \\
\bottomrule
\end{tabular}}
\end{table}

\subsection{Models and Generation}
\label{app:generation}
Section~\ref{sec:setup} lists which model answers which task.

\paragraph{Decoding for LLM UQ.}
Each input receives one greedy answer and ten sampled answers, and every answer is scored on its first 20 tokens. %
Open models sample at temperature 0.7 with top-$k$ 50 and top-$p$ 1.0. %
Gemini models answer greedily at temperature 0 and sample at temperature 0.7; Gemini 3.8 Flash and Gemini 3.1 Flash-Lite use a thinking budget of 0 and a 256-token output limit, Gemini 3.1 Pro a thinking budget of 128 and a 2,048-token limit, and all Gemini answers are cut to 20 tokens before scoring. The benchmark questions ask for short answers, and explicit thinking is not always needed for perception-centred visual tasks~\citep{li2025thinkornot}. %
Qwen2.5-VL-7B (marked $^{\ddagger}$ in Table~\ref{tab:det_llmuq_full_auroc}) keeps the answers and samples of its original runs, which use an earlier sampler, also at temperature 0.7 with top-$k$ 50 and top-$p$ 1.0; rerunning it with the sampler of the other models changes its semantic entropy by at most 0.04 AUROC and its other sampling-based scores by up to 0.17 (Appendix~\ref{app:ext_retest}). %
For five open models (marked \S{} in Table~\ref{tab:det_llmuq_full_auroc}) the 20-token cap truncates most greedy answers, so their scores are computed on answer prefixes. An answer counts as truncated when it reaches 20 tokens without an end-of-sequence token; by this definition the cap cuts \detCapGemma\% of the natural VizWiz answers of Gemma 4 12B, \detCapInternVL\% of InternVL3.5-8B, \detCapMistral\% of Mistral-Small-3.2-24B, \detCapQwenThreeFive\% of Qwen3.5-9B and \detCapQwenThreeEight\% of Qwen3.8-27B. Qwen2.5-VL-7B's original runs store no tokens. %
VLM-specific UQ uses method-specific budgets (Appendix~\ref{app:native-protocol}).

\paragraph{Output budget of Gemini 3.8 Flash.}
Although thinking is switched off, Gemini 3.8 Flash still spends part of its 256-token output budget on thinking, so its visible greedy answer is truncated on 18\% of 450 clean VizWiz and CLEVR inputs and on up to 73\% at its strongest pixelation, where truncated answers reach an accuracy of 0.07 against 0.43 for complete ones. %
Its ten sampled answers, from which all four of its scores are computed, hit the cap on 19\% of clean inputs and on up to 72\% at strong pixelation. %
Gemini 3.1 Pro, with a 2,048-token budget, is barely affected: 91 of its 13,950 greedy answers in the same analysis reach the cap. %
The main benchmark run uses the same Gemini 3.8 Flash configuration and records the finish reason returned for every greedy answer. On the CLEVR pairs of Figure~\ref{fig:claim2_hallucination_hcc}, the budget ends \cTwoFlashMaxTokAnswers{} of the \cTwoFlashPairs{} perturbed answers; other Gemini 3.8 Flash results, especially on strongly degraded inputs, can also include truncated answers. Appendix~\ref{app:res_hcc} reports how many of its correct-to-wrong transitions involve such answers. %

\subsection{LLM UQ Scores}
\label{app:uq_scores}\label{app:scores}
We call the following nine scores \emph{LLM UQ}: they were developed for language models and read only the generated answers and their token probabilities. The Gemini models expose only the four sampling-based scores (semantic entropy, lexical similarity, DegMat and LUQ). Higher values mean more uncertainty throughout. Answer length, the number of words in the scored greedy answer, is reported as a control and is not a UQ score.
All nine scores use the LM-Polygraph implementations; Semantic Entropy, DegMat and LUQ judge entailment with DeBERTa-large fine-tuned on MNLI (checkpoint \texttt{microsoft/deberta-large-mnli}).

\paragraph{Token-probability scores.}
With $y = (y_1,\dots,y_T)$ the greedy answer to input $x$, \textbf{MSP} is $-\log p_\theta(y \mid x)$, the negative log-probability of the whole answer. The same log-likelihood also scores partial answers in beam search at test time~\citep{he2026beam}.
\textbf{Perplexity}~\cite{fomicheva2020unsupervised} is the negative mean log-probability of the answer tokens.
\textbf{Mean Token Entropy}~\cite{fomicheva2020unsupervised} is the entropy of the next-token distribution averaged over answer positions.
\textbf{PMI}~\cite{takayama2019relevant} is $-\frac{1}{T}\sum_{t}\big[\log p(y_t \mid y_{<t}, x) - \log p(y_t \mid y_{<t})\big]$, which is high when the input adds little to a language-model prior.
In the LM-Polygraph implementation used for the PMI rows of our tables, $\log p(y_t \mid y_{<t})$ does not come from a text-only model but from the same forward pass over the image and question, at the last prompt positions. Table~\ref{tab:hcc_cc} compares it with the text-only PMI defined above, which we stored for Gemma 4 12B, Qwen3.5-9B and Qwen3.8-27B.

\paragraph{Sampling-based scores.}
\textbf{Lexical Similarity}~\cite{fomicheva2020unsupervised} is the negative mean pairwise ROUGE-L between the sampled answers.
\textbf{Semantic Entropy}~\cite{kuhn2023semantic} groups the $N$ sampled answers into clusters by bidirectional NLI entailment (two answers share a cluster when each entails the other) and averages $-\log p(c_j \mid x)$ over the samples, where $c_j$ is the cluster of the $j$-th answer. For the open models $p(c \mid x)$ is the summed sequence probability of the answers in cluster $c$; for the Gemini models it is the fraction of the $N$ answers in $c$, so the score is the entropy of the empirical cluster distribution.
\textbf{DegMat}~\cite{lin2024generating} is $\mathrm{tr}(MI - D)/M^2$, where $D$ is the degree matrix of the $M \times M$ matrix of pairwise entailment probabilities between the $M$ sampled answers.
\textbf{LUQ}~\cite{zhang2024luq} is one minus the mean, over sampled answers, of each answer's mean similarity to the others, with similarity $\exp e / (\exp e + \exp c)$ computed from the entailment and contradiction logits $e$ and $c$ of the pair. Unlike the original method, the LM-Polygraph implementation does not split responses into sentences.

\paragraph{P(True)~\cite{kadavath2022language}.} The model is shown its own answer and asked whether it is true: $\mathcal{U}_{\mathrm{PTrue}} = -\log p_\theta(\textrm{``True''} \mid x, y)$.
The per-source detection tables of Appendix~\ref{app:results} use this answer-conditioned prompt wherever the model's answers were stored. For Qwen2.5-VL-7B, whose VizWiz answers were not stored, they use the question-only variant $-\log p_\theta(\textrm{``True''} \mid x)$, whose self-check prompt carries the question but not the answer.
With either prompt, P(True) ranks the uncertain VizWiz inputs \emph{below} the clean ones in several settings (Table~\ref{tab:det_llmuq_full_auroc}; e.g.\ 0.27 / 0.26 / 0.43 on Image / Text / Cross-modal for Qwen2.5-VL-7B with the question-only prompt and 0.29 / 0.26 / 0.33 for Gemma 4 12B with the answer-conditioned prompt); in these settings the model states \emph{higher} confidence on the degraded inputs. We keep the defined direction of P(True) everywhere. %

\subsection{Evaluation and Statistics}
\label{app:stats}
\paragraph{Per-source detection.}
For each VizWiz source (Image, Text or Cross-modal) against Clean we report the threshold-free AUROC and AUPRC. F1 needs a threshold, which we choose on training folds (five-fold nested cross-validation repeated five times, grouped by image), so F1 is never measured with a threshold tuned on the items it scores. The direction of the answer-length control is chosen on the same training folds. A table that reports a best score states how it was chosen: a fixed score; a split choice (chosen on one random stratified half of the items and scored on the other, and vice versa, averaged over 200 random splits); or an in-sample choice, marked as such. VLM-specific methods use their own published scores without selection. %

\paragraph{Uncertainty-rise rate.}
URR (Sec.~\ref{sec:evaluation}) is \emph{strict}: a tie between the clean and the perturbed score counts as no rise. A \emph{tie-corrected} variant counts a tie as one half, and the two are reported separately. Because ties push strict URR below 0.50 even when a perturbation has no effect, URR is read against a \emph{same-input sampling-noise level}. For the three sampling-based scores of Figure~\ref{fig:urr_summary} (lexical similarity, DegMat, LUQ), we recompute each score on two disjoint halves (answers 1--5 and 6--10) of the ten samples of each clean input and take the URR of the second half over the first as the reference. Each half has five answers, half the budget of the perturbation scores, so the reference measures sampling variation at half the budget; Appendix~\ref{app:res_urr} repeats the comparison against two independent ten-answer draws of the same clean input. Pooled over the six clean sets (the VizWiz clean subset, the VQAv2 items and the four CLEVR subsets; 900 images, 2,700 pairs per model), the strict reference lies between 0.45 and 0.48 for the four core models, and between 0.49 and 0.51 when a tie counts one half; per set (150 images, 450 pairs) the strict reference ranges from 0.35 to 0.52. %

\paragraph{HCC.}
On CLEVR, $\Delta U_i = u(x_i^{\mathrm{pert}}) - u(x_i)$, and $H_i = 1$ marks a correct-to-wrong transition, the operational hallucination label of Sec.~\ref{sec:evaluation}: a correct clean answer whose perturbed answer is incorrect under the task-aware scorer (Appendix~\ref{app:corrections}). HCC (Sec.~\ref{sec:evaluation}) is the point-biserial correlation between $\Delta U$ and $H$ over the scene-audited pairs whose perturbed answer is not an abstention; initially incorrect items stay in with $H_i = 0$. For a fixed VLM, UQ method and perturbation setting,
\begin{equation}
\begin{aligned}
\mathrm{HCC}(\Delta u) &= \operatorname{Corr}(\Delta U, H) \\
&= \frac{\overline{\Delta U}_1 - \overline{\Delta U}_0}{s_{\Delta U}}
\sqrt{\frac{n_1 n_0}{n^2}},
\end{aligned}
\end{equation}
where $\overline{\Delta U}_1$ and $\overline{\Delta U}_0$ are the mean score changes of pairs with and without a correct-to-wrong transition, respectively, $n_1$ and $n_0$ their counts, $n = n_1 + n_0$, and $s_{\Delta U}$ the population standard deviation of the score changes over the included pairs. A positive HCC means a larger mean score change for $H=1$ than for $H=0$; it is a correlation, not the share of transitions a score would flag.

\paragraph{Intervals.}
AUROC, URR and HCC use 95\% intervals from 2,000 bootstrap draws that resample images rather than rows, keeping all rows of a sampled image together; AUROC half-widths are 0.04--0.07 on the VizWiz sources. All intervals are pointwise, and a difference between two methods or models is a point estimate unless a paired interval is reported. Holm correction is applied only where an analysis says so. The VLM-specific analyses report conditional pointwise intervals (Appendix~\ref{app:vlmuq}). %

\subsection{Answer Scoring on CLEVR}
\label{app:changes}\label{app:corrections}
Free-form model outputs are hard to evaluate for automatic metrics, human preferences and LLM judges alike~\citep{li2025rationales}. Judging correctness by whether the gold string occurs inside the answer fails on number words and on sentences containing other digits, credits ``no'' inside ``not'', and counts abstentions as errors. We therefore use a \emph{task-aware scorer} that maps number words to digits, matches whole words, reads a leading yes/no and records an abstention as a third outcome. The choice matters: with substring matching, Qwen2.5-VL-7B's clean counting accuracy would be 12.7\% instead of 98.7\%. We have not validated the task-aware scorer against human ratings (see Limitations). %
\section{Evidence for the Three Findings}
\label{app:results}
This appendix gives the complete results behind the three findings of the main text: detection depends on both the source and the model (\S\ref{app:res_detection}), uncertainty responds unevenly to perturbations and is weakly consistent with the hallucinations they induce (\S\ref{app:res_urr_hcc}), and part of the detection signal is explained by answer length (\S\ref{app:ext_length}).
\ifdefined\detVarNineModel\else\newcommand{\detBestFlashCross}{0.75}
\newcommand{\detBestFlashImage}{0.82}
\newcommand{\detBestFlashLiteCross}{0.72}
\newcommand{\detBestFlashLiteImage}{0.73}
\newcommand{\detBestFlashLiteText}{0.80}
\newcommand{\detBestFlashText}{0.85}
\newcommand{\detBestGemmaCross}{0.81}
\newcommand{\detBestGemmaImage}{0.77}
\newcommand{\detBestGemmaText}{0.84}
\newcommand{\detBestInternVLCross}{0.77}
\newcommand{\detBestInternVLImage}{0.83}
\newcommand{\detBestInternVLText}{0.85}
\newcommand{\detBestIsSECount}{16}
\newcommand{\detBestMistralCross}{0.81}
\newcommand{\detBestMistralImage}{0.87}
\newcommand{\detBestMistralText}{0.87}
\newcommand{\detBestProCross}{0.71}
\newcommand{\detBestProImage}{0.73}
\newcommand{\detBestProText}{0.87}
\newcommand{\detBestQwenThreeEightCross}{0.67}
\newcommand{\detBestQwenThreeEightImage}{0.76}
\newcommand{\detBestQwenThreeEightText}{0.83}
\newcommand{\detBestQwenThreeFiveCross}{0.73}
\newcommand{\detBestQwenThreeFiveImage}{0.82}
\newcommand{\detBestQwenThreeFiveText}{0.86}
\newcommand{\detBestQwenVLCross}{0.82}
\newcommand{\detBestQwenVLImage}{0.85}
\newcommand{\detBestQwenVLText}{0.89}
\newcommand{\detCFApiCells}{9}
\newcommand{\detCFApiPool}{4}
\newcommand{\detCFCrossLowestCount}{8}
\newcommand{\detCFCrossLowestExceptions}{Gemma 4 12B (Image lowest)}
\newcommand{\detCFCrossLowestNearTieCount}{1}
\newcommand{\detCFCrossLowestNearTieGap}{0.015}
\newcommand{\detCFFlashCross}{0.74}
\newcommand{\detCFFlashImage}{0.81}
\newcommand{\detCFFlashLiteCross}{0.71}
\newcommand{\detCFFlashLiteImage}{0.72}
\newcommand{\detCFFlashLiteText}{0.79}
\newcommand{\detCFFlashText}{0.85}
\newcommand{\detCFGemmaCross}{0.80}
\newcommand{\detCFGemmaImage}{0.75}
\newcommand{\detCFGemmaText}{0.84}
\newcommand{\detCFInternVLCross}{0.76}
\newcommand{\detCFInternVLImage}{0.81}
\newcommand{\detCFInternVLText}{0.84}
\newcommand{\detCFMaxCross}{0.82}
\newcommand{\detCFMaxCrossModel}{Qwen2.5-VL-7B}
\newcommand{\detCFMaxImage}{0.86}
\newcommand{\detCFMaxImageModel}{Mistral-Small-3.2-24B}
\newcommand{\detCFMaxText}{0.89}
\newcommand{\detCFMaxTextModel}{Qwen2.5-VL-7B}
\newcommand{\detCFMeanCross}{0.75}
\newcommand{\detCFMeanImage}{0.78}
\newcommand{\detCFMeanText}{0.85}
\newcommand{\detCFMinCross}{0.65}
\newcommand{\detCFMinCrossModel}{Qwen3.8-27B}
\newcommand{\detCFMinImage}{0.72}
\newcommand{\detCFMinImageModel}{Gemini 3.1 Flash-Lite}
\newcommand{\detCFMinText}{0.79}
\newcommand{\detCFMinTextModel}{Gemini 3.1 Flash-Lite}
\newcommand{\detCFMistralCross}{0.81}
\newcommand{\detCFMistralImage}{0.86}
\newcommand{\detCFMistralText}{0.87}
\newcommand{\detCFModalLSApiCount}{7}
\newcommand{\detCFModalLSOpenCount}{0}
\newcommand{\detCFModalSEApiCount}{1}
\newcommand{\detCFModalSECount}{16}
\newcommand{\detCFModalSEOpenCount}{15}
\newcommand{\detCFOpenCells}{18}
\newcommand{\detCFOpenPool}{9}
\newcommand{\detCFProCross}{0.71}
\newcommand{\detCFProImage}{0.73}
\newcommand{\detCFProText}{0.87}
\newcommand{\detCFQwenThreeEightCross}{0.65}
\newcommand{\detCFQwenThreeEightImage}{0.74}
\newcommand{\detCFQwenThreeEightText}{0.81}
\newcommand{\detCFQwenThreeFiveCross}{0.71}
\newcommand{\detCFQwenThreeFiveImage}{0.82}
\newcommand{\detCFQwenThreeFiveText}{0.85}
\newcommand{\detCFQwenVLCross}{0.82}
\newcommand{\detCFQwenVLImage}{0.84}
\newcommand{\detCFQwenVLText}{0.89}
\newcommand{\detCFSourceSpreadMax}{0.16}
\newcommand{\detCFSourceSpreadMaxModel}{Gemini 3.1 Pro}
\newcommand{\detCFSourceSpreadMin}{0.06}
\newcommand{\detCFSpreadCross}{0.16}
\newcommand{\detCFSpreadImage}{0.14}
\newcommand{\detCFSpreadText}{0.09}
\newcommand{\detCFTextHighestCount}{9}
\newcommand{\detCFTextHighestExceptions}{none}
\newcommand{\detCFminusSEMax}{0.08}
\newcommand{\detCFminusSEMin}{-0.02}
\newcommand{\detCapGemma}{84.2}
\newcommand{\detCapInternVL}{84.8}
\newcommand{\detCapMax}{96.5}
\newcommand{\detCapMin}{84.2}
\newcommand{\detCapMistral}{93.5}
\newcommand{\detCapQwenThreeEight}{95.7}
\newcommand{\detCapQwenThreeFive}{96.5}
\newcommand{\detLenAboveBestCells}{Gemma 4 12B Image, Gemma 4 12B Text, Qwen3.5-9B Cross-modal, Gemini 3.1 Pro Image, Gemini 3.1 Flash-Lite Image}
\newcommand{\detLenAboveBestCount}{5}
\newcommand{\detLenAboveSECount}{8}
\newcommand{\detLenCells}{24}
\newcommand{\detLenFlashCross}{0.72}
\newcommand{\detLenFlashImage}{0.76}
\newcommand{\detLenFlashLiteCross}{0.72}
\newcommand{\detLenFlashLiteImage}{0.84}
\newcommand{\detLenFlashLiteText}{0.80}
\newcommand{\detLenFlashText}{0.71}
\newcommand{\detLenGemmaCross}{0.75}
\newcommand{\detLenGemmaImage}{0.87}
\newcommand{\detLenGemmaText}{0.86}
\newcommand{\detLenInternVLCross}{0.70}
\newcommand{\detLenInternVLImage}{0.50}
\newcommand{\detLenInternVLText}{0.65}
\newcommand{\detLenMinusBestFlashLiteImage}{0.11}
\newcommand{\detLenMinusBestFlashLiteImageHi}{0.17}
\newcommand{\detLenMinusBestFlashLiteImageLo}{0.05}
\newcommand{\detLenMinusSEGemmaImage}{0.10}
\newcommand{\detLenMinusSEGemmaImageHi}{0.16}
\newcommand{\detLenMinusSEGemmaImageLo}{0.04}
\newcommand{\detLenMinusSEProImage}{0.12}
\newcommand{\detLenMinusSEProImageHi}{0.19}
\newcommand{\detLenMinusSEProImageLo}{0.04}
\newcommand{\detLenMistralCross}{0.69}
\newcommand{\detLenMistralImage}{0.65}
\newcommand{\detLenMistralText}{0.67}
\newcommand{\detLenProCross}{0.66}
\newcommand{\detLenProImage}{0.78}
\newcommand{\detLenProText}{0.70}
\newcommand{\detLenQwenThreeEightCross}{0.65}
\newcommand{\detLenQwenThreeEightImage}{0.74}
\newcommand{\detLenQwenThreeEightText}{0.76}
\newcommand{\detLenQwenThreeFiveCross}{0.74}
\newcommand{\detLenQwenThreeFiveImage}{0.71}
\newcommand{\detLenQwenThreeFiveText}{0.76}
\newcommand{\detNApi}{3}
\newcommand{\detNCells}{27}
\newcommand{\detNClean}{150}
\newcommand{\detNModels}{9}
\newcommand{\detNOpen}{6}
\newcommand{\detNPosCross}{122}
\newcommand{\detNPosImage}{150}
\newcommand{\detNPosText}{150}
\newcommand{\detOptimismMax}{0.02}
\newcommand{\detOptimismMedian}{0.01}
\newcommand{\detQwenPipeOtherMax}{0.17}
\newcommand{\detQwenPipeSEMax}{0.04}
\newcommand{\detSEAboveChanceCount}{27}
\newcommand{\detSECILowMin}{0.59}
\newcommand{\detSECrossLowestCount}{8}
\newcommand{\detSECrossLowestExceptions}{Gemma 4 12B (Image lowest)}
\newcommand{\detSECrossLowestNearTieCount}{2}
\newcommand{\detSECrossLowestNearTieGap}{0.015}
\newcommand{\detSEFlashCross}{0.75}
\newcommand{\detSEFlashImage}{0.81}
\newcommand{\detSEFlashLiteCross}{0.71}
\newcommand{\detSEFlashLiteImage}{0.72}
\newcommand{\detSEFlashLiteText}{0.76}
\newcommand{\detSEFlashText}{0.83}
\newcommand{\detSEGemmaCross}{0.81}
\newcommand{\detSEGemmaImage}{0.77}
\newcommand{\detSEGemmaText}{0.84}
\newcommand{\detSEInternVLCross}{0.77}
\newcommand{\detSEInternVLImage}{0.82}
\newcommand{\detSEInternVLText}{0.85}
\newcommand{\detSEMaxCross}{0.81}
\newcommand{\detSEMaxCrossModel}{Mistral-Small-3.2-24B}
\newcommand{\detSEMaxImage}{0.87}
\newcommand{\detSEMaxImageModel}{Mistral-Small-3.2-24B}
\newcommand{\detSEMaxText}{0.89}
\newcommand{\detSEMaxTextModel}{Qwen2.5-VL-7B}
\newcommand{\detSEMeanCross}{0.74}
\newcommand{\detSEMeanImage}{0.79}
\newcommand{\detSEMeanText}{0.84}
\newcommand{\detSEMinCross}{0.66}
\newcommand{\detSEMinCrossModel}{Gemini 3.1 Pro}
\newcommand{\detSEMinImage}{0.67}
\newcommand{\detSEMinImageModel}{Gemini 3.1 Pro}
\newcommand{\detSEMinText}{0.76}
\newcommand{\detSEMinTextModel}{Gemini 3.1 Flash-Lite}
\newcommand{\detSEMistralCross}{0.81}
\newcommand{\detSEMistralImage}{0.87}
\newcommand{\detSEMistralText}{0.87}
\newcommand{\detSEProCross}{0.66}
\newcommand{\detSEProImage}{0.67}
\newcommand{\detSEProText}{0.79}
\newcommand{\detSEQwenThreeEightCross}{0.67}
\newcommand{\detSEQwenThreeEightImage}{0.76}
\newcommand{\detSEQwenThreeEightText}{0.83}
\newcommand{\detSEQwenThreeFiveCross}{0.73}
\newcommand{\detSEQwenThreeFiveImage}{0.82}
\newcommand{\detSEQwenThreeFiveText}{0.86}
\newcommand{\detSEQwenVLCross}{0.80}
\newcommand{\detSEQwenVLImage}{0.85}
\newcommand{\detSEQwenVLText}{0.89}
\newcommand{\detSESourceSpreadMax}{0.16}
\newcommand{\detSESourceSpreadMaxModel}{Qwen3.8-27B}
\newcommand{\detSESourceSpreadMin}{0.05}
\newcommand{\detSESpreadCross}{0.15}
\newcommand{\detSESpreadCrossApi}{0.09}
\newcommand{\detSESpreadCrossOpen}{0.15}
\newcommand{\detSESpreadImage}{0.20}
\newcommand{\detSESpreadImageApi}{0.15}
\newcommand{\detSESpreadImageOpen}{0.12}
\newcommand{\detSESpreadText}{0.13}
\newcommand{\detSESpreadTextApi}{0.07}
\newcommand{\detSESpreadTextOpen}{0.07}
\newcommand{\detSETextHighestCount}{8}
\newcommand{\detSETextHighestExceptions}{Mistral-Small-3.2-24B (Image highest)}
\newcommand{\detSFourAboveChanceCount}{27}
\newcommand{\detSFourFlashCross}{0.69}
\newcommand{\detSFourFlashImage}{0.77}
\newcommand{\detSFourFlashLiteCross}{0.68}
\newcommand{\detSFourFlashLiteImage}{0.71}
\newcommand{\detSFourFlashLiteText}{0.75}
\newcommand{\detSFourFlashText}{0.80}
\newcommand{\detSFourGemmaCross}{0.73}
\newcommand{\detSFourGemmaImage}{0.70}
\newcommand{\detSFourGemmaText}{0.78}
\newcommand{\detSFourInternVLCross}{0.70}
\newcommand{\detSFourInternVLImage}{0.82}
\newcommand{\detSFourInternVLText}{0.79}
\newcommand{\detSFourMaxCross}{0.75}
\newcommand{\detSFourMaxCrossModel}{Mistral-Small-3.2-24B}
\newcommand{\detSFourMaxImage}{0.83}
\newcommand{\detSFourMaxImageModel}{Mistral-Small-3.2-24B}
\newcommand{\detSFourMaxText}{0.83}
\newcommand{\detSFourMaxTextModel}{Mistral-Small-3.2-24B}
\newcommand{\detSFourMeanCross}{0.69}
\newcommand{\detSFourMeanImage}{0.75}
\newcommand{\detSFourMeanText}{0.79}
\newcommand{\detSFourMinCross}{0.65}
\newcommand{\detSFourMinCrossModel}{Gemini 3.1 Pro}
\newcommand{\detSFourMinImage}{0.67}
\newcommand{\detSFourMinImageModel}{Gemini 3.1 Pro}
\newcommand{\detSFourMinText}{0.75}
\newcommand{\detSFourMinTextModel}{Gemini 3.1 Flash-Lite}
\newcommand{\detSFourMistralCross}{0.75}
\newcommand{\detSFourMistralImage}{0.83}
\newcommand{\detSFourMistralText}{0.83}
\newcommand{\detSFourProCross}{0.65}
\newcommand{\detSFourProImage}{0.67}
\newcommand{\detSFourProText}{0.80}
\newcommand{\detSFourQwenThreeEightCross}{0.66}
\newcommand{\detSFourQwenThreeEightImage}{0.73}
\newcommand{\detSFourQwenThreeEightText}{0.80}
\newcommand{\detSFourQwenThreeFiveCross}{0.69}
\newcommand{\detSFourQwenThreeFiveImage}{0.77}
\newcommand{\detSFourQwenThreeFiveText}{0.81}
\newcommand{\detSFourQwenVLCross}{0.67}
\newcommand{\detSFourQwenVLImage}{0.76}
\newcommand{\detSFourQwenVLText}{0.77}
\newcommand{\detSFourSpreadCross}{0.10}
\newcommand{\detSFourSpreadImage}{0.16}
\newcommand{\detSFourSpreadText}{0.07}
\newcommand{\detVarNineLomoModelMax}{28}
\newcommand{\detVarNineLomoModelMin}{15}
\newcommand{\detVarNineLomoSourceMax}{65}
\newcommand{\detVarNineLomoSourceMin}{53}
\newcommand{\detVarNineModel}{25}
\newcommand{\detVarNineModelHi}{47}
\newcommand{\detVarNineModelLo}{14}
\newcommand{\detVarNineModelP}{0.04}
\newcommand{\detVarNineQwenRerunModel}{29}
\newcommand{\detVarNineQwenRerunModelP}{0.02}
\newcommand{\detVarNineQwenRerunResidual}{17}
\newcommand{\detVarNineQwenRerunSource}{54}
\newcommand{\detVarNineQwenRerunSourceP}{$<$0.001}
\newcommand{\detVarNineResidual}{18}
\newcommand{\detVarNineResidualHi}{31}
\newcommand{\detVarNineResidualLo}{12}
\newcommand{\detVarNineSEModel}{55}
\newcommand{\detVarNineSEModelHi}{71}
\newcommand{\detVarNineSEModelLo}{36}
\newcommand{\detVarNineSEModelP}{$<$0.001}
\newcommand{\detVarNineSEResidual}{12}
\newcommand{\detVarNineSEResidualHi}{23}
\newcommand{\detVarNineSEResidualLo}{8}
\newcommand{\detVarNineSESource}{33}
\newcommand{\detVarNineSESourceHi}{49}
\newcommand{\detVarNineSESourceLo}{15}
\newcommand{\detVarNineSESourceP}{$<$0.001}
\newcommand{\detVarNineSource}{57}
\newcommand{\detVarNineSourceHi}{68}
\newcommand{\detVarNineSourceLo}{32}
\newcommand{\detVarNineSourceP}{$<$0.001}
\newcommand{\detVarSixModel}{27}
\newcommand{\detVarSixModelHi}{49}
\newcommand{\detVarSixModelLo}{13}
\newcommand{\detVarSixModelP}{0.11}
\newcommand{\detVarSixResidual}{23}
\newcommand{\detVarSixResidualHi}{38}
\newcommand{\detVarSixResidualLo}{13}
\newcommand{\detVarSixSource}{50}
\newcommand{\detVarSixSourceHi}{66}
\newcommand{\detVarSixSourceLo}{27}
\newcommand{\detVarSixSourceP}{0.003}
\fi
\providecommand{\appvTwoUrrFamDsMin}{2}
\providecommand{\appvTwoUrrFamDsMax}{6}
\providecommand{\appvTwoUrrFamAbove}{194}
\providecommand{\appvTwoUrrFamCells}{328}
\providecommand{\appvTwoHallImgMin}{0.0}
\providecommand{\appvTwoHallImgMax}{71.7}
\providecommand{\appvTwoUrrMapAbove}{244}
\providecommand{\appvTwoUrrMapCells}{648}
\providecommand{\appvTwoTransferFamQwen}{9078}
\providecommand{\appvTwoTransferCrossQwen}{4278}
\providecommand{\appvTwoTransferFamLlava}{9078}
\providecommand{\appvTwoTransferCrossLlava}{4278}

\subsection{Source and Model Dependence}
\label{app:res_detection}
\definecolor{ctrlabove}{RGB}{0,115,90}\definecolor{ctrlbelow}{RGB}{200,50,0}\definecolor{vlmbg}{RGB}{229,240,250}
\begin{table}[t]
\centering
\footnotesize
\setlength{\tabcolsep}{3pt}
\renewcommand{\arraystretch}{1.0}
\caption{\textbf{Detecting each uncertainty source with UQ on Gemma 4 12B and Gemini 3.1 Pro (AUROC$\uparrow$; chance 0.50).} Same layout as Table~\ref{tab:det_llmuq_main}. Each column compares the VizWiz items of one source (Image, Text, Cross-modal) with the same clean items. Shaded rows: VLM-specific UQ; unshaded: LLM UQ. Indented rows are baselines of the VLM-specific method above them; green or red marks a method above or below its baseline(s). ``--'': not applicable (no white-box access to closed-source models). Bold: highest value of the column within each block. $^{\ast}$Question-only P(True) prompt.}
\label{tab:det_merged_app}
\resizebox{\ifdim\width>\columnwidth\columnwidth\else\width\fi}{!}{%
\begin{tabular}{lcccccc}
\toprule
 & \multicolumn{3}{c}{\textbf{Open-source}} & \multicolumn{3}{c}{\textbf{Closed-source}} \\
\cmidrule(lr){2-4} \cmidrule(lr){5-7}
 & \multicolumn{3}{c}{Gemma 4 12B} & \multicolumn{3}{c}{Gemini 3.1 Pro} \\
\cmidrule(lr){2-4} \cmidrule(lr){5-7}
\textbf{Method} & Image & Text & Cross & Image & Text & Cross \\
\midrule
\rowcolor{black!6}\multicolumn{7}{l}{\textbf{Black-box: sampled answers}} \\
\multicolumn{7}{l}{\textit{LLM UQ}} \\
Semantic Entropy & \textbf{0.77} & \textbf{0.84} & \textbf{0.81} & 0.67 & 0.79 & 0.66 \\
Lexical Similarity & 0.74 & 0.81 & 0.74 & \textbf{0.73} & \textbf{0.87} & \textbf{0.71} \\
DegMat & 0.69 & 0.79 & 0.73 & 0.68 & 0.80 & 0.65 \\
LUQ & 0.59 & 0.68 & 0.64 & 0.62 & 0.75 & 0.58 \\
\addlinespace[2pt]
\rowcolor{vlmbg}\multicolumn{7}{l}{\textit{VLM-specific UQ}} \\
\rowcolor{vlmbg}VL-Uncertainty & 0.48 & 0.59 & 0.64 & \textbf{0.64} & 0.74 & 0.65 \\
\rowcolor{vlmbg}\quad SE, same sampler & \textbf{0.50} & \textbf{0.63} & \textbf{0.70} & 0.63 & \textbf{0.77} & \textbf{0.69} \\
\midrule
\rowcolor{black!6}\multicolumn{7}{l}{\textbf{White-box: logits, attention or hidden states}} \\
\multicolumn{7}{l}{\textit{LLM UQ}} \\
MSP & \textbf{0.75} & \textbf{0.79} & 0.76 & {\color{black!35}--} & {\color{black!35}--} & {\color{black!35}--} \\
Perplexity & 0.70 & 0.76 & 0.74 & {\color{black!35}--} & {\color{black!35}--} & {\color{black!35}--} \\
Mean token entropy & 0.74 & \textbf{0.79} & \textbf{0.79} & {\color{black!35}--} & {\color{black!35}--} & {\color{black!35}--} \\
PMI & 0.72 & 0.63 & 0.65 & {\color{black!35}--} & {\color{black!35}--} & {\color{black!35}--} \\
P(True) & 0.29 & 0.26 & 0.33 & {\color{black!35}--} & {\color{black!35}--} & {\color{black!35}--} \\
\addlinespace[2pt]
\rowcolor{vlmbg}\multicolumn{7}{l}{\textit{VLM-specific UQ}} \\
\rowcolor{vlmbg}VAUQ & 0.87 & \textbf{0.89} & 0.78 & {\color{black!35}--} & {\color{black!35}--} & {\color{black!35}--} \\
\rowcolor{vlmbg}\quad unmasked entropy & 0.79 & 0.80 & 0.80 & {\color{black!35}--} & {\color{black!35}--} & {\color{black!35}--} \\
\rowcolor{vlmbg}\quad random mask & 0.85 & 0.88 & 0.78 & {\color{black!35}--} & {\color{black!35}--} & {\color{black!35}--} \\
\rowcolor{vlmbg}Response SVAR & 0.83 & 0.84 & 0.71 & {\color{black!35}--} & {\color{black!35}--} & {\color{black!35}--} \\
\rowcolor{vlmbg}Dropout Dec. (token entropy) & 0.64 & 0.62 & 0.68 & {\color{black!35}--} & {\color{black!35}--} & {\color{black!35}--} \\
\rowcolor{vlmbg}Dropout Dec. (token disagreement) & 0.32 & 0.33 & 0.32 & {\color{black!35}--} & {\color{black!35}--} & {\color{black!35}--} \\
\rowcolor{vlmbg}BICR & \textbf{0.87} & 0.87 & \textbf{0.80} & {\color{black!35}--} & {\color{black!35}--} & {\color{black!35}--} \\
\rowcolor{vlmbg}\quad no ranking term & 0.83 & 0.85 & 0.80 & {\color{black!35}--} & {\color{black!35}--} & {\color{black!35}--} \\
\bottomrule
\end{tabular}}
\end{table}
Table~\ref{tab:det_merged_app} repeats Table~\ref{tab:det_llmuq_main} for Gemma 4 12B and Gemini 3.1 Pro. For all nine VLMs, Table~\ref{tab:det_llmuq_full_auroc} reports the AUROC of every LLM UQ score for separating each VizWiz source from the clean items. It also reports answer length as a control, the mean of the four sampling-based scores and the best score. Table~\ref{tab:det_llmuq_ci} gives 95\% intervals for the AUROCs of Semantic Entropy, answer length and the four-score mean.
By point estimate, Semantic Entropy ranks Cross-modal as the hardest source for \detSECrossLowestCount{} of the \detNModels{} VLMs. For \detSECrossLowestNearTieCount{} of them, Cross-modal lies within \detSECrossLowestNearTieGap{} AUROC of the next source.

\paragraph{Model versus source.} Table~\ref{tab:det_llmuq_variance} splits the variation of the AUROCs into a share due to the model and a share due to the source. Which share is larger depends on the score. For the mean of the four sampling-based scores, the source share is larger: \detVarNineSource\% against \detVarNineModel\% over all nine VLMs, and \detVarSixSource\% against \detVarSixModel\% over the six VLMs other than Qwen3.5-9B, Qwen3.8-27B and Gemini 3.1 Flash-Lite. For Semantic Entropy alone, the model share is larger (\detVarNineSEModel\% vs.\ \detVarNineSESource\%). The bootstrap intervals of the two shares overlap in both comparisons.
Differences between VLMs can come from their language backbones, which also limit, for example, hierarchical visual recognition~\citep{tan2026llmbottleneck}. Part of the model share, however, reflects how the scores are computed rather than the model itself. Open models compute Semantic Entropy from summed sequence probabilities and closed-source models from sample frequencies (Appendix~\ref{app:scores}), and the best score is chosen from \detCFOpenPool{} scores for open models but from \detCFApiPool{} for closed-source models. Within each group, the AUROC of Semantic Entropy on Image spans \detSESpreadImageOpen{} across the six open models and \detSESpreadImageApi{} across the three closed-source models (\detSESpreadImage{} over all nine). Semantic Entropy is the most frequent best score in \detCFModalSEOpenCount{} of the \detCFOpenCells{} open-model settings, but in only \detCFModalSEApiCount{} of the \detCFApiCells{} closed-source settings, where Lexical Similarity is most frequent (\detCFModalLSApiCount{}). Detection therefore depends on both the source and the model, and neither dominates.

\begin{table*}[t]
\centering
\scriptsize
\setlength{\tabcolsep}{3.4pt}
\renewcommand{\arraystretch}{1.0}
\caption{\textbf{All LLM UQ scores for detecting each uncertainty source (AUROC$\uparrow$; chance 0.50).} Same task and items as Table~\ref{tab:det_llmuq_main}, for all nine VLMs. Scores: maximum sequence probability (MSP), perplexity (PPL), mean token entropy (MTE), pointwise mutual information (PMI), P(True) (P(T)), semantic entropy (SE), lexical similarity (LS), degree matrix (DM) and LUQ; the closed-source models expose only the four sampling-based scores (``--''). Higher scores mean more uncertain for every score; the direction is fixed in advance, so a value below 0.50 means the score ranks the source \emph{below} the clean items. Bold: best LLM UQ score of the row chosen on all items (in-sample). P(True) uses the answer-conditioned self-check prompt, except $^{\ast}$the question-only prompt for Qwen2.5-VL-7B, whose VizWiz answers were not stored. $^{\dagger}$One Mistral-Small-3.2-24B image has no PMI value (149 + 150 items). \emph{Len.}: answer length alone (control, not a UQ score; no stored answers for Qwen2.5-VL-7B; greedy answers truncated at 20 tokens, which most answers of the models marked $^{\S}$ reach). \emph{Samp.\ mean}: mean of the four sampling-based scores (SE, LS, DM, LUQ), the summary every model exposes. \emph{Best score}: best LLM UQ score of the row, chosen on half of the items and evaluated on the other half; choosing the score on all items instead changes an entry by at most 0.02 AUROC. $^{\ddagger}$Answers and samples from the original runs; $^{\S}$most greedy answers reach the 20-token cap (Appendix~\ref{app:generation}).}
\label{tab:det_llmuq_full_auroc}
\resizebox{\ifdim\width>\textwidth\textwidth\else\width\fi}{!}{%
\begin{tabular}{@{}llccccccccc@{\hspace{9pt}}c@{\hspace{9pt}}cc@{}}
\toprule
\textbf{Model} & \textbf{Source} & MSP & PPL & MTE & PMI & P(T) & SE & LS & DM & LUQ & Len. & Samp.\ mean & Best score \\
\midrule
\multicolumn{14}{@{}l}{\emph{Open-weight VLMs}} \\
\multirow{3}{*}{Qwen2.5-VL-7B$^{\ddagger}$} & Image & 0.84 & 0.80 & 0.81 & 0.79 & 0.27$^{\ast}$ & \textbf{0.85} & \textbf{0.85} & 0.73 & 0.61 & -- & 0.76 & 0.84 \\
 & Text & 0.88 & 0.81 & 0.81 & 0.82 & 0.26$^{\ast}$ & \textbf{0.89} & 0.87 & 0.70 & 0.63 & -- & 0.77 & 0.89 \\
 & Cross-modal & \textbf{0.82} & 0.77 & 0.78 & 0.78 & 0.43$^{\ast}$ & 0.80 & 0.76 & 0.60 & 0.54 & -- & 0.67 & 0.82 \\
\addlinespace[2pt]
\multirow{3}{*}{Gemma 4 12B$^{\S}$} & Image & 0.75 & 0.70 & 0.74 & 0.72 & 0.29 & \textbf{0.77} & 0.74 & 0.69 & 0.59 & 0.87 & 0.70 & 0.75 \\
 & Text & 0.79 & 0.76 & 0.79 & 0.63 & 0.26 & \textbf{0.84} & 0.81 & 0.79 & 0.68 & 0.86 & 0.78 & 0.84 \\
 & Cross-modal & 0.76 & 0.74 & 0.79 & 0.65 & 0.33 & \textbf{0.81} & 0.74 & 0.73 & 0.64 & 0.75 & 0.73 & 0.80 \\
\addlinespace[2pt]
\multirow{3}{*}{InternVL3.5-8B$^{\S}$} & Image & 0.79 & 0.81 & \textbf{0.83} & 0.59 & 0.56 & 0.82 & 0.81 & 0.82 & 0.82 & 0.50 & 0.82 & 0.81 \\
 & Text & 0.83 & 0.82 & 0.83 & 0.56 & 0.45 & \textbf{0.85} & 0.80 & 0.78 & 0.73 & 0.65 & 0.79 & 0.84 \\
 & Cross-modal & \textbf{0.77} & 0.76 & 0.76 & 0.46 & 0.54 & \textbf{0.77} & 0.70 & 0.70 & 0.63 & 0.70 & 0.70 & 0.76 \\
\addlinespace[2pt]
\multirow{3}{*}{Mistral-Small-3.2-24B$^{\S}$} & Image & 0.83 & 0.84 & 0.86 & 0.64$^{\dagger}$ & 0.40 & \textbf{0.87} & 0.84 & 0.82 & 0.79 & 0.65 & 0.83 & 0.86 \\
 & Text & 0.83 & 0.82 & 0.84 & 0.72 & 0.39 & \textbf{0.87} & 0.84 & 0.83 & 0.77 & 0.67 & 0.83 & 0.87 \\
 & Cross-modal & 0.77 & 0.78 & 0.78 & 0.57 & 0.39 & \textbf{0.81} & 0.75 & 0.75 & 0.67 & 0.69 & 0.75 & 0.81 \\
\addlinespace[2pt]
\multirow{3}{*}{Qwen3.5-9B$^{\S}$} & Image & 0.77 & 0.77 & 0.79 & 0.71 & 0.58 & \textbf{0.82} & 0.79 & 0.74 & 0.74 & 0.71 & 0.77 & 0.82 \\
 & Text & 0.80 & 0.79 & 0.81 & 0.59 & 0.77 & \textbf{0.86} & 0.84 & 0.77 & 0.76 & 0.76 & 0.81 & 0.85 \\
 & Cross-modal & 0.66 & 0.64 & 0.63 & 0.51 & 0.62 & \textbf{0.73} & 0.68 & 0.70 & 0.67 & 0.74 & 0.69 & 0.71 \\
\addlinespace[2pt]
\multirow{3}{*}{Qwen3.8-27B$^{\S}$} & Image & 0.72 & 0.72 & 0.74 & 0.61 & 0.65 & \textbf{0.76} & 0.72 & 0.71 & 0.72 & 0.74 & 0.73 & 0.74 \\
 & Text & 0.79 & 0.78 & 0.80 & 0.60 & 0.51 & \textbf{0.83} & 0.81 & 0.80 & 0.76 & 0.76 & 0.80 & 0.81 \\
 & Cross-modal & 0.62 & 0.60 & 0.62 & 0.57 & 0.45 & \textbf{0.67} & 0.66 & \textbf{0.67} & 0.63 & 0.65 & 0.66 & 0.65 \\
\addlinespace[2pt]
\midrule
\multicolumn{14}{@{}l}{\emph{Closed-source VLMs}} \\
\multirow{3}{*}{Gemini 3.8 Flash} & Image & -- & -- & -- & -- & -- & 0.81 & \textbf{0.82} & 0.80 & 0.66 & 0.76 & 0.77 & 0.81 \\
 & Text & -- & -- & -- & -- & -- & 0.83 & \textbf{0.85} & 0.81 & 0.70 & 0.71 & 0.80 & 0.85 \\
 & Cross-modal & -- & -- & -- & -- & -- & \textbf{0.75} & 0.72 & 0.70 & 0.57 & 0.72 & 0.69 & 0.74 \\
\addlinespace[2pt]
\multirow{3}{*}{Gemini 3.1 Pro} & Image & -- & -- & -- & -- & -- & 0.67 & \textbf{0.73} & 0.68 & 0.62 & 0.78 & 0.67 & 0.73 \\
 & Text & -- & -- & -- & -- & -- & 0.79 & \textbf{0.87} & 0.80 & 0.75 & 0.70 & 0.80 & 0.87 \\
 & Cross-modal & -- & -- & -- & -- & -- & 0.66 & \textbf{0.71} & 0.65 & 0.58 & 0.66 & 0.65 & 0.71 \\
\addlinespace[2pt]
\multirow{3}{*}{Gemini 3.1 Flash-Lite} & Image & -- & -- & -- & -- & -- & 0.72 & \textbf{0.73} & \textbf{0.73} & 0.65 & 0.84 & 0.71 & 0.72 \\
 & Text & -- & -- & -- & -- & -- & 0.76 & \textbf{0.80} & 0.77 & 0.69 & 0.80 & 0.75 & 0.79 \\
 & Cross-modal & -- & -- & -- & -- & -- & 0.71 & \textbf{0.72} & 0.69 & 0.60 & 0.72 & 0.68 & 0.71 \\
\addlinespace[2pt]
\bottomrule
\end{tabular}}
\end{table*}

\begin{table*}[t]
\centering
\small
\setlength{\tabcolsep}{5pt}
\caption{\textbf{95\% intervals for the detection AUROCs.} Same task as Table~\ref{tab:det_llmuq_main}; brackets are percentile intervals over 2,000 bootstrap replicates that resample images. The semantic-entropy and answer-length intervals are those of the analysis tables (one replicate set per model); the interval of the mean of the four sampling-based scores (SE, LS, DM, LUQ) uses one replicate set shared by all nine models. $^{\ddagger}$Answers and samples from the original runs; $^{\S}$most greedy answers reach the 20-token cap (Appendix~\ref{app:generation}).}
\label{tab:det_llmuq_ci}
\resizebox{\ifdim\width>\textwidth\textwidth\else\width\fi}{!}{%
\begin{tabular}{@{}llccc@{}}
\toprule
\textbf{Model} & \textbf{Source} & \textbf{Semantic entropy} & \textbf{Sampling-score mean} & \textbf{Answer length (control)} \\
\midrule
\multicolumn{5}{@{}l}{\emph{Open-weight VLMs}} \\
\multirow{3}{*}{Qwen2.5-VL-7B$^{\ddagger}$} & Image & 0.85 [0.81, 0.90] & 0.76 [0.72, 0.80] & -- \\
 & Text & 0.89 [0.85, 0.93] & 0.77 [0.73, 0.81] & -- \\
 & Cross-modal & 0.80 [0.75, 0.85] & 0.67 [0.63, 0.72] & -- \\
\addlinespace[2pt]
\multirow{3}{*}{Gemma 4 12B$^{\S}$} & Image & 0.77 [0.71, 0.82] & 0.70 [0.64, 0.74] & 0.87 [0.82, 0.91] \\
 & Text & 0.84 [0.80, 0.88] & 0.78 [0.74, 0.82] & 0.86 [0.81, 0.90] \\
 & Cross-modal & 0.81 [0.75, 0.85] & 0.73 [0.68, 0.78] & 0.75 [0.69, 0.80] \\
\addlinespace[2pt]
\multirow{3}{*}{InternVL3.5-8B$^{\S}$} & Image & 0.82 [0.77, 0.87] & 0.82 [0.77, 0.86] & 0.50 [0.44, 0.57] \\
 & Text & 0.85 [0.80, 0.89] & 0.79 [0.74, 0.83] & 0.65 [0.60, 0.72] \\
 & Cross-modal & 0.77 [0.72, 0.82] & 0.70 [0.64, 0.76] & 0.70 [0.64, 0.76] \\
\addlinespace[2pt]
\multirow{3}{*}{Mistral-Small-3.2-24B$^{\S}$} & Image & 0.87 [0.83, 0.91] & 0.83 [0.79, 0.87] & 0.65 [0.59, 0.71] \\
 & Text & 0.87 [0.83, 0.91] & 0.83 [0.79, 0.86] & 0.67 [0.61, 0.73] \\
 & Cross-modal & 0.81 [0.76, 0.86] & 0.75 [0.69, 0.79] & 0.69 [0.62, 0.75] \\
\addlinespace[2pt]
\multirow{3}{*}{Qwen3.5-9B$^{\S}$} & Image & 0.82 [0.78, 0.87] & 0.77 [0.73, 0.82] & 0.71 [0.66, 0.77] \\
 & Text & 0.86 [0.82, 0.90] & 0.81 [0.77, 0.85] & 0.76 [0.71, 0.81] \\
 & Cross-modal & 0.73 [0.66, 0.78] & 0.69 [0.64, 0.75] & 0.74 [0.69, 0.80] \\
\addlinespace[2pt]
\multirow{3}{*}{Qwen3.8-27B$^{\S}$} & Image & 0.76 [0.70, 0.81] & 0.73 [0.68, 0.77] & 0.74 [0.68, 0.79] \\
 & Text & 0.83 [0.78, 0.87] & 0.80 [0.76, 0.84] & 0.76 [0.71, 0.81] \\
 & Cross-modal & 0.67 [0.60, 0.73] & 0.66 [0.60, 0.71] & 0.65 [0.58, 0.71] \\
\addlinespace[2pt]
\midrule
\multicolumn{5}{@{}l}{\emph{Closed-source VLMs}} \\
\multirow{3}{*}{Gemini 3.8 Flash} & Image & 0.81 [0.77, 0.86] & 0.77 [0.73, 0.81] & 0.76 [0.70, 0.82] \\
 & Text & 0.83 [0.78, 0.87] & 0.80 [0.76, 0.84] & 0.71 [0.65, 0.77] \\
 & Cross-modal & 0.75 [0.69, 0.81] & 0.69 [0.63, 0.74] & 0.72 [0.66, 0.78] \\
\addlinespace[2pt]
\multirow{3}{*}{Gemini 3.1 Pro} & Image & 0.67 [0.60, 0.73] & 0.67 [0.62, 0.73] & 0.78 [0.73, 0.83] \\
 & Text & 0.79 [0.74, 0.84] & 0.80 [0.76, 0.84] & 0.70 [0.64, 0.76] \\
 & Cross-modal & 0.66 [0.59, 0.72] & 0.65 [0.59, 0.71] & 0.66 [0.60, 0.72] \\
\addlinespace[2pt]
\multirow{3}{*}{Gemini 3.1 Flash-Lite} & Image & 0.72 [0.66, 0.78] & 0.71 [0.66, 0.76] & 0.84 [0.80, 0.89] \\
 & Text & 0.76 [0.71, 0.81] & 0.75 [0.71, 0.80] & 0.80 [0.75, 0.84] \\
 & Cross-modal & 0.71 [0.64, 0.77] & 0.68 [0.62, 0.73] & 0.72 [0.65, 0.77] \\
\addlinespace[2pt]
\bottomrule
\end{tabular}}
\end{table*}

\begin{table*}[t]
\centering
\small
\setlength{\tabcolsep}{5pt}
\caption{\textbf{How much of the detection AUROC varies with the model and with the source?} Two-way decomposition (no replication) of a model $\times$ source table of AUROCs for the task of Table~\ref{tab:det_llmuq_main}: share (\%) of the total sum of squares due to the model, the source and the residual (interaction plus noise), with 95\% intervals from 2,000 image-bootstrap replicates shared across models, and the F-test $p$ of each effect (models treated as fixed). \emph{Summary}: the AUROC entered per model--source setting, either the mean of the four sampling-based scores (SE, LS, DM, LUQ; exposed by every model) or semantic entropy alone. The six-model row omits Qwen3.5-9B, Qwen3.8-27B and Gemini 3.1 Flash-Lite; the last row replaces Qwen2.5-VL-7B's AUROCs with those of a rerun through the other models' sampler (point estimates only). Dropping any one of the nine models keeps the source share between 53\% and 65\% (sampling-score mean).}
\label{tab:det_llmuq_variance}
\begin{tabular}{@{}llccccc@{}}
\toprule
\textbf{Models} & \textbf{Summary} & \textbf{Model (\%)} & \textbf{Source (\%)} & \textbf{Residual (\%)} & $p_{\text{model}}$ & $p_{\text{source}}$ \\
\midrule
6 (no Qwen3.5/3.8, Flash-Lite) & sampling-score mean & 27 [13, 49] & 50 [27, 66] & 23 & 0.11 & 0.003 \\
9 & sampling-score mean & 25 [14, 47] & 57 [32, 68] & 18 & 0.04 & $<$0.001 \\
9 & semantic entropy & 55 [36, 71] & 33 [15, 49] & 12 & $<$0.001 & $<$0.001 \\
9, Qwen2.5-VL-7B resampled & sampling-score mean & 29 & 54 & 17 & 0.02 & $<$0.001 \\
\bottomrule
\end{tabular}
\end{table*}

\subsection{Uncertainty Response and Hallucination Consistency}
\label{app:res_urr_hcc}

\subsubsection{Uncertainty Rise under Perturbation}
\label{app:res_urr}
Table~\ref{tab:app_urr_family} reports the strict URR of three sampling-based LLM UQ scores (Lexical Similarity, DegMat and LUQ) for every perturbation family on the seven models that answered the perturbed inputs. For the four core models, it also counts the datasets on which a family lies above the sampling-noise level of Appendix~\ref{app:stats}. Figure~\ref{fig:urr_summary} shows, per model and modality, the share of these settings above that level.

\begin{table*}[t]
\centering
\scriptsize
\setlength{\tabcolsep}{3pt}
\caption{\textbf{Uncertainty-rise rate per perturbation family (strict URR$\uparrow$).} Share of paired items whose score is strictly higher on the perturbed than on the clean input (ties count as no rise), averaged over the three sample-based LLM UQ scores (lexical similarity, DegMat, LUQ; ten sampled answers per input) and the strengths of the family, then over the datasets on which the family is applied (VizWiz, VQAv2 and the four CLEVR tasks; family--dataset settings with fewer than 20 paired items dropped; \appvTwoUrrFamDsMin--\appvTwoUrrFamDsMax{} datasets per entry). \emph{$>$ref}: datasets on which the family's 95\% image-cluster bootstrap interval lies above the dataset-matched split-half reference of the model (strict URR of answers 6--10 over answers 1--5 of the same clean input, five answers per half; \appvTwoUrrFamAbove{} of \appvTwoUrrFamCells{} settings in total). The three added models have no split-half reference.}
\label{tab:app_urr_family}
\resizebox{\textwidth}{!}{%
\begin{tabular}{@{}lccccccccccc@{}}
\toprule
\textbf{Family} & \multicolumn{2}{c}{\textbf{Qwen2.5-VL-7B}} & \multicolumn{2}{c}{\textbf{Gemma 4 12B}} & \multicolumn{2}{c}{\textbf{Gemini 3.8 Flash}} & \multicolumn{2}{c}{\textbf{Gemini 3.1 Pro}} & \textbf{Qwen3.5-9B} & \textbf{Qwen3.8-27B} & \textbf{Gemini 3.1 Flash-Lite} \\
 & URR & $>$ref & URR & $>$ref & URR & $>$ref & URR & $>$ref & URR & URR & URR \\
\cmidrule(lr){2-3} \cmidrule(lr){4-5} \cmidrule(lr){6-7} \cmidrule(lr){8-9}
\midrule
\multicolumn{12}{@{}l}{\textit{Image perturbations}} \\
Blur & 0.67 & 5/6 & 0.56 & 3/6 & 0.70 & 6/6 & 0.68 & 6/6 & 0.80 & 0.68 & 0.64 \\
Brightness & 0.61 & 3/6 & 0.53 & 1/6 & 0.60 & 3/6 & 0.55 & 1/6 & 0.61 & 0.57 & 0.54 \\
Cutout & 0.76 & 6/6 & 0.60 & 4/6 & 0.65 & 5/6 & 0.63 & 4/6 & 0.75 & 0.68 & 0.60 \\
Gaussian noise & 0.80 & 6/6 & 0.61 & 4/6 & 0.73 & 6/6 & 0.71 & 6/6 & 0.78 & 0.73 & 0.70 \\
Pixelate & 0.70 & 6/6 & 0.57 & 3/6 & 0.67 & 6/6 & 0.66 & 6/6 & 0.71 & 0.66 & 0.60 \\
Salt-and-pepper & 0.79 & 6/6 & 0.59 & 4/6 & 0.66 & 4/6 & 0.63 & 4/6 & 0.77 & 0.74 & 0.56 \\
Solarize & 0.63 & 3/6 & 0.57 & 1/6 & 0.62 & 3/6 & 0.60 & 2/6 & 0.54 & 0.54 & 0.57 \\
\multicolumn{12}{@{}l}{\textit{Text perturbations (question edits)}} \\
Drop words & 0.60 & 3/6 & 0.59 & 2/6 & 0.55 & 1/6 & 0.57 & 1/6 & 0.58 & 0.54 & 0.55 \\
Invalid rewrite & 0.67 & 5/6 & 0.64 & 4/6 & 0.62 & 3/6 & 0.67 & 4/6 & 0.60 & 0.50 & 0.63 \\
Subjective rewrite & 0.80 & 6/6 & 0.80 & 6/6 & 0.72 & 5/6 & 0.74 & 5/6 & 0.77 & 0.69 & 0.81 \\
Shuffle words & 0.58 & 4/6 & 0.55 & 2/6 & 0.50 & 0/6 & 0.55 & 1/6 & 0.56 & 0.52 & 0.53 \\
Typos & 0.61 & 4/6 & 0.58 & 3/6 & 0.48 & 0/6 & 0.54 & 2/6 & 0.66 & 0.57 & 0.53 \\
\multicolumn{12}{@{}l}{\textit{Cross-modal perturbations}} \\
Ambiguous rewrite & 0.55 & 1/2 & 0.54 & 0/2 & 0.54 & 0/2 & 0.52 & 0/2 & 0.46 & 0.48 & 0.53 \\
Evidence removed (question) & 0.64 & 2/2 & 0.60 & 0/2 & 0.61 & 1/2 & 0.60 & 0/2 & 0.58 & 0.57 & 0.63 \\
Evidence removed (image) & 0.62 & 4/6 & 0.57 & 3/6 & 0.64 & 4/6 & 0.58 & 1/6 & 0.64 & 0.59 & 0.57 \\
\bottomrule
\end{tabular}}
\end{table*}

\paragraph{Matched-budget reference.} The sampling-noise level compares two halves of five answers each, whereas the perturbation scores use ten answers. We therefore also compare two independent ten-answer draws of the same clean input, with the same prompt and generation setup but a new seed (Table~\ref{tab:urr_matched}). Against this reference, \rfUrrFwdImage{} of the 168 image settings, \rfUrrFwdText--\rfUrrRevText{} of the 120 text settings and \rfUrrRevCross--\rfUrrFwdCross{} of the 40 cross-modal settings of Figure~\ref{fig:urr_summary} lie above it, depending on which draw is compared with which; \rfUrrBothImage, \rfUrrBothText{} and \rfUrrBothCross{} lie above it in both directions. The order image, text, cross-modal holds in every case. The Qwen2.5-VL-7B settings come from its earlier sampler, so its reference matches the number of answers but not the sampler. %

\begin{table}[t]
\centering
\scriptsize
\setlength{\tabcolsep}{3pt}
\caption{\textbf{URR against a matched-budget reference.} Settings of the four core models whose URR interval lies above the
reference interval of the same model and dataset (the rule of Figure~\ref{fig:urr_summary}). The matched reference
compares two independent ten-answer draws of the same clean input (same prompt and runner, new seed).}
\label{tab:urr_matched}
\begin{tabular}{@{}lccc@{}}
\toprule
Reference & Image & Text & Cross-modal \\
\midrule
Split-half, five vs.\ five answers of one draw & 117/168 & 61/120 & 16/40 \\
Ten-answer resample above the original draw & 112/168 & 61/120 & 14/40 \\
Original draw above the ten-answer resample & 112/168 & 62/120 & 10/40 \\
Above both directions & 95/168 & 51/120 & 7/40 \\
\bottomrule
\end{tabular}
\end{table}

\subsubsection{Perturbation-Induced Transitions and HCC}
\label{app:res_hcc}
Table~\ref{tab:app_halluc_rates} reports, for each of the 13 representative perturbations and the seven models that answered the perturbed inputs, the share of scene-audited CLEVR pairs whose clean answer is correct and whose perturbed answer is incorrect. The drop-words, INV and SBJ edits can change the correct answer, so we list their transitions but count them as hallucinations neither here nor in HCC. Solarization at strength 1 inverts almost every pixel, so for colour questions the clean answer need not describe the solarized image; its attribute transitions should be read with this in mind.
Figure~\ref{fig:claim2_hallucination_hcc}(b) summarizes, per score, the sign of HCC($\Delta U$) over every representative perturbation and CLEVR task on the scene-audited items. It uses the population of Appendix~\ref{app:stats}: perturbed abstentions are excluded, and initially incorrect items are kept with $H=0$. Because HCC is a correlation, a weak or negative value does not tell how many transitions a score would flag (Appendix~\ref{app:stats}).

\paragraph{Details of Figure~\ref{fig:claim2_hallucination_hcc}.}
Panel (a) averages the \cTwoNImage{} image perturbations of Table~\ref{tab:app_halluc_rates} (the 7 families, with brightness in both directions) and its \cTwoNTextKept{} text edits that remove no word (Shuffle and Typos). We did not verify that these two edits preserve the meaning of the question, and a shuffle can reorder relation terms. Panel (b) uses the same perturbations, the population of Appendix~\ref{app:stats} and the answer-conditioned P(True) prompt. It omits \cTwoHCCcellsUndef{} undefined settings, and \cTwoHCCcellsLowN{} of the \cTwoHCCcells{} settings have fewer than 10 transitions. With HCC as defined in Sec.~\ref{sec:evaluation}, \cTwoHCCcellsPos{} of the \cTwoHCCcells{} settings have an interval above zero and \cTwoHCCcellsNeg{} below (Figure~\ref{fig:claim2_hallucination_hcc}(b)); Table~\ref{tab:hcc_cc} repeats the count on initially correct pairs. Per perturbation family, the share of LLM UQ settings with an interval above zero ranges from 24\% (solarization) to 46\% (cutout). %
When only answers that state a wrong yes/no, number or attribute count as transitions, \cTwoHCCcommittedPos{} of \cTwoHCCcommittedCells{} settings have an interval above zero.
For the Gemini models, scoring the full answers instead of their first 20 tokens changes \cTwoProCapHdiffers{} (Pro), \cTwoLiteCapHdiffers{} (Flash-Lite) and \cTwoFlashCapHdiffers{} (Flash) of the \cTwoFlashPairs{} pair labels per model, in either direction (transitions \cTwoProCapH{}$\to$\cTwoProFullH{}, \cTwoLiteCapH{}$\to$\cTwoLiteFullH{} and \cTwoFlashCapH{}$\to$\cTwoFlashFullH{}). The output budget cut off \cTwoFlashMaxTokAnswers{} of the \cTwoFlashPairs{} perturbed Gemini 3.8 Flash answers, including \cTwoFlashMaxTokH{} of its \cTwoFlashH{} transitions, \cTwoFlashMaxTokHcommitted{} of which state a wrong answer (Appendix~\ref{app:generation}).

\begin{table*}[t]
\centering
\scriptsize
\setlength{\tabcolsep}{2.6pt}
\caption{\textbf{Correct-to-wrong transitions under each representative perturbation (\%, scene-audited CLEVR items).} Clean acc.: share of clean answers judged correct. Other columns: pairs whose clean answer is correct and whose perturbed answer is incorrect, as a percentage of all scene-audited pairs of the task (abstentions stay in the denominator and are not counted as wrong). Task-aware scorer on the 20-token answers, not validated against human ratings. One strength per family: blur 10, darkening 0.2, brightening 4, cutout 0.2, Gaussian noise 50, pixelation 5, salt-and-pepper 0.2, solarization 1, typos (strong), shuffle and drop words (minor), INV and SBJ rewrites. $^{\S}$Drop words can delete a referent word and the INV and SBJ rewrites change the question, so the clean gold answer need not apply to the perturbed question; these columns are not hallucinations in the sense of the main text. $^{\dagger}$Gemini 3.8 Flash: the API output budget ends many perturbed answers before the answer (Appendix~\ref{app:generation}).}
\label{tab:app_halluc_rates}
\resizebox{\textwidth}{!}{%
\begin{tabular}{@{}lcccccccccccccc@{}}
\toprule
 & & \multicolumn{8}{c}{\textit{Image}} & \multicolumn{5}{c}{\textit{Text (question edits)}} \\
\cmidrule(lr){3-10} \cmidrule(lr){11-15}
\textbf{Model} & \textbf{Clean acc.} & Blur & Dark & Bright & Cutout & Noise & Pixel. & S\&P & Solar. & Shuffle & Typos & Drop$^{\S}$ & INV$^{\S}$ & SBJ$^{\S}$ \\
\midrule
\multicolumn{15}{@{}l}{\textit{Attribute} ($n=127$ scene-audited items)} \\
Qwen2.5-VL-7B & 78.0 & 26.0 & 10.2 & 27.6 & 25.2 & 33.1 & 11.0 & 40.9 & 55.9 & 3.1 & 7.1 & 10.2 & 44.9 & 19.7 \\
Gemma 4 12B & 51.2 & 18.9 & 9.4 & 23.6 & 11.0 & 11.8 & 11.0 & 15.0 & 40.9 & 2.4 & 3.9 & 7.1 & 28.3 & 11.0 \\
Gemini 3.8 Flash$^{\dagger}$ & 92.1 & 66.1 & 8.7 & 30.7 & 19.7 & 43.3 & 26.8 & 34.6 & 71.7 & 3.1 & 3.1 & 27.6 & 44.1 & 32.3 \\
Gemini 3.1 Pro & 78.0 & 32.3 & 7.9 & 23.6 & 10.2 & 21.3 & 14.2 & 13.4 & 56.7 & 3.1 & 5.5 & 24.4 & 48.8 & 22.8 \\
Qwen3.5-9B & 72.4 & 63.8 & 7.9 & 38.6 & 37.8 & 24.4 & 20.5 & 49.6 & 62.2 & 4.7 & 30.7 & 20.5 & 59.8 & 27.6 \\
Qwen3.8-27B & 69.3 & 51.2 & 18.9 & 34.6 & 29.1 & 42.5 & 30.7 & 52.8 & 64.6 & 3.1 & 7.9 & 16.5 & 57.5 & 34.6 \\
Gemini 3.1 Flash-Lite & 59.1 & 29.9 & 3.1 & 18.1 & 13.4 & 15.7 & 14.2 & 15.7 & 38.6 & 4.7 & 6.3 & 12.6 & 30.7 & 19.7 \\
\midrule
\multicolumn{15}{@{}l}{\textit{Counting} ($n=150$ scene-audited items)} \\
Qwen2.5-VL-7B & 98.7 & 49.3 & 4.7 & 24.0 & 34.7 & 34.0 & 13.3 & 47.3 & 27.3 & 4.0 & 18.7 & 12.0 & 38.7 & 28.0 \\
Gemma 4 12B & 64.0 & 28.0 & 22.0 & 27.3 & 32.7 & 37.3 & 26.7 & 29.3 & 31.3 & 7.3 & 16.7 & 17.3 & 16.0 & 14.7 \\
Gemini 3.8 Flash$^{\dagger}$ & 89.3 & 46.7 & 13.3 & 26.7 & 21.3 & 36.0 & 24.7 & 28.0 & 29.3 & 1.3 & 4.7 & 16.7 & 35.3 & 5.3 \\
Gemini 3.1 Pro & 88.0 & 54.0 & 9.3 & 22.7 & 28.0 & 37.3 & 30.0 & 28.0 & 28.0 & 2.7 & 5.3 & 18.0 & 38.0 & 7.3 \\
Qwen3.5-9B & 64.0 & 47.3 & 24.0 & 31.3 & 45.3 & 38.0 & 29.3 & 45.3 & 24.0 & 14.0 & 30.0 & 10.0 & 24.0 & 10.0 \\
Qwen3.8-27B & 30.0 & 27.3 & 11.3 & 15.3 & 14.7 & 20.7 & 10.0 & 21.3 & 16.7 & 3.3 & 12.7 & 1.3 & 10.7 & 5.3 \\
Gemini 3.1 Flash-Lite & 87.3 & 39.3 & 6.7 & 22.0 & 24.7 & 26.0 & 17.3 & 21.3 & 20.7 & 6.0 & 12.0 & 14.7 & 24.7 & 20.0 \\
\midrule
\multicolumn{15}{@{}l}{\textit{Existence} ($n=129$ scene-audited items)} \\
Qwen2.5-VL-7B & 100.0 & 34.9 & 17.1 & 26.4 & 21.7 & 25.6 & 7.0 & 39.5 & 52.7 & 0.8 & 31.8 & 8.5 & 36.4 & 79.8 \\
Gemma 4 12B & 79.8 & 9.3 & 4.7 & 13.2 & 7.8 & 6.2 & 9.3 & 10.1 & 36.4 & 0.8 & 10.1 & 10.9 & 14.7 & 44.2 \\
Gemini 3.8 Flash$^{\dagger}$ & 99.2 & 39.5 & 0.8 & 14.0 & 5.4 & 10.9 & 10.1 & 10.9 & 45.0 & 0.8 & 4.7 & 9.3 & 31.8 & 63.6 \\
Gemini 3.1 Pro & 98.4 & 38.0 & 2.3 & 17.8 & 4.7 & 15.5 & 8.5 & 9.3 & 45.0 & 6.2 & 3.1 & 11.6 & 32.6 & 39.5 \\
Qwen3.5-9B & 94.6 & 34.1 & 5.4 & 30.2 & 12.4 & 13.2 & 14.0 & 19.4 & 54.3 & 0.8 & 38.8 & 7.0 & 37.2 & 34.9 \\
Qwen3.8-27B & 97.7 & 37.2 & 6.2 & 17.8 & 7.0 & 14.0 & 10.1 & 30.2 & 51.2 & 2.3 & 22.5 & 7.8 & 33.3 & 26.4 \\
Gemini 3.1 Flash-Lite & 96.9 & 33.3 & 3.9 & 17.8 & 6.2 & 14.7 & 11.6 & 8.5 & 48.1 & 6.2 & 13.2 & 9.3 & 31.8 & 82.9 \\
\midrule
\multicolumn{15}{@{}l}{\textit{Relation} ($n=56$ scene-audited items)} \\
Qwen2.5-VL-7B & 44.6 & 16.1 & 3.6 & 14.3 & 17.9 & 17.9 & 10.7 & 19.6 & 14.3 & 1.8 & 17.9 & 12.5 & 30.4 & 35.7 \\
Gemma 4 12B & 21.4 & 5.4 & 0.0 & 8.9 & 8.9 & 7.1 & 3.6 & 7.1 & 14.3 & 1.8 & 5.4 & 7.1 & 8.9 & 14.3 \\
Gemini 3.8 Flash$^{\dagger}$ & 26.8 & 16.1 & 3.6 & 10.7 & 3.6 & 8.9 & 10.7 & 12.5 & 12.5 & 1.8 & 3.6 & 10.7 & 17.9 & 17.9 \\
Gemini 3.1 Pro & 33.9 & 16.1 & 7.1 & 10.7 & 12.5 & 7.1 & 8.9 & 7.1 & 19.6 & 5.4 & 5.4 & 14.3 & 30.4 & 28.6 \\
Qwen3.5-9B & 28.6 & 26.8 & 3.6 & 16.1 & 23.2 & 16.1 & 7.1 & 25.0 & 3.6 & 3.6 & 19.6 & 8.9 & 16.1 & 17.9 \\
Qwen3.8-27B & 33.9 & 26.8 & 0.0 & 12.5 & 19.6 & 23.2 & 5.4 & 28.6 & 8.9 & 3.6 & 12.5 & 10.7 & 16.1 & 12.5 \\
Gemini 3.1 Flash-Lite & 55.4 & 21.4 & 12.5 & 19.6 & 17.9 & 12.5 & 7.1 & 16.1 & 16.1 & 14.3 & 19.6 & 28.6 & 30.4 & 51.8 \\
\bottomrule
\end{tabular}}
\end{table*}

\paragraph{Initially correct pairs.} The negative class of HCC also contains pairs whose clean answer is already wrong, a median of \rfHccInitWrongMed{} per setting. Table~\ref{tab:hcc_cc} therefore repeats the analysis of Figure~\ref{fig:claim2_hallucination_hcc}(b) on pairs whose clean answer is correct. It also reports the AUROC with which the score change, or the score of the perturbed input alone, ranks correct-to-wrong transitions. HCC has an interval above zero in \rfHccCcAbove{} and below zero in \rfHccCcBelow{} of the 1,902 settings (median \rfHccCcMed). The seven scores other than PMI and P(True) have \rfHccSevenAbove{} of their \rfHccSevenCells{} settings above zero and \rfHccSevenBelow{} below. P(True) has \rfHccPtrueAbove{} above and \rfHccPtrueBelow{} below, and the library implementation of PMI accounts for most of the remaining negative settings. On the three models for which both PMI variants are available, the text-only PMI of Appendix~\ref{app:scores} has a median HCC of \rfHccFixedMed{} (AUROC \rfHccFixedAuroc), against \rfHccLibSameMed{} for the library PMI on the same settings. %

\paragraph{Answers cut by the output budget.} For the Gemini models, we record the API finish reason and the full answer. Of Gemini 3.8 Flash's \rfTrFlashTrans{} correct-to-wrong transitions, \rfTrFlashCut{} (\rfTrFlashCutPct\%) involve a perturbed answer cut by the output budget. Excluding these pairs lowers its median AUROC of the score change from \rfTrFlash{} to \rfTrFlashNoapi{} (\rfTrFlashNoapiAbove{} settings above 0.5 and \rfTrFlashNoapiBelow{} below; \rfTrFlashNoapiDefined{} of \rfTrFlashCells{} settings defined). For Gemini 3.1 Pro (\rfTrPro{} to \rfTrProNoapi) and Gemini 3.1 Flash-Lite (\rfTrLite{} to \rfTrLiteNoapi), the change is small. Grading the full rather than the 20-token answers gives median AUROCs of \rfTrFlashFull{} (Flash), \rfTrProFull{} (Pro) and \rfTrLiteFull{} (Flash-Lite). %

\begin{table*}[t]
\centering
\scriptsize
\setlength{\tabcolsep}{2.5pt}
\caption{\textbf{HCC on initially correct pairs.} The 1,902 settings of Figure~\ref{fig:claim2_hallucination_hcc}(b),
restricted to pairs whose clean answer is correct ($H=1$: the perturbed answer is wrong). HCC$_{\text{all}}$: the paper's
HCC, which keeps initially incorrect pairs with $H=0$. CI $>0$ / $<0$: settings whose 95\% bootstrap interval lies above or
below zero. AUROC: median AUROC with which the score change $\Delta U$, or the perturbed-input score $U(x')$ alone,
ranks correct-to-wrong transitions above pairs that stay correct (0.5 = no signal). Medians over settings.
$^{\dagger}$The PMI of all our tables, as implemented in the LM-Polygraph fork: its prior term is read from the
image-and-question forward pass rather than from the text-only model of Appendix~\ref{app:scores}. The last row gives the
text-only PMI of Appendix~\ref{app:scores}, stored for Gemma 4 12B, Qwen3.5-9B and Qwen3.8-27B; it is not part of the
1,902 settings.}
\label{tab:hcc_cc}
\begin{tabular}{@{}lrcccccc@{}}
\toprule
Score & Settings & HCC$_{\text{all}}$ & HCC, correct pairs [IQR] & CI $>0$ / $<0$ & AUROC $\Delta U$ & AUROC $U(x')$ \\
\midrule
Semantic Entropy & 278 & 0.16 & 0.17 [0.03, 0.31] & 96 / 5 & 0.61 & 0.72 \\
Lexical Similarity & 278 & 0.14 & 0.14 [0.02, 0.30] & 95 / 4 & 0.60 & 0.70 \\
DegMat & 278 & 0.16 & 0.17 [0.04, 0.29] & 105 / 7 & 0.62 & 0.73 \\
LUQ & 278 & 0.18 & 0.22 [0.07, 0.36] & 113 / 4 & 0.65 & 0.72 \\
MSP & 158 & 0.18 & 0.21 [0.07, 0.35] & 61 / 7 & 0.62 & 0.71 \\
Perplexity & 158 & 0.17 & 0.20 [0.06, 0.37] & 57 / 6 & 0.62 & 0.71 \\
MTE & 158 & 0.20 & 0.24 [0.09, 0.41] & 70 / 5 & 0.64 & 0.71 \\
P(True) & 158 & $-$0.01 & 0.01 [$-$0.16, 0.27] & 40 / 21 & 0.50 & 0.49 \\
PMI (library)$^{\dagger}$ & 158 & $-$0.11 & $-$0.12 [$-$0.28, 0.03] & 6 / 46 & 0.42 & 0.47 \\
\midrule
All & 1,902 & 0.14 & 0.15 [0.00, 0.32] & 643 / 105 & 0.60 & 0.69 \\
PMI, text-only prior & 118 & -- & 0.30 [0.08, 0.44] & 62 / 9 & 0.68 & 0.71 \\
\bottomrule
\end{tabular}
\end{table*}

\subsection{Answer-Length Confounding}
\label{app:ext_length}

\begin{table*}[t]
\centering
\scriptsize
\setlength{\tabcolsep}{3pt}
\renewcommand{\arraystretch}{1.05}
\caption{\textbf{Do the scores detect a source beyond answer length? (all nine VLMs)} AUROC$\uparrow$ for separating each
VizWiz source from the clean items. \emph{Length only} ranks items by the number of words in the scored greedy answer,
which is capped at 20 tokens. \emph{Length-controlled} is the best of the four sampling-based scores after length control:
the lower of its AUROC over clean--source pairs of identical answer length and its AUROC after removing the length trend
fitted on clean answers. Bold: above 0.60 after Holm correction, over the 36 pre-registered tests for the first three
models and over each model's 12 tests for the others (a secondary analysis with the same estimator). The best score is
chosen on the evaluation items. Cap: share of the Image-source greedy answers that reach the 20-token cap. Refusals: share
of greedy answers that are refusals. $^{\ast}$Rerun of Qwen2.5-VL-7B with the sampling pipeline of the other models.
Brackets give 95\% bootstrap intervals over images.}
\label{tab:ext_length}
\resizebox{\ifdim\width>\textwidth\textwidth\else\width\fi}{!}{%
\begin{tabular}{@{}lcccccccc@{}}
\toprule
 & \multicolumn{2}{c}{\textbf{Image}} & \multicolumn{2}{c}{\textbf{Text}} & \multicolumn{2}{c}{\textbf{Cross-modal}} & & \\
\cmidrule(lr){2-3} \cmidrule(lr){4-5} \cmidrule(lr){6-7}
\textbf{Model} & Length only & Length-controlled & Length only & Length-controlled & Length only & Length-controlled & Cap (\%) & Refusals, Image / clean (\%) \\
\midrule
Gemma 4 12B & 0.87 & 0.67 [0.58, 0.73] & 0.86 & \textbf{0.77 [0.70, 0.83]} & 0.75 & \textbf{0.77 [0.69, 0.82]} & 95 & 71 / 3 \\
Gemini 3.8 Flash & 0.76 & \textbf{0.73 [0.64, 0.79]} & 0.71 & \textbf{0.81 [0.75, 0.85]} & 0.72 & 0.65 [0.57, 0.71] & 81 & 51 / 1 \\
Gemini 3.1 Pro & 0.78 & 0.51 [0.44, 0.58] & 0.70 & \textbf{0.78 [0.72, 0.83]} & 0.66 & 0.62 [0.54, 0.69] & 95 & 46 / 2 \\
\midrule
InternVL3.5-8B & 0.50 & \textbf{0.85 [0.80, 0.89]} & 0.65 & \textbf{0.84 [0.78, 0.88]} & 0.70 & \textbf{0.74 [0.67, 0.80]} & 76 & 8 / 1 \\
Mistral-Small-3.2-24B & 0.65 & \textbf{0.82 [0.75, 0.87]} & 0.67 & \textbf{0.82 [0.75, 0.88]} & 0.69 & \textbf{0.77 [0.70, 0.83]} & 95 & 19 / 2 \\
Qwen3.5-9B & 0.71 & \textbf{0.81 [0.75, 0.86]} & 0.76 & \textbf{0.85 [0.79, 0.88]} & 0.74 & \textbf{0.70 [0.62, 0.76]} & 93 & 19 / 0 \\
Qwen3.8-27B & 0.74 & \textbf{0.72 [0.65, 0.78]} & 0.76 & \textbf{0.80 [0.74, 0.85]} & 0.65 & 0.65 [0.56, 0.71] & 95 & 24 / 0 \\
Gemini 3.1 Flash-Lite & 0.84 & 0.62 [0.54, 0.68] & 0.80 & \textbf{0.72 [0.64, 0.78]} & 0.72 & 0.68 [0.59, 0.74] & 97 & 63 / 3 \\
\midrule
Qwen2.5-VL-7B$^{\ast}$ & 0.75 & \textbf{0.79 [0.72, 0.85]} & 0.73 & \textbf{0.82 [0.76, 0.87]} & 0.73 & 0.67 [0.59, 0.74] & 95 & 27 / 2 \\
\bottomrule
\end{tabular}}
\end{table*}

Answer length alone, which is not an uncertainty score, separates Image items from clean items with AUROC \detLenGemmaImage, \detLenProImage{} and \detLenFlashLiteImage{} on Gemma 4 12B, Gemini 3.1 Pro and Gemini 3.1 Flash-Lite, above semantic entropy on the same models (\detSEGemmaImage, \detSEProImage{} and \detSEFlashLiteImage; Table~\ref{tab:det_llmuq_full_auroc}). The detection AUROCs could therefore partly measure how much a model says, a shortcut of the kind that also inflates measured abilities in other settings~\citep{zhong2026shortcuts}.
For the three models of the pre-registered length analysis (Gemma 4 12B, Gemini 3.8 Flash and Gemini 3.1 Pro), we compute each sampling-based score's AUROC in two ways: over clean and source answers of identical length, and after removing the length trend fitted on clean answers. We keep the lower of the two values and test it against the pre-registered bar of 0.60 (Table~\ref{tab:ext_length}). The two estimators and the rule of keeping the lower value were fixed before the first result but are not part of the pre-registration. %
As a secondary analysis, we apply the same estimator, which reproduces the values of the three pre-registered models exactly, to the other five VLMs with stored answers (InternVL3.5-8B, Mistral-Small-3.2-24B, Qwen3.5-9B, Qwen3.8-27B and Gemini 3.1 Flash-Lite), with Holm correction over each model's 12 tests. %

Text-source items stay separable on all three models: the best length-controlled AUROC is 0.77 on Gemma 4 12B (semantic entropy), 0.81 on Gemini 3.8 Flash and 0.78 on Gemini 3.1 Pro (both lexical similarity). %
Image-source items stay above the bar after Holm correction only on Gemini 3.8 Flash (0.73). %
On Gemma 4 12B and Gemini 3.1 Pro, the Image AUROC falls to 0.67 and 0.51 after length control, and neither is shown above 0.60. %
On these two models, answer length alone reaches 0.87 and 0.78, and adding the score to a length-only classifier changes AUROC by +0.01 and $-$0.01. %
Gemini 3.1 Pro's 0.51 depends on the form of the control: it is 0.66 with the full answer length and 0.73 on the answer lengths shared by at least five clean and five Image answers. %
Cross-modal items stay above the bar only on Gemma 4 12B (0.77). %
Because the best score is chosen on the evaluation items, we also choose it on one half of the items and evaluate it on the other; this lowers the length-controlled AUROC by at most 0.03 on the three pre-registered models. %
The length effect largely coincides with refusals: 46--71\% of Image-source answers are refusals, against 1--3\% of clean answers (counted on the 20-token scored answer for Gemma 4 12B and on the full answer for the Gemini models). A refusal can itself be read as a verbal expression of uncertainty, and rejecting when no valid answer is available remains a weak point of current models~\citep{zhong2026rejection}; the length-controlled scores show how much the sampled UQ scores add beyond it. %
A rerun of Qwen2.5-VL-7B with the other models' sampling pipeline, not included in the Holm correction, keeps length-controlled AUROCs of 0.79, 0.82 and 0.67 on the Image, Text and Cross-modal sources. %
The pre-registered prediction (at least one score above 0.60 after length control and Holm correction on each of the three models) is supported, and the pre-registered failure criterion (every score at or below 0.55 on all three models) is not met. %
The scores thus separate Text-source items from clean items beyond answer length on all three models, whereas for degraded images on Gemma 4 12B and Gemini 3.1 Pro they are not shown to add to answer length.

On the five added VLMs, Text detection stays above 0.60 on all five. Image detection does so on InternVL3.5-8B, Mistral-Small-3.2-24B, Qwen3.5-9B and Qwen3.8-27B (0.72--0.85), but not on Gemini 3.1 Flash-Lite (0.62; \rfLenLiteTok{} with the full answer length), which refuses 63\% of the Image items. Cross-modal detection stays above 0.60 on InternVL3.5-8B, Mistral-Small-3.2-24B and Qwen3.5-9B, although the Qwen3.5-9B result is no longer above 0.60 when the Holm correction covers all 60 added tests. %
The four open models refuse \rfLenLowRefLo--\rfLenLowRefHi\% of the Image items (judged on the 20-token answer), and adding the score to a length-only classifier raises their Image AUROC by 0.05 to 0.43 (point estimates; the interval of Qwen3.8-27B includes zero). %
Because \rfLenCapLo--\rfLenCapHi\% of their greedy answers reach the 20-token cap, their answer lengths vary little, and the control acts mostly within the capped range.

\paragraph{Answer prompt.}
\label{app:ext_prompt}
The scores read the sampled answers, so an instruction that changes the answers, including their length, could change what the scores detect; the same holds for models that reason in longer visual chains of thought~\citep{li2026visreason,li2025tirbench}. For four models, we re-ask the 572 VizWiz questions with two alternative prompts, one that appends ``Answer with a short phrase.'' (prompt A) and one that places the instruction before the question (prompt B), and recompute every score (Table~\ref{tab:ext_prompt}). Under prompt B, only two small shifts (+0.06 on Image for Gemma 4 12B and +0.04 on Text for Qwen2.5-VL-7B) differ from the benchmark prompt after Holm correction. %
Prompt A shortens the answers and cuts refusals of low-quality images from 27--71\% to 11--43\%; the best Image score of Gemma 4 12B gains 0.11 AUROC, the best Cross-modal score loses 0.06--0.07 on all four models, and answer length no longer beats every score on Gemma 4 12B's Image items (0.87 vs.\ 0.77 under the benchmark prompt, 0.82 vs.\ 0.88 under prompt A). %
P(True) is the least stable score: the answer-conditioned P(True) of Gemma 4 12B, the variant in our tables, moves by up to 0.51 AUROC from its benchmark-prompt value (on Image: 0.29 under the benchmark prompt, 0.79 under prompt A and 0.25 under prompt B), so the tables of this appendix report the benchmark-prompt values. %
The benchmark-prompt answers of Gemma 4 12B and the Gemini models come from an earlier sampling run with other seeds, so each change also contains sampling noise. Where the instruction is placed matters little, but which score is best, and whether answer length beats the scores, depend on whether the prompt asks for a short answer.

\begin{table}[t]
\centering
\scriptsize
\setlength{\tabcolsep}{3pt}
\renewcommand{\arraystretch}{1.05}
\caption{\textbf{Does the answer prompt change what the scores detect?} Change in AUROC of each model's best benchmark-prompt score when the same 572 VizWiz questions are asked with prompt A (``Answer with a short phrase.'' appended) or prompt B (the instruction placed before the question), all scores recomputed. 0.00 means no change; bold values differ from the benchmark prompt after Holm correction within the model. Qwen2.5-VL-7B is compared with a rerun of the benchmark prompt through the same sampling pipeline.}
\label{tab:ext_prompt}
\resizebox{\ifdim\width>\columnwidth\columnwidth\else\width\fi}{!}{%
\begin{tabular}{@{}lcccccc@{}}
\toprule
 & \multicolumn{2}{c}{\textbf{Image}} & \multicolumn{2}{c}{\textbf{Text}} & \multicolumn{2}{c}{\textbf{Cross-modal}} \\
\cmidrule(lr){2-3} \cmidrule(lr){4-5} \cmidrule(lr){6-7}
\textbf{Model} & A & B & A & B & A & B \\
\midrule
Qwen2.5-VL-7B & +0.01 & +0.02 & $-$0.02 & \textbf{+0.04} & $-$0.07 & +0.04 \\
Gemma 4 12B & \textbf{+0.11} & \textbf{+0.06} & $-$0.04 & +0.02 & $-$0.06 & +0.03 \\
Gemini 3.8 Flash & $-$0.01 & +0.02 & $-$0.01 & +0.02 & $-$0.07 & +0.02 \\
Gemini 3.1 Pro & +0.07 & +0.03 & +0.01 & +0.01 & $-$0.06 & +0.01 \\
\bottomrule
\end{tabular}}
\end{table}
\section{VLM-Specific UQ Methods}
\label{app:vlmuq}
We evaluate five VLM-specific UQ methods on Qwen2.5-VL-7B and LLaVA-1.5-7B: VL-Uncertainty (VLU)~\citep{zhang2024vl}, VAUQ~\citep{park2026vauq}, response SVAR, the score module of Dropout Decoding~\citep{fang2025dropout} and BICR~\citep{khanmohammadi2026grounded}.
These five are a selection rather than an exhaustive set; Table~\ref{tab:slim-native-methods} lists the score, endpoints, coverage, supervision and matched control of each. Inconsistency between models is also used as an uncertainty signal outside VQA, for example to detect out-of-distribution objects with a text-to-image model~\citep{nguyen2026ood}.
Each method uses the score proposed by its authors, and we choose no score, layer or mask setting on the evaluation outcomes. All method settings and analysis rules were set before these runs, although the benchmark data had been used in earlier development.
The LLM UQ rows (MSP, Perplexity and Mean Token Entropy) and the answer-length control in this appendix score the same 128-token greedy answer as the VLM-specific methods. This run is separate from the nine-VLM LLM UQ results (Appendix~\ref{app:res_detection}), so the two should not be compared value by value: for Qwen2.5-VL-7B, MSP separates Image from Clean with AUROC 0.777 here and 0.84 in Table~\ref{tab:det_llmuq_full_auroc}. %

\begin{table*}[t]
\centering\scriptsize
\setlength{\tabcolsep}{3pt}
\begin{tabular}{p{.09\textwidth}p{.27\textwidth}p{.16\textwidth}p{.19\textwidth}p{.15\textwidth}}
\toprule
Method & Evaluated score & Endpoints; coverage & Supervision / auxiliary model & Matched control\\
\midrule
\raggedright VLU & \raggedright entropy of entailment clusters over 5 answers to blurred-image and rephrased-question pairs (32 tokens) & \raggedright all; every query scored & \raggedright none; Qwen2.5-3B-Instruct rephrases and judges & \raggedright same-input SE with the same sampler and judge\tabularnewline
\raggedright VAUQ & \raggedright $H-\alpha(H_{\mathrm{mask}}-H)$ on the greedy answer, published $K,\alpha$ per backbone & \raggedright all; every query scored & \raggedright none & \raggedright unmasked $H$; equal-size random mask\tabularnewline
\raggedright Response SVAR & \raggedright negated answer-to-image attention, summed over layers & \raggedright all; every query scored & \raggedright none & \raggedright LLM UQ on the same answer\tabularnewline
\raggedright Dropout score & \raggedright mean visual-token entropy (aleatoric) and KL to the mean (epistemic); no decoding intervention & \raggedright all; every query scored & \raggedright none & \raggedright both components reported\tabularnewline
\raggedright BICR & \raggedright MLP probe on the last prompt state; uncertainty $=1-$confidence & \raggedright all; every query scored; CLEVR scores from probes that exclude the test images & \raggedright correctness labels of 430 CLEVR originals; 360 fits per backbone & \raggedright same grid without ranking (and without Brier) term\tabularnewline
\bottomrule
\end{tabular}
\caption{VLM-specific UQ methods as evaluated on Qwen2.5-VL-7B and LLaVA-1.5-7B. ``all\textquotedblright\ means VizWiz source-versus-Clean detection and the 150 perturbation settings; every query of these endpoints has a finite score on each backbone. Response SVAR and the Dropout score module are the scores we evaluate, not full reproductions of the original pipelines. BICR is trained only on scene-graph-verified CLEVR originals: original (unperturbed) questions with a known correctness label whose answer the official scene graph verifies; perturbed versions and other descendants of these images are excluded. VizWiz scores come from the probe trained on all of them, CLEVR scores from five probes that each exclude the test images.}
\label{tab:slim-native-methods}
\end{table*}

\subsection{Implementation and Matched Baselines}
\label{app:native-protocol}

\paragraph{VAUQ.}
\label{app:native-vauq}
VAUQ~\citep{park2026vauq} uses visual attention to mask informative image tokens.
For a fixed greedy answer, it combines the mean predictive entropy $H$ with the entropy change after masking the salient visual tokens:
\begin{equation}
  s_{\mathrm{VAUQ}}=H-\alpha\left(H_{\mathrm{mask}}-H\right),
\end{equation}
with entropy in bits and higher scores indicating greater uncertainty.
We transfer the published ViLP settings without tuning them on our labels: $K=\alpha=0.8$ for Qwen and $0.6$ for LLaVA, where $K$ is the fraction of masked visual tokens. %
The matched control masks the same number of randomly chosen visual tokens, with one fixed mask ordering per image, so its intervals do not include variation over mask seeds.
At the published setting, the attention-selected mask has a higher AUROC than the equal-size random mask on all three VizWiz endpoints of both backbones (Qwen2.5-VL-7B: 0.800 vs.\ 0.746 for Image, 0.821 vs.\ 0.777 for Text and 0.757 vs.\ 0.736 for Cross-modal; LLaVA-1.5-7B: 0.773 vs.\ 0.747, 0.712 vs.\ 0.701 and 0.707 vs.\ 0.674; point estimates), and it does so at 29 or 30 of the 30 settings of a sensitivity grid over $K$ and $\alpha$ for every backbone and endpoint. %

\paragraph{VL-Uncertainty (VLU).}
VL-Uncertainty pairs progressively blurred images with independently rephrased questions: the five blur radii $[0.6,0.8,1.0,1.2,1.4]$ are paired in order with the rephrasing temperatures $[0.1,0.2,0.3,0.4,0.5]$. %
As in the original implementation, an auxiliary Qwen2.5-3B-Instruct model writes the rephrasings and judges whether two answers entail each other. %
The VLM gives one sampled answer per variant. An answer joins a semantic cluster when the auxiliary model judges it and the cluster's first member to entail each other, and the score is the Shannon entropy of the cluster frequencies, in bits.
The transformations are meant to preserve meaning, but we did not check that every blur or rephrasing keeps its visual evidence or proposition.
VLU's own schedule uses temperature 1, top-$k$ 50, top-$p$ 0.95, a 32-token limit and repetition penalty 1.05; as a sensitivity analysis we also run it with the common 128-token limit and repetition penalty 1.0. %
At each budget, a matched Semantic Entropy control samples the unchanged image and question with the same number of answers, decoding settings and judge.
Image preprocessing is shared with the rest of the benchmark, so matching VLU's sampling rule does not reproduce its published benchmark results.

\paragraph{Response SVAR and the Dropout score module.}
Response SVAR follows the scalar comparison in \citet{park2026vauq}: it averages the attention from answer tokens to visual tokens over heads and answer tokens, sums the selected layers, and negates the result so that larger values mean greater uncertainty.
This response-level statistic leaves out the object-token classifier of the original SVAR pipeline.
For the score module of Dropout Decoding~\citep{fang2025dropout}, let $p_i$ be the vocabulary distribution obtained by projecting the final decoder hidden state at visual position $i$ through the language-model head, and let $\bar p$ be their mean over visual positions.
We keep the two published components,
\begin{align}
 U_{\mathrm{alea}} &= \frac{1}{N}\sum_i H(p_i),\\
 U_{\mathrm{epi}} &= \frac{1}{N}\sum_i D_{\mathrm{KL}}(p_i\Vert\bar p),
\end{align}
in nats, without the uncertainty-guided masking and decoding ensemble that follow them in the original method.

\paragraph{BICR.}
BICR takes the final decoder state at the last prompt position for the real image and for a black image of the same size; during training, one shared MLP scores both vectors.
The loss combines class-weighted binary cross-entropy, a Brier term and a margin-ranking term that rewards higher confidence on the real image, applied only to correct training examples with a valid blank view.
At test time BICR uses the real-image vector alone, and its uncertainty is one minus the mean confidence of the selected seed models.
The matched controls are weighted BCE alone and BCE plus Brier without the ranking term.
All three losses search the same grid of four configurations and five seeds, and each loss selects its own configuration, so the controls share the search space and seeds with full BICR but may end up with different hyperparameters. %
We use six probe contexts per backbone, a corpus probe and five CLEVR probes, each with 60 candidate fits (four configurations, five seeds and three losses), i.e.\ 360 fits per backbone. The corpus probe is trained only on scene-graph-verified CLEVR originals (Table~\ref{tab:slim-native-methods}) and is applied unchanged to VizWiz, whose labels neither train nor select it. For the CLEVR perturbation analyses, the original and perturbed queries are scored by the five CLEVR probes, each of which excludes the test images. %

\paragraph{Scope.}
None of these components is designed to separate the benchmark's Image, Text or Cross-modal sources, so names such as aleatoric and epistemic are not source labels.
We do not evaluate two further methods. ContextualLens~\citep{phukan2025beyond}, a training-free answer-level embedding score, is outside our scope. For VL-Calibration~\citep{xiao2026vlcalibration}, we found no released fine-tuned checkpoint for either backbone, and we did not replace the learned method with an untrained confidence prompt.

\subsection{Source Detection}
\label{app:native-scalar-matrix}\label{app:native-detection}
Table~\ref{tab:slim-native-detection-full} reports every VLM-specific score, component and control on the three VizWiz detection endpoints, each of which separates one uncertain source from Clean.
Every score keeps its published orientation, so some AUROCs fall below chance, and differences between scores or backbones are point estimates.

\begin{table*}[t]
\centering\scriptsize
\setlength{\tabcolsep}{2.5pt}
\begin{tabular}{lccc}
\toprule
Score & Image vs.\ Clean & Text vs.\ Clean & Cross-modal vs.\ Clean\\
 & ($n=300$) & ($n=300$) & ($n=272$)\\
\midrule
\multicolumn{4}{l}{\textbf{Qwen2.5-VL-7B}}\\
\multicolumn{4}{l}{\textit{VLM-specific UQ}}\\
VLU, 32 tok. & 0.605 [0.550, 0.658] & 0.643 [0.591, 0.693] & 0.626 [0.567, 0.682]\\
\quad SE, 5 samples, 32 tok. & 0.648 [0.593, 0.700] & 0.667 [0.613, 0.718] & 0.629 [0.570, 0.685]\\
\quad VLU, 128 tok. & 0.607 [0.553, 0.662] & 0.651 [0.602, 0.704] & 0.666 [0.615, 0.718]\\
\quad SE, 5 samples, 128 tok. & 0.661 [0.607, 0.714] & 0.665 [0.611, 0.720] & 0.676 [0.620, 0.727]\\
VAUQ & 0.800 [0.749, 0.849] & 0.821 [0.774, 0.864] & 0.757 [0.697, 0.810]\\
\quad VAUQ unmasked $H$ & 0.780 [0.728, 0.831] & 0.805 [0.757, 0.850] & 0.741 [0.681, 0.798]\\
\quad VAUQ random mask & 0.746 [0.688, 0.802] & 0.777 [0.727, 0.826] & 0.736 [0.677, 0.792]\\
\quad VAUQ term $H_{\mathrm{mask}}{-}H$ & 0.291 [0.232, 0.350] & 0.294 [0.239, 0.353] & 0.334 [0.272, 0.402]\\
Response SVAR & 0.832 [0.784, 0.876] & 0.815 [0.765, 0.859] & 0.679 [0.613, 0.742]\\
Dropout score: aleatoric & 0.813 [0.763, 0.860] & 0.701 [0.637, 0.760] & 0.785 [0.729, 0.839]\\
Dropout score: epistemic & 0.227 [0.177, 0.281] & 0.319 [0.260, 0.380] & 0.348 [0.284, 0.413]\\
BICR & 0.903 [0.865, 0.940] & 0.894 [0.852, 0.930] & 0.535 [0.466, 0.606]\\
\quad BICR loss control: BCE+Brier & 0.855 [0.810, 0.899] & 0.872 [0.825, 0.911] & 0.511 [0.442, 0.582]\\
\quad BICR loss control: BCE & 0.770 [0.715, 0.826] & 0.826 [0.775, 0.871] & 0.465 [0.397, 0.537]\\
\multicolumn{4}{l}{\textit{LLM UQ on the same greedy answer}}\\
MSP & 0.777 [0.724, 0.829] & 0.815 [0.766, 0.861] & 0.761 [0.705, 0.814]\\
PPL (mean token NLL) & 0.770 [0.719, 0.822] & 0.806 [0.758, 0.852] & 0.745 [0.686, 0.801]\\
MTE & 0.789 [0.735, 0.840] & 0.810 [0.761, 0.855] & 0.745 [0.684, 0.801]\\
Answer length (control) & 0.751 [0.697, 0.805] & 0.790 [0.737, 0.837] & 0.744 [0.688, 0.797]\\
\midrule
\multicolumn{4}{l}{\textbf{LLaVA-1.5-7B}}\\
\multicolumn{4}{l}{\textit{VLM-specific UQ}}\\
VLU, 32 tok. & 0.567 [0.524, 0.606] & 0.550 [0.506, 0.593] & 0.575 [0.531, 0.616]\\
\quad SE, 5 samples, 32 tok. & 0.604 [0.564, 0.645] & 0.579 [0.539, 0.621] & 0.586 [0.542, 0.630]\\
\quad VLU, 128 tok. & 0.583 [0.542, 0.619] & 0.549 [0.504, 0.594] & 0.574 [0.530, 0.617]\\
\quad SE, 5 samples, 128 tok. & 0.571 [0.532, 0.610] & 0.551 [0.510, 0.594] & 0.562 [0.521, 0.603]\\
VAUQ & 0.773 [0.719, 0.826] & 0.712 [0.653, 0.769] & 0.707 [0.643, 0.765]\\
\quad VAUQ unmasked $H$ & 0.745 [0.688, 0.799] & 0.697 [0.638, 0.755] & 0.673 [0.607, 0.736]\\
\quad VAUQ random mask & 0.747 [0.689, 0.801] & 0.701 [0.644, 0.758] & 0.674 [0.608, 0.737]\\
\quad VAUQ term $H_{\mathrm{mask}}{-}H$ & 0.277 [0.222, 0.337] & 0.347 [0.284, 0.411] & 0.321 [0.260, 0.383]\\
Response SVAR & 0.612 [0.549, 0.674] & 0.722 [0.661, 0.779] & 0.695 [0.630, 0.759]\\
Dropout score: aleatoric & 0.620 [0.556, 0.684] & 0.469 [0.406, 0.534] & 0.349 [0.284, 0.418]\\
Dropout score: epistemic & 0.385 [0.324, 0.448] & 0.408 [0.344, 0.471] & 0.571 [0.506, 0.638]\\
BICR & 0.534 [0.469, 0.597] & 0.461 [0.394, 0.526] & 0.547 [0.482, 0.615]\\
\quad BICR loss control: BCE+Brier & 0.535 [0.470, 0.598] & 0.454 [0.387, 0.520] & 0.556 [0.489, 0.624]\\
\quad BICR loss control: BCE & 0.553 [0.489, 0.618] & 0.433 [0.368, 0.499] & 0.525 [0.459, 0.592]\\
\multicolumn{4}{l}{\textit{LLM UQ on the same greedy answer}}\\
MSP & 0.682 [0.621, 0.742] & 0.667 [0.604, 0.727] & 0.711 [0.649, 0.770]\\
PPL (mean token NLL) & 0.732 [0.671, 0.788] & 0.685 [0.627, 0.743] & 0.688 [0.624, 0.749]\\
MTE & 0.743 [0.684, 0.798] & 0.695 [0.636, 0.753] & 0.673 [0.607, 0.735]\\
Answer length (control) & 0.608 [0.545, 0.672] & 0.646 [0.584, 0.708] & 0.712 [0.649, 0.772]\\
\bottomrule
\end{tabular}
\caption{All VLM-specific scores, components and controls on the three VizWiz detection endpoints (AUROC with pointwise 95\% image-bootstrap intervals, 2,000 draws, conditional on the saved outputs). Within a backbone and endpoint all 18 scores are finite on the same items. Indented rows are matched controls or published components of the method above; the SE rows are LLM UQ controls sharing VLU\textquoteright s sampler, answer count and judge. VLU and SE use the native 32-token schedule and a 128-token sensitivity with repetition penalty 1.0. The VAUQ term $H_{\mathrm{mask}}{-}H$ and the two Dropout components are reported as published components, not as separately proposed scores. BICR rows use the probe trained on CLEVR originals; the loss controls share its configuration grid and seeds but select their own configuration. Intervals are marginal; overlap or non-overlap is not a test of a difference.}
\label{tab:slim-native-detection-full}
\end{table*}

\subsection{Perturbation Response}
\label{app:native-perturbation-matrices}
Table~\ref{tab:slim-native-risk} summarizes the full results of each score: the median URR over the image, text and cross-modal perturbation settings, the median HCC($\Delta U$) over the CLEVR settings, and the number of settings whose pointwise interval lies above or below zero.
URR is the fraction of original--perturbed pairs whose uncertainty strictly increases, with ties counted as no increase. HCC($\Delta U$) is the point-biserial correlation between the uncertainty change and a transition from a correct original answer to an incorrect perturbed answer, computed on CLEVR only; undefined correlations are reported as unavailable rather than as zero.

The HCC population follows the same rules as for the LLM UQ results (Appendix~\ref{app:stats}): it uses the scene-audited CLEVR items of Sec.~\ref{sec:benchmark}, keeps originally incorrect answers with a zero transition indicator and excludes perturbed refusals, after checks for score availability, test images and probe contexts.
It differs in the answers and the settings. The answers come from a separate 128-token greedy run on Qwen2.5-VL-7B and LLaVA-1.5-7B, and the settings cover every strength of each family and all text edits, including INV, SBJ and drop words, which can change the gold answer. Figure~\ref{fig:claim2_hallucination_hcc} instead uses one representative strength per family and omits these three edits; its panel (b) restricts the VLM-specific settings to these representative perturbations, but the transitions still come from the separate run.
For BICR, we combine the correlations of the five probes that exclude the test images by an item-weighted mean.

\begin{table*}[t]
\centering\scriptsize
\setlength{\tabcolsep}{2.5pt}
\begin{tabular}{lccccccc}
\toprule
 & \multicolumn{3}{c}{URR, median over settings} & \multicolumn{4}{c}{HCC($\Delta U$), CLEVR settings}\\
\cmidrule(lr){2-4}\cmidrule(lr){5-8}
\midrule
\multicolumn{8}{l}{\textbf{Qwen2.5-VL-7B} (correct$\to$wrong transitions per eligible pair: image edits 3,076/7,392 = 41.6\%; text edits 670/3,696 = 18.1\%)}\\
Score & Image (96) & Text (48) & Cross-modal (6) & Image (64) & Text (32) & CI$>$0 & CI$<$0\\
\multicolumn{8}{l}{\textit{VLM-specific UQ}}\\
VLU, 32 tok. & 0.393 & 0.317 & 0.343 & 0.056 & 0.014 & 14 & 3\\
\quad SE, 5 samples, 32 tok. & 0.387 & 0.310 & 0.314 & 0.051 & $-$0.023 & 14 & 11\\
\quad VLU, 128 tok. & 0.393 & 0.320 & 0.300 & 0.014 & 0.038 & 16 & 5\\
\quad SE, 5 samples, 128 tok. & 0.407 & 0.323 & 0.319 & 0.056 & $-$0.025 & 11 & 7\\
VAUQ & 0.707 & 0.543 & 0.584 & 0.177 & 0.103 & 32 & 2\\
\quad VAUQ unmasked $H$ & 0.763 & 0.637 & 0.622 & 0.163 & 0.086 & 36 & 2\\
\quad VAUQ random mask & 0.683 & 0.577 & 0.571 & 0.193 & 0.095 & 33 & 3\\
\quad VAUQ term $H_{\mathrm{mask}}{-}H$ & 0.513 & 0.473 & 0.463 & $-$0.082 & $-$0.077 & 1 & 20\\
Response SVAR & 0.843 & 0.487 & 0.607 & 0.100 & 0.001 & 28 & 8\\
Dropout score: aleatoric & 1.000 & 0.417 & 0.535 & 0.019 & 0.036 & 3 & 1\\
Dropout score: epistemic & 0.237 & 0.403 & 0.426 & 0.002 & $-$0.022 & 3 & 4\\
BICR & 0.830 & 0.647 & 0.639 & 0.137 & 0.153 & 36 & 6\\
\quad BICR loss control: BCE+Brier & 0.717 & 0.633 & 0.618 & 0.150 & 0.144 & 40 & 6\\
\quad BICR loss control: BCE & 0.673 & 0.610 & 0.560 & 0.103 & 0.124 & 35 & 4\\
\multicolumn{8}{l}{\textit{LLM UQ on the same greedy answer}}\\
MSP & 0.713 & 0.593 & 0.559 & 0.129 & 0.048 & 27 & 8\\
PPL (mean token NLL) & 0.760 & 0.603 & 0.589 & 0.156 & 0.075 & 33 & 3\\
MTE & 0.790 & 0.627 & 0.618 & 0.152 & 0.094 & 33 & 2\\
Answer length (control) & 0.547 & 0.387 & 0.491 & 0.069 & 0.023 & 24 & 5\\
\midrule
\multicolumn{8}{l}{\textbf{LLaVA-1.5-7B} (correct$\to$wrong transitions per eligible pair: image edits 502/7,392 = 6.8\%; text edits 159/3,696 = 4.3\%)}\\
Score & Image (96) & Text (48) & Cross-modal (6) & Image (64) & Text (32) & CI$>$0 & CI$<$0\\
\multicolumn{8}{l}{\textit{VLM-specific UQ}}\\
VLU, 32 tok. & 0.197 & 0.197 & 0.190 & $-$0.001 (60) & $-$0.045 (30) & 4 & 13\\
\quad SE, 5 samples, 32 tok. & 0.237 & 0.263 & 0.243 & 0.021 (60) & $-$0.020 (30) & 10 & 9\\
\quad VLU, 128 tok. & 0.227 & 0.237 & 0.208 & 0.039 (60) & $-$0.025 (30) & 9 & 8\\
\quad SE, 5 samples, 128 tok. & 0.273 & 0.290 & 0.227 & $-$0.004 (60) & $-$0.010 (30) & 6 & 11\\
VAUQ & 0.590 & 0.700 & 0.573 & $-$0.019 (60) & 0.023 (30) & 26 & 18\\
\quad VAUQ unmasked $H$ & 0.647 & 0.777 & 0.560 & $-$0.005 (60) & 0.060 (30) & 24 & 19\\
\quad VAUQ random mask & 0.623 & 0.757 & 0.561 & $-$0.007 (60) & 0.046 (30) & 24 & 20\\
\quad VAUQ term $H_{\mathrm{mask}}{-}H$ & 0.447 & 0.500 & 0.426 & $-$0.022 (60) & $-$0.038 (30) & 9 & 22\\
Response SVAR & 0.747 & 0.277 & 0.618 & 0.020 (60) & $-$0.031 (30) & 11 & 10\\
Dropout score: aleatoric & 0.640 & 0.420 & 0.485 & $-$0.011 (60) & 0.021 (30) & 12 & 8\\
Dropout score: epistemic & 0.333 & 0.407 & 0.482 & 0.006 (60) & $-$0.021 (30) & 5 & 11\\
BICR & 0.707 & 0.470 & 0.564 & 0.071 (60) & 0.069 (30) & 30 & 20\\
\quad BICR loss control: BCE+Brier & 0.617 & 0.460 & 0.591 & 0.023 (60) & 0.058 (30) & 17 & 22\\
\quad BICR loss control: BCE & 0.620 & 0.450 & 0.602 & 0.021 (60) & 0.053 (30) & 16 & 20\\
\multicolumn{8}{l}{\textit{LLM UQ on the same greedy answer}}\\
MSP & 0.560 & 0.743 & 0.527 & $-$0.017 (60) & 0.019 (30) & 17 & 23\\
PPL (mean token NLL) & 0.560 & 0.733 & 0.550 & $-$0.017 (60) & 0.056 (30) & 22 & 21\\
MTE & 0.647 & 0.777 & 0.571 & $-$0.008 (60) & 0.071 (30) & 23 & 20\\
Answer length (control) & 0.143 & 0.317 & 0.526 & $-$0.015 (57) & $-$0.014 (30) & 13 & 31\\
\bottomrule
\end{tabular}
\caption{Descriptive summary of the VLM-specific perturbation-response grids over all perturbation settings. Indented rows are matched controls or published components of the method above; the SE rows are LLM UQ controls sharing VLU\textquoteright s sampler, answer count and judge. URR is strict: a tie between original and perturbed uncertainty counts as no increase, which lowers URR for discrete sampling scores (VLU, SE); tie shares were not recorded for these runs. Setting counts are in parentheses; Cross-modal settings (CLIP mask, AMB, IVE) exist only on VizWiz and VQAv2. HCC($\Delta U$) is the point-biserial correlation between the uncertainty change and a correct$\to$wrong transition on scene-audited CLEVR pairs; originally wrong pairs stay in the population with transition 0, audit-flagged gold and perturbed refusals are excluded. This population differs in its answers and settings from that of the LLM UQ results. A number in parentheses after a median is the count of settings with a defined correlation. BICR rows use the item-weighted mean of correlations from five probes that exclude the test images. CI$>$0 / CI$<$0 count settings whose pointwise 95\% cluster-bootstrap interval excludes zero; they are unadjusted for the number of settings and serve only as a description of sign consistency.}
\label{tab:slim-native-risk}
\end{table*}
\section{Validity Checks}
\label{app:validity}
This appendix tests the detection and perturbation results against simpler explanations: edits that keep the answer, wording-only disagreement among the crowd answers and sampling noise. It closes with a control in which the ambiguity of the question is exact by construction.

\subsection{Answer-Preserving Edits}
\label{app:ext_benign}

\begin{table*}[t]
\centering
\scriptsize
\setlength{\tabcolsep}{3pt}
\renewcommand{\arraystretch}{1.05}
\caption{\textbf{Do scores rise for edits that keep the answer?} Uncertainty-rise rate (URR, ties counted as one half; 0.50 is no effect) over the 900 clean VizWiz, VQAv2 and CLEVR items, averaged over the four shared scores. \emph{Benign} averages five edits intended to preserve the gold answer (JPEG re-encoding, a thin border crop, a small brightness change, a punctuation or casing change, an entailment-checked paraphrase); \emph{Split-half reference} compares two halves of the samples of the same unedited input; \emph{Harmful} averages the benchmark's visual and question perturbations on the same items. $^\ddagger$Qwen2.5-VL-7B's harmful values come from an earlier sampling pipeline. Brackets give 95\% bootstrap intervals over images.}
\label{tab:ext_benign}
\resizebox{\ifdim\width>\textwidth\textwidth\else\width\fi}{!}{%
\begin{tabular}{@{}lccccc@{}}
\toprule
\textbf{Model} & \textbf{Benign URR} & \textbf{Sampling-noise level} & \textbf{Benign $-$ noise level} & \textbf{Harmful URR} & \textbf{Benign $-$ harmful} \\
\midrule
Qwen2.5-VL-7B$^\ddagger$ & 0.49 [0.47, 0.51] & 0.50 [0.48, 0.53] & $-$0.01 [$-$0.04, 0.01] & 0.71 [0.69, 0.72] & $-$0.22 [$-$0.24, $-$0.20] \\
Gemma 4 12B & 0.52 [0.50, 0.54] & 0.50 [0.48, 0.52] & +0.02 [$-$0.01, 0.05] & 0.62 [0.61, 0.64] & $-$0.10 [$-$0.12, $-$0.08] \\
Gemini 3.8 Flash & 0.51 [0.49, 0.53] & 0.51 [0.49, 0.53] & 0.00 [$-$0.03, 0.03] & 0.64 [0.63, 0.65] & $-$0.13 [$-$0.15, $-$0.11] \\
Gemini 3.1 Pro & 0.50 [0.48, 0.52] & 0.52 [0.49, 0.54] & $-$0.02 [$-$0.05, 0.01] & 0.63 [0.61, 0.64] & $-$0.13 [$-$0.15, $-$0.11] \\
\bottomrule
\end{tabular}}
\end{table*}

URR (Sec.~\ref{sec:evaluation}) should respond to the kind of change made to the input, not to any change.
We apply five edits meant to keep the gold answer to 900 clean VizWiz, VQAv2 and CLEVR items: JPEG re-encoding at quality 70, a 2.5\% crop on each side, a $\pm$10\% brightness change, a punctuation or casing change, and a paraphrase that we keep only if it entails the original question and is entailed by it. We compare their URR with the split-half sampling reference and with the benchmark's harmful perturbations of the same items (Table~\ref{tab:ext_benign}). %

On all four models (Qwen2.5-VL-7B, Gemma 4 12B and the two Gemini models), benign URR lies between $-$0.02 and +0.02 of the sampling-noise level. %
Harmful perturbations of the same items raise URR by 0.11--0.20 over the reference, so benign URR stays 0.10--0.22 below harmful URR on every model. %
On Gemma 4 12B, the small benign rise (0.17 of its harmful rise) comes from the two question edits (URR 0.56), while its image edits stay at 0.50. %
The scores therefore rise less for these small edits than for the benchmark perturbations, which indicates that they respond to the kind of change. We did not verify on every item that the edits keep the answer.

\subsection{Wording-Only Disagreement}
\label{app:ext_wording}

\begin{table*}[t]
\centering
\scriptsize
\setlength{\tabcolsep}{3pt}
\renewcommand{\arraystretch}{1.05}
\caption{\textbf{Does wording-only disagreement explain the Cross-modal result?} AUROC for ranking VizWiz training-pool items whose crowd answers differ only by synonyms or granularity above clean items selected with our Clean rule (Appendix~\ref{app:vizwiz}; 0.50: no difference). \emph{Unmatched} uses 300 and 300 items; \emph{Matched}, added after the data, keeps only items without a crowd answer ``unanswerable'' and clean items within the wording set's reason limits (258 and 73 items). Brackets give 95\% bootstrap intervals over images.}
\label{tab:ext_wording}
\resizebox{\ifdim\width>\textwidth\textwidth\else\width\fi}{!}{%
\begin{tabular}{@{}lcccccc@{}}
\toprule
 & \multicolumn{2}{c}{\textbf{Qwen2.5-VL-7B}} & \multicolumn{2}{c}{\textbf{Gemma 4 12B}} & \multicolumn{2}{c}{\textbf{Gemini 3.8 Flash}} \\
\cmidrule(lr){2-3} \cmidrule(lr){4-5} \cmidrule(lr){6-7}
\textbf{Score} & Unmatched & Matched & Unmatched & Matched & Unmatched & Matched \\
\midrule
Lexical similarity & 0.35 [0.30, 0.39] & 0.46 [0.38, 0.53] & 0.39 [0.34, 0.43] & 0.50 [0.42, 0.58] & 0.30 [0.26, 0.34] & 0.42 [0.35, 0.50] \\
Degree matrix & 0.37 [0.32, 0.41] & 0.47 [0.40, 0.56] & 0.42 [0.37, 0.46] & 0.49 [0.41, 0.57] & 0.37 [0.33, 0.41] & 0.50 [0.42, 0.57] \\
Semantic entropy & 0.36 [0.31, 0.40] & 0.48 [0.40, 0.55] & 0.39 [0.34, 0.43] & 0.49 [0.41, 0.57] & 0.35 [0.31, 0.40] & 0.47 [0.40, 0.55] \\
LUQ & 0.38 [0.33, 0.42] & 0.46 [0.38, 0.53] & 0.43 [0.39, 0.48] & 0.51 [0.43, 0.59] & 0.39 [0.35, 0.44] & 0.50 [0.43, 0.58] \\
\bottomrule
\end{tabular}}
\end{table*}

Cross-modal items differ from clean ones in answer wording as well as in source (Appendix~\ref{app:vizwiz}), so their AUROCs could reflect how differently the crowd phrased its answers.
On a separate VizWiz training pool, we let each sampling-based score of Qwen2.5-VL-7B, Gemma 4 12B and Gemini 3.8 Flash rank 300 items whose crowd answers differ only by synonyms or granularity against 300 clean items selected with our Clean rule (Appendix~\ref{app:vizwiz}; Table~\ref{tab:ext_wording}). %

Without matching, the wording-only items receive \emph{lower} uncertainty than the clean items (AUROC 0.30--0.43 for all 12 model--score pairs). %
This reversal follows from the composition of the clean set: 21.4\% of its crowd answers are ``unanswerable'', against 1.9\% for the wording-only items, and 64\% of its items exceed the wording set's reason limits. %
When we match the two sets on answerability and on these reason limits (258 and 73 items), the AUROCs lie between 0.42 and 0.51, and none differs from 0.50 after Holm correction. %
Within our Cross-modal subset, the 20 items without a synonym majority separate from clean items about as well as the 102 with one (median AUROC 0.75 vs.\ 0.72 over the 12 model--score pairs; semantic entropy 0.89 vs.\ 0.78 on Qwen2.5-VL-7B and 0.81 vs.\ 0.81 on Gemma 4 12B). %
We defined the matched comparison after seeing the data, and neither comparison matches the two sets on image content.
Once the two sets are matched, wording-only disagreement does not raise sampled uncertainty, and we find no evidence that it accounts for the Cross-modal AUROCs.

\subsection{Sampling Robustness}
\label{app:ext_retest}

\begin{table}[t]
\centering
\scriptsize
\setlength{\tabcolsep}{3pt}
\renewcommand{\arraystretch}{1.05}
\caption{\textbf{Sampling pipeline and detection for Qwen2.5-VL-7B.} AUROC$\uparrow$ for separating each VizWiz source from the clean items. \emph{Orig.}: the earlier generation pipeline behind Table~\ref{tab:det_llmuq_full_auroc}; \emph{Rerun}: the sampling pipeline of the other models (first draw).}
\label{tab:ext_retest}
\resizebox{\ifdim\width>\columnwidth\columnwidth\else\width\fi}{!}{%
\begin{tabular}{@{}lcccccc@{}}
\toprule
 & \multicolumn{2}{c}{\textbf{Image}} & \multicolumn{2}{c}{\textbf{Text}} & \multicolumn{2}{c}{\textbf{Cross-modal}} \\
\cmidrule(lr){2-3} \cmidrule(lr){4-5} \cmidrule(lr){6-7}
\textbf{Score} & Orig. & Rerun & Orig. & Rerun & Orig. & Rerun \\
\midrule
Semantic entropy & 0.85 & 0.84 & 0.89 & 0.87 & 0.80 & 0.78 \\
Degree matrix & 0.73 & 0.83 & 0.70 & 0.81 & 0.60 & 0.76 \\
Lexical similarity & 0.85 & 0.79 & 0.87 & 0.82 & 0.76 & 0.69 \\
LUQ & 0.61 & 0.77 & 0.63 & 0.77 & 0.54 & 0.70 \\
\bottomrule
\end{tabular}}
\end{table}

For four models, we draw a second set of ten answers with a new seed on the VizWiz items and correlate each sampling-based score across the two draws (Spearman $\rho$). Within a single source, the lowest correlation is 0.72 for Qwen2.5-VL-7B, 0.71 for Gemma 4 12B, 0.64 for Gemini 3.1 Pro and 0.52 for Gemini 3.8 Flash, so small leads between scores are not stable: the best score of Gemini 3.8 Flash on Cross-modal items changes on the second draw, although it led by 0.041 AUROC on the first. %
The generation pipeline matters more than the draw: Qwen2.5-VL-7B's AUROCs under its original, earlier sampler differ from both draws under the other models' sampler by up to 0.17 (LUQ on Cross-modal items: 0.537 vs.\ 0.700 and 0.705; Table~\ref{tab:ext_retest}), while semantic entropy remains its best sampling-based score on all three VizWiz sources in every run. %

\subsection{Exact Referential Ambiguity on CLEVR}
\label{app:ext_clevr_amb}

On CLEVR, the scene graph gives the answer set $A$ of every question exactly. We edit each attribute or relation question that passed the label audit (Appendix~\ref{app:clevr}) so that two or more objects fit the description and the queried value differs across them (DUP, $|A| \ge 2$; e.g., ``What shape is the yellow metal object?'' becomes ``What shape is the metal object?'' in a scene with a metal cube and a metal sphere), and pair it with a control that keeps the answer unique (DROP-UNIQ, $|A| = 1$; a deletion of the same number of attribute words on 85 of 131 pairs, a paraphrase otherwise). We report $\Delta_{\text{ctrl}} = \text{URR(DUP)} - \text{URR(DROP-UNIQ)}$, with the tie-corrected URR averaged over the four sampling-based scores (Table~\ref{tab:new_pipe_clevr}). %

\begin{table*}[t]
\centering
\small
\setlength{\tabcolsep}{4pt}
\caption{\textbf{Exact referential ambiguity on CLEVR} (tie-corrected URR, mean of the four sampling-based scores, against the same fresh clean run). $\Delta_{\text{ctrl}}$: URR(DUP) $-$ URR(DROP-UNIQ) on 131 attribute and relation items (108 attribute), 113 images. One-sided image-cluster sign-flip $p \le 10^{-4}$ for every model (the floor of 10{,}000 draws); Holm-adjusted $p \le 3 \times 10^{-4}$ over Gemma 4 12B, Qwen2.5-VL-7B and Gemini 3.1 Pro. \emph{True deletions}: the 85 pairs whose control is a deletion rather than a paraphrase. \emph{FP}: false premise ($n = 145$); \emph{denial rise}: greedy abstention or denial share minus that on the clean items. $^{\dagger}$Excluded because of answer truncation: Flash's greedy answers are truncated by the output budget in 0.75 (attribute) and 0.97 (relation) of DUP items, against 0.06 and 0.59 on the clean items, and its no-truncation sensitivity keeps 7 pairs. Its values are shown in parentheses for reference only. Brackets give 95\% image-cluster bootstrap intervals.}
\label{tab:new_pipe_clevr}
\begin{tabular}{@{}lccccc@{}}
\toprule
\textbf{Model} & \textbf{URR(DUP)} & \textbf{$\Delta_{\text{ctrl}}$} & \textbf{$\Delta_{\text{ctrl}}$, true deletions} & \textbf{URR(FP)} & \textbf{FP denial rise} \\
\midrule
Gemma 4 12B & 0.789 & 0.279 $[0.195, 0.355]$ & 0.301 $[0.194, 0.401]$ & 0.676 & $+0.59$ \\
Qwen2.5-VL-7B & 0.875 & 0.282 $[0.215, 0.348]$ & 0.307 $[0.224, 0.390]$ & 0.788 & $+0.94$ \\
Gemini 3.1 Pro & 0.761 & 0.220 $[0.125, 0.310]$ & 0.237 $[0.120, 0.347]$ & 0.608 & $+0.81$ \\
\midrule
Gemini 3.8 Flash$^{\dagger}$ & (0.816) & (0.269 $[0.196, 0.346]$) & (0.216 $[0.122, 0.309]$) & (0.684) & ($+0.75$) \\
\bottomrule
\end{tabular}
\end{table*}

DUP raises URR more than the answer-preserving control: $\Delta_{\text{ctrl}}$ is 0.279 for Gemma 4 12B, 0.282 for Qwen2.5-VL-7B and 0.220 for Gemini 3.1 Pro. Part of this rise comes from answers that list the matching objects (on attribute questions, 0.79 of Gemini 3.1 Pro's DUP samples mention several referents, against 0.03 of its clean samples), and Gemini 3.8 Flash is shown only for reference because its answers are truncated by the output budget. Gemini 3.1 Pro's greedy answers to attribute DUP questions also hit its output budget more often than its clean answers (0.25 against 0.00); on the 50 pairs without any truncated output, its $\Delta_{\text{ctrl}}$ is 0.085 $[-0.086, 0.235]$. %
This result shows that sampled uncertainty responds to an answer-divergent ambiguity that exists by construction, which is the best case for scores based on sampling diversity. It does not show that the scores detect subtle cross-modal rewrites, which change the meaning of the input without making the answer set diverge, and on which URR exceeds the sampling-noise level far less often (Appendix~\ref{app:res_urr}).
\end{document}